\documentclass[lettersize,journal]{IEEEtran}
\usepackage{amsmath,amsfonts}
\usepackage[caption=false,font=normalsize,labelfont=sf,textfont=sf]{subfig}
\usepackage{times}
\IEEEoverridecommandlockouts
\usepackage{enumitem}
\usepackage{tabularx}
\usepackage{lipsum}
\usepackage{cite}
 \usepackage[usenames,dvipsnames]{xcolor} \usepackage{soul}
\usepackage{graphicx,color}
\usepackage{algorithm}
\usepackage[noend]{algpseudocode}
\usepackage[font=singlespacing]{caption}
\usepackage{lettrine}
\usepackage{wrapfig}
\usepackage{float}
 \usepackage{graphicx}\usepackage{amsmath,amssymb,amsfonts,amsthm}
\usepackage[utf8]{inputenc}
\theoremstyle{definition}

\usepackage{graphicx}
\usepackage{graphics}
\usepackage{xcolor}
\usepackage{float} 
\newcommand\MM{\mathcal{M}}

\newcommand\XX{\mathcal{X}}

\newcommand\LL{\mathcal{L}}

\def\1{\mathbf{1}}
\newcommand{\Ni}{\mathcal{N}_i}

\title{How Physicality Enables Trust: A New Era of Trust-Centered Cyberphysical Systems}
\author{Stephanie Gil, Michal Yemini, Arsenia Chorti, Angelia Nedi\'{c}, H.~Vincent Poor, Andrea J.~Goldsmith
\thanks{Stephanie Gil is with the School of Engineering and Applied Sciences,
Harvard University, Allston, MA 02134 USA (e-mail: sgil@seas.harvard.edu). Michal Yemini is with the Faculty of Engineering, Bar-Ilan University, Ramat-Gan 5290002 Israel (email: michal.yemini@biu.ac.il). Arsenia Chorti is with the Wireless Connectivity Group, Barkhausen
Institut gGmbH, 01187 Dresden, Germany, and also with ETIS UMR 8051,
CY Paris Cergy University, ENSEA, CNRS, 95011 Cergy, France (e-mail:
arsenia.chorti@barkhauseninstitut.org).
Angelia Nedi\'{c} is with the School of Electrical, Computer and Energy
Engineering, Arizona State University, Tempe, AZ 85281 USA (e-mail:
Angelia.Nedich@asu.edu). H.~Vincent Poor and Andrea J.\ Goldsmith are with the Electrical and
Computer Engineering Department, Princeton University, Princeton, NJ 08544
USA (e-mails: poor@princeton.edu; goldsmith@princeton.edu).}
\thanks{%This work has been supported by the NSF award CNS 2147641.
This work was partially supported by NSF awards CNS-2147631,  CNS-2147641 and CNS-2147694 and ONR YIP award $\#$N00014-21-1-2714. M.~Yemini and A.~J.~Goldsmith were partially supported by  AFOSR award \#002484665. 
}
}

\newcommand{\an}{\textcolor{violet}}

\newcommand{\my}{\textcolor{cyan}}

\newcounter{sideremark}

\usepackage{tcolorbox}
\usepackage{lipsum}
\usepackage{xpatch}
\tcbuselibrary{skins,breakable}
\usetikzlibrary{shadings,shadows}

\newenvironment{svblock}[1]{%
    \tcolorbox[beamer,breakable,enhanced,
    colback=white!95!NavyBlue,colframe=white!78!NavyBlue,%
    colbacktitle = white!80!NavyBlue,
    colbacklower=white!,
    title={#1}, coltitle=black]}%
    {\endtcolorbox}

\makeatletter
\newinsert\copyrightnote@ins
\count\copyrightnote@ins=1000
\dimen\copyrightnote@ins=8in
\newcommand*\copyrightnote@hook
  {%
    \unvbox\copyrightnote@ins
    \global\let\@makecol\copyrightnote@makecol
  }
\let\copyrightnote@AtBeginDocument\AtBeginDocument
\AtBeginDocument{\let\copyrightnote@AtBeginDocument\@firstofone}
\newcommand*\copyrightnote@firstuse
  {%
    \gdef\copyrightnote@firstuse
      {\global\skip\copyrightnote@ins=\bigskipamount\gdef\copyrightnote@firstuse{}}%
    \global\let\copyrightnote@makecol\@makecol
    \xpatchcmd\@makecol{\unvbox\footins}{\unvbox\footins\copyrightnote@hook}
      {}{\GenericError{}{patching @makecol failed}{}{}}
    \copyrightnote@AtBeginDocument
      {%
        \ifvoid\footins
          \insert\footins{}% 
        \else
          \global\skip\copyrightnote@ins=\bigskipamount
        \fi
      }%
  }
\newcommand\copyrightnote[1]
  {%
    \copyrightnote@firstuse
    \copyrightnote@AtBeginDocument
      {%
        \insert\copyrightnote@ins
          {%
            \reset@font\footnotesize
            \interlinepenalty\interfootnotelinepenalty
            \splittopskip\footnotesep
            \splitmaxdepth\dp\strutbox
            \floatingpenalty\@MM
            \hsize\columnwidth
            \@parboxrestore
            \color@begingroup
            \parindent1em
            \noindent
            \hskip1.8em
            \ignorespaces #1\@finalstrut\strutbox
            \color@endgroup
          }%
      }%
  }
\makeatother

\begin{document}

\maketitle

\copyrightnote{This work has been submitted to the IEEE for possible publication. Copyright may be transferred without notice, after which this version may no longer be accessible.}

% Points and ideas to take from meeting:
% \begin{itemize}
%     \item Goal oriented security (or context aware).
%     \item We should include the term ``sensing" explictly so it is clear that we are not limited to RF measurements only.
%     \item We use the physicality of the problem to create trust, this is in-contrast to current approach which leaves security to the application level.
%     \item We should consider the case where a system is completely spoofed, even if to state that this case is beyond the scope of our survey.
% \end{itemize}

\begin{abstract}
Multi-agent cyberphysical systems enable new capabilities in efficiency, resilience, and security. The unique characteristics of these systems prompt a reevaluation of their security concepts, including their vulnerabilities, and mechanisms to mitigate these vulnerabilities.  This survey paper examines how advancement in wireless networking, coupled with the sensing and computing in cyberphysical systems, can foster novel security capabilities. This study delves into three main themes related to securing multi-agent cyberphysical systems. First, we discuss the threats that are particularly relevant to multi-agent cyberphysical systems given the potential lack of trust between agents. Second, we present prospects for sensing, contextual awareness, and authentication, enabling the inference and measurement of ``inter-agent trust" for these systems. Third, we elaborate on the application of quantifiable trust notions to enable ``resilient coordination," where ``resilient" signifies sustained functionality amid attacks on multiagent cyberphysical systems. We refer to the capability of cyberphysical systems to self-organize, and coordinate to achieve a task, as \emph{autonomy}.  This survey unveils the cyberphysical character of future interconnected systems as a pivotal catalyst for realizing robust, trust-centered autonomy in tomorrow's world.

%The increasingly critical role of autonomy and networked cyberphysical systems in society, is prompting a reevaluation of current security concepts, their limitations, and the discovery of fresh avenues for safeguarding these systems.

% The increasing prevalence of autonomy and networked cyberphysical systems (CPS) is motivating a need to revisit concepts of security, and identify new opportunities for securing for these systems. This survey paper reviews how the expected evolution of wireless networking, along with onboard sensing and computing capabilities of future CPS, can support new capabilities and opportunities for security in cyberphysical systems. To achieve this, this survey paper reviews the state of the art in i) threats that are of particular relevance to distributed, multiagent cyberphysical systems, 2) new opportunities for sensing, contextual awareness, and authentication towards inferring and measuring \emph{inter-agent trust}, and 3) using quantifiable concepts of trust to provide \emph{resilient coordination}. Here, ``resilient'' refers to nominal or close to nominal operation of multiagent cyberphysical systems in the case of an attack. In this way, we reveal the cyberphysicality of future networked systems as an important enabler for achieving resilient, trust-centered autonomy in tomorrow's world.
\end{abstract}

% \section*{Links to the Overleaf project}
% Editable: https://www.overleaf.com/6332291453zhxhtmbyrwdn

% View-only: https://www.overleaf.com/read/vfdxnjykgfmd
% `\mynote{TODO: add Andrea' NSF and AFSOR grants, Stephnie's grants and all others. Let's ask everyone to add their funding when we send them the paper}
\section{Introduction}
\label{sec:intro}

The future is ushering in ubiquitous multi-agent networks of physical devices, such as multi-robot systems, smart and connected vehicles, and fleets of delivery drones. These systems are 1) connected, either globally through a cloud, or through a local network, or through a hybrid of these two, and where this network could be any wireless or wired network, 2) embodied, and thus they \emph{physically interact} with each other and/or with their environment, and 3) have computation ability on board, i.e., they have \emph{a cyber component} whereby they perform computations based on the information they receive from the network and they can act on this information. Throughout, we will refer to these systems as \emph{multi-agent cyberphysical systems}. These multi-agent cyberphysical systems will be actuated by running algorithms onboard and communicating with other agents to cooperatively and distributively achieve a wide array of critical tasks. From automated deliveries, to environmental monitoring, to search and rescue -- these multi-agent cyberphysical systems have the potential to change the way that societies and economies operate. 

The integration of multi-agent cyberphysical systems into the world around us will impose new challenges and \emph{new opportunities} for securing these systems. This article focuses on how the physicality of these systems can be exploited to provide novel avenues for security.  We will review advances in sensing and communication for future cellular and broader wireless networks. The emphasis here is not on the type of network used, but rather on how \emph{the physicality of the multi-agent system might be exploited for security}, and for attaining resilient coordination for the multi-agent cyberphysical systems that are connected by such networks. This is possible both in cellular and broader wireless networks and thus our survey article will review advances relevant to both network types. In particular, the promise of new sensing and communication abilities brought forth by 6G and future wireless networks, together with onboard intelligence and sensors that will be pervasive in future cyberphysical networks, can be combined to provide powerful new paradigms for securing these systems.

% \my{(MY) TODO: I revised the above paragraph to convey the message of 1) the new dependence of humanity on physical systems 2) these systems are composed of multiple agents, and are thus distributed in nature and rely on collaboration. Then, the discussion in the next paragraph becomes a natural extension of this one where we discuss the threats to the reliable operation of distributed multi-agent physical systems.}
% \\ 

Achieving this overarching promise requires a new understanding across traditionally disparate fields such as communication and robotics. This is because the concept of security for these multi-agent emerging networks is inherently different from that in today's networks. One challenge arises due to the heterogeneity of devices that will be supported in wireless networks, including smartphones, tablets, Internet of Things (IoT) devices\footnote{By IoT devices we specifically refer to those user devices that are highly computationally and communication resource constrained.}, smart vehicles, and drones. These are highly heterogeneous in their computing and energy resources and have widely differing latency and privacy requirements. 

Additionally, future cyberphysical systems will make decisions and act without a human-in-the-loop. As a result their behavior or performance on a given task must be taken into account when considering their trustworthiness. In contrast, today's networks rely heavily on authentication for securing trust. Here, \emph{authentication} is the validation of a particular networked agents' identity, and \emph{trust} corresponds to an agents' rightful access to the network. Oftentimes these two concepts are taken to be synonymous, however, for systems where agents act in the physical world, their behavior should also factor into permitting their access to the network. That is to say, our bar for trustworthiness must be high, and must consider the full loop that includes an agents' behavior. The behavior of these systems results from both the inputs (network) and algorithms (controls) running onboard each of the user devices.  Thus, authentication alone is not enough and \emph{authentication does not equal trustworthiness}. We highlight this statement in the following ``key challenge,'' which is the first of several that we will provide to summarize main takeaways of this survey article:  \\

 %\vspace{0.1in}
 
\noindent \emph{\textbf{Key Challenge 1: Authentication alone is no longer sufficient for newly emerging multiagent cyberphysical systems. \emph{Trustworthiness} will additionally need to capture trust in the \emph{algorithms} that control or generate the networked devices' behavior. }}\\

 %\vspace{0.02in}
A required evolution of prior control mechanisms for individual agents should include a feedback loop whereby both the network and the physical behavior (performance) of the system will feed back to a new measure of trustworthiness. In turn, an agent's controller should account for the trustworthiness of its peers in the network when determining its own actions. In order to accomplish this we must understand: \textit{how can communication networks be leveraged to provide new mechanisms of data validation and “trust” for cyberphysical systems, and how does this trust translate to resilience in performance for these systems?} 

\begin{figure}
\centering
      \includegraphics[scale=0.06]{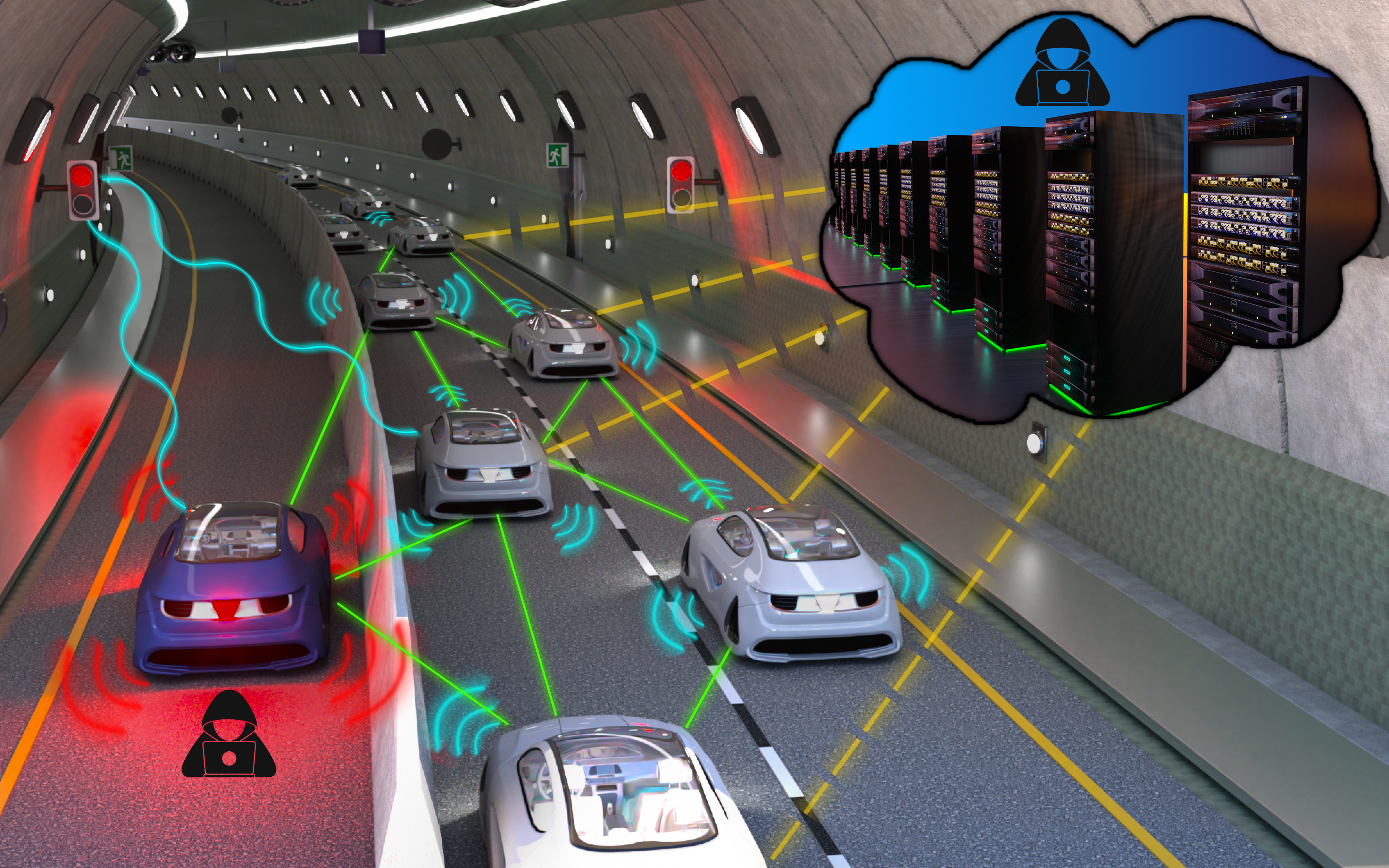}
      % \vspace{-0.1in}
  \caption{Emerging networks, applications, and vulnerabilities. Vulnerabilities can include hacking of the devices (including their onboard sensors), network links, and/or servers.}
  \vspace{-0.2in}
  \label{fig:emergingNetworks}
  \end{figure}

  %We argue that quantifying \emph{trust} over the links in the network, or between agents, will emerge as a powerful abstraction to bridge these two pieces.    
To start answering these questions, we must better characterize the connection between the network and the performance of the networked system. Future networks will support many multi-agent applications and assessing their performance will depend on the task. We divide these tasks into two broad categories of \textbf{\emph{Democratized Sensing}} and \textbf{\emph{Democratized Computing}}. \emph{Democratized Sensing} tasks are those where agents use the communication network to share locally sensed data and form a global view of the world. Some examples of democratized sensing tasks include crowdsensing, distributed localization, coverage, distributed event detection, and distributed mapping~\cite{STAR,crowdsensing,mobileCrowdsensing,coverageSensorNetworks,bulloCoverage,eventDetectionDisasterManagement,CPSdistributedEventDetection,sensorNetworksEventDetection,carloneDistMappingKimera,distributedMappingCarlone,beltrameDistMapping,distLocalizationMappingSastry,distributedExplorationMapping}.  \emph{Democratized Computing} tasks are those where agents use the communication network to exchange partial computational results with one another to compute a quantity that is more global in nature. Some examples of democratized computing tasks include distributed optimization, federated learning, distributed estimation, and decentralized control~\cite{DistOptAngelia,distoptNotarstefano,distributedControlSonia,distributedDynamicConsensusSonia,distributedEstimationSonia,distributedOptimizationControlAngelia,fedLearning6G,fedLearningSurvey,pushpullDistributedOptimization,zeno,krum}. More details of these examples are as follows:
\begin{itemize}[leftmargin=*]
    \item \textbf{Crowd-sourced sensing} where multiple agents share information that is locally sensed in order to make a decision on some event of global interest~\cite{cloudClusterTWC,STAR,SundaramHypTesting}. This could include smart and connected vehicles and infrastructure, sharing information about the current state of traffic on the road, or agents deploying themselves to provide surveillance coverage of a particular area~\cite{schwager2017}.
    \item \textbf{Consensus tasks} where networked agents must reach \emph{agreement} on some quantity of interest such as a heading direction, velocity, or estimation of some quantity~\cite{distributedEstimationSonia,distributedDynamicConsensusSonia}. Some future application areas for these types of users include traffic and intersection management for autonomous vehicles, flocking tasks, and crowdsensing measurements of some average quantity over an environment by a sensor network.
    \item \textbf{Distributed optimization tasks} where networked agents must distributedly compute some quantity of interest by jointly minimizing a cost function~\cite{DistOptAngelia}. Some application areas for these use cases include federated learning in future mobile multi-agent systems~\cite{fedLearning6G}.
    %\item \textbf{Blockchain-enabled} use cases include Mobile Edge Computing that require a trust mechanism to resist selfish edge attacks in order to fruitfully utilize the computational resources of various edge devices~\cite{blockchainPoor}. Blockchain in this context can improve security and protect data privacy for IoT applications.\\ 
\end{itemize}
%\vspace{-0.1in}
In all aforementioned use-cases there is some objective to be completed, for example degree of agreement achieved in the case of consensus, effectiveness of deployment for the multi-agent system in the case of coverage, or accuracy of the computed quantity in the case of distributed optimization. The degree to which success can be achieved for this objective is what we refer to as the \textbf{\emph{performance}} of the system. The degree to which this performance can be attained in the face of an attack or untrustworthy agents in the network, is what we refer to as the \textbf{\emph{resilience}} of the system. Currently, \emph{there is a need for quantifying the \emph{performance/resilience} tradeoff in a multi-agent system resulting from untrustworthy agents or links in the network.}  

Some works have made progress in quantifying this impact (see~\cite{bulloUnreliable,Scaglione19,TRO_foundations,SundaramHypTesting} for example), nonetheless, theoretical results in this area remain sparse. A previous survey paper on the ability to provide graceful degradation of operational goals for cyberphysical systems when under attack, pointed to the lack of tools necessary for attaining theoretical guarantees along these lines~\cite{Sastry_survivableCPS}. We argue that we still lack a unified framework by which to reason quantitatively about the full feedback loop, namely: quantifying the impact of untrustworthy links and untrustworthy agents on the final performance guarantees available for cyberphysical systems. We summarize this observation in the following challenge.\\

\vspace{-0.03in}
\noindent \emph{\textbf{Key Challenge 2: We currently lack a unified framework for quantifying the impact of attacks at the network level on performance guarantees of multi-agent cyberphysical systems.}}\\
\vspace{-0.04in}

This survey article argues that the emergence of ubiquitous networks of multi-agent cyberphysical devices brings with it a new opportunity for redefining security or trustworthiness for these systems. We refer to trustworthiness centered on the physicality of these systems as \emph{physicality-based trust}. Indeed, today we are on the cusp of a paradigm shift in how we design and use future networks, with physics-based trust emerging as a principal enabler for resilience and security.

%\vspace{-0.1in}
  % \begin{minipage}{\linewidth}
  %     \centering
  %     \hspace{-0.7cm}
  %     \begin{minipage}{0.5\linewidth}
      \begin{svblock}{}
\textbf{\textit{The time is ripe to consider a new security paradigm in which the physicality of emerging networks is exploited to create and quantify trust. This is in contrast to current approaches that 
%largely leave security to the application level \my{and }.}}  
largely ignore opportunities to exploit the physical layer in security and trust building.
}}
\end{svblock}
      % \end{minipage}
      % \hspace{0.05\linewidth}
      % \begin{minipage}{0.45\linewidth}
      % \vspace{-0.5cm}
      \begin{figure}
      \centering
           \includegraphics[scale=0.2]{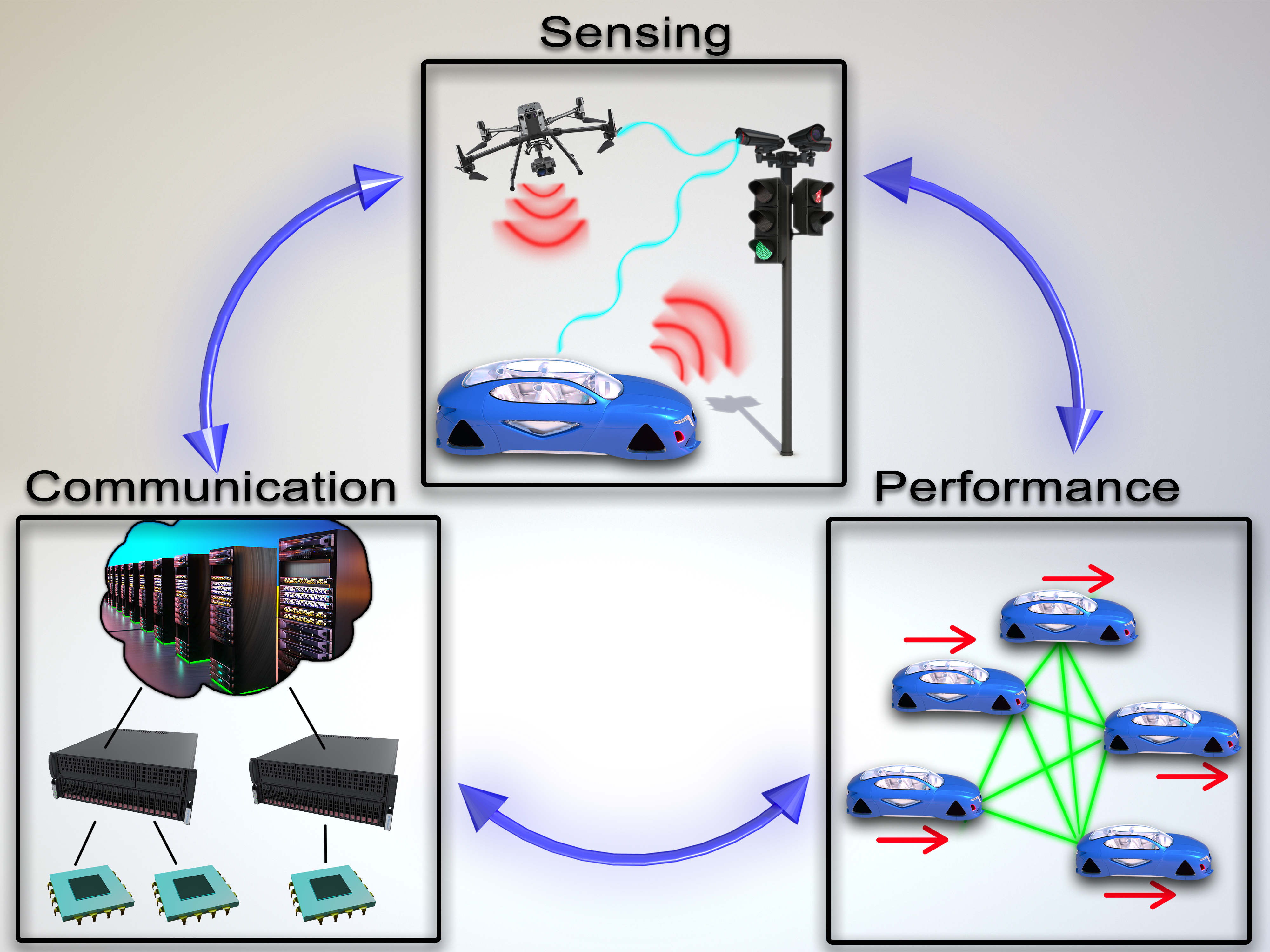}
    % \vspace{-0.1in}
  \caption{Interconnection between cyber and physical sides of emerging networks give rise to novel opportunities for security.}
  % \vspace{-0.25in}
  \label{fig:feedback}
  \end{figure}

Importantly, we distinguish between security and resilience: \emph{security} signifies the ability to detect or prevent an attack, whereas \emph{resilience} indicates the ability of an agent or a whole system to overcome an attack once it occurs. Note that in some cases one may not need to explicitly detect an attacking agent to be resilient to it \cite{DOLEV198214,resilience_no_detection_2008}.

The use of physicality-based trust metrics for allowing the \emph{quantification of trust} makes way for \emph{new opportunities} of characterizing performance impact more readily. This leads to the following key challenge:\\

\vspace{-0.02in}
\noindent \emph{\textbf{Key Challenge 3: The quantification of trust allows for connecting physicality-based methods of detection on the communication and networking side with the control and performance outcomes of multi-agent cyberphysical systems. This leads to new challenges of quantifying resilience guarantees for these systems.}}\\
\vspace{-0.02in}

%\label{sec:trust_utilization}
\begin{figure*}
  \centering
    \includegraphics[height=7cm, width=13cm]{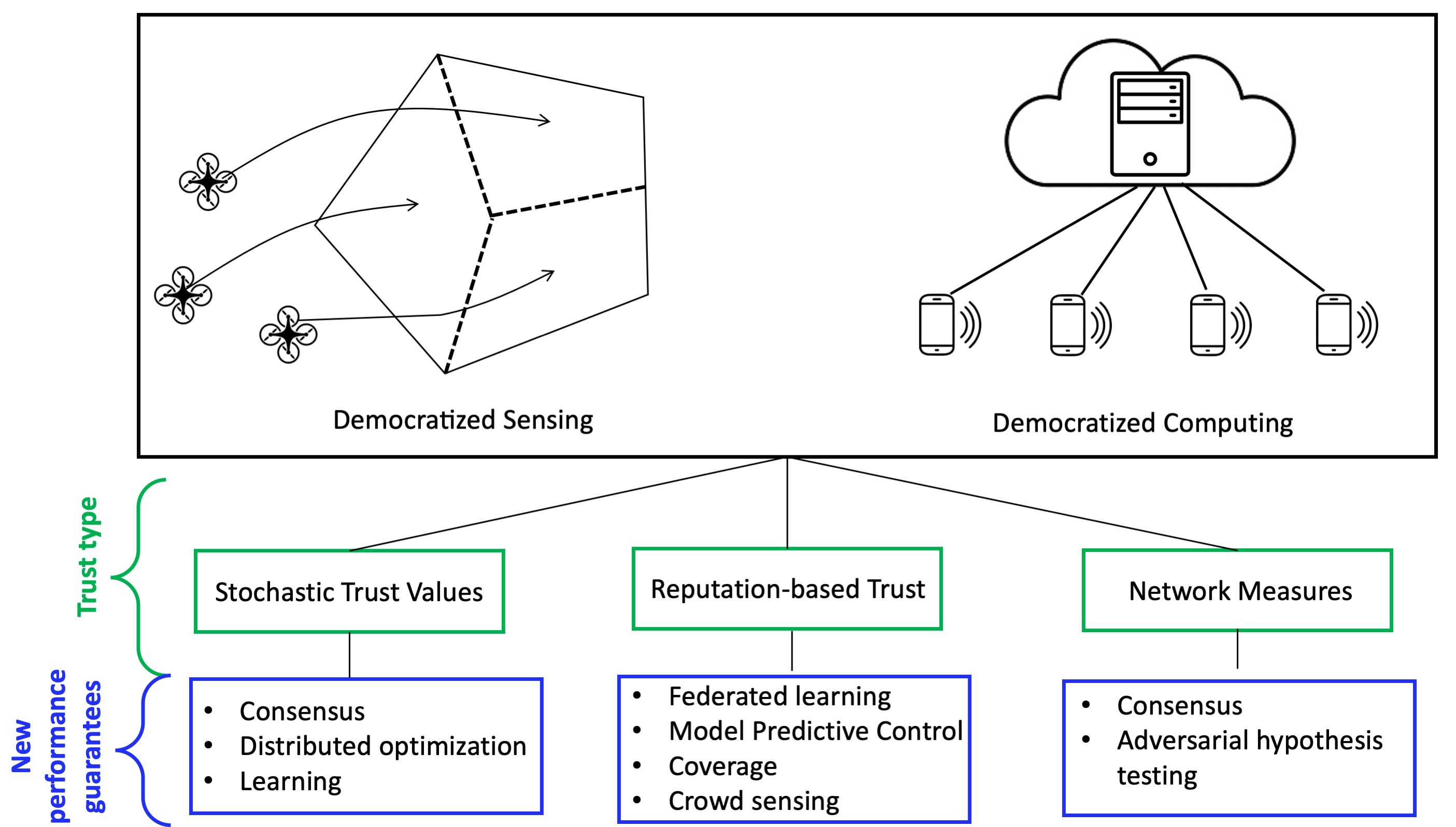}
  \caption{Emerging network applications and performance areas. %\my{Can we flip the colored test on the left (it's hard to read)?}
  }
  \label{fig:trustSchematic}
\end{figure*}

New developments for future networking infrastructure may present the option of the quantification of trust for the first time. Research in this direction is motivated by several challenges related to future network demands. Paramount among these are securing wireless links under aggressive latency constraints, and scaling authentication and key distribution to massive numbers, for example, to accommodate IoT devices. It is envisioned that Physical Layer Security (PLS) solutions will provide new levels of security by complimenting conventional upper-layer security schemes and strengthening the overall trust and resilience of the sixth generation (6G) networks~\cite{Chorti22}. These methods work by exploiting the characteristics of physical phenomena to provide security, including phenomena such as fading, interference, diversity, reciprocity, directivity and energy concentration through beamforming, among others. PLS can be used to provide keyless and innately secure communications, generate and distribute keys for symmetric encryption~\cite{Chorti-BITS}, and provide authentication through localization, fingerprinting and physical unclonable functions~\cite{Pappu02}, for example.\\ 

\vspace{-0.02in}
\noindent \emph{\textbf{{Key Challenge 4:  Further development of Physical Layer Security (PLS) technologies is required in order for PLS to provide competitive alternatives or complementary schemes to standard cryptography; of particular interest are use cases involving limited computation power and memory resources.}}}\\
\vspace{-0.02in}

As an example, in \cite{Location_spoofing_2016_geo_spatial, GilAURO} the angle of arrival has been used to identify Sybil attacks in robotic systems, while in the same direction, range estimation has been used in \cite{Mitev22} to provide resilience against more general impersonation attacks. The availability of positioning information will be provided in future 6G networks as a default service~\cite{hexa-x}, and several protocols have already been proposed that incorporate positioning information as a second soft authentication factor \cite{Nokia11}. The availability and ``standardization'' of location information with 6G also motivates the development of algorithms that utilize this to learn trust of other devices on the network.

However, we currently lack sufficient research into these emerging areas. Research must catch up to this technological opportunity whereby the detection and mitigation of attacks must exploit both the cyber \emph{and} physical sides of cyberphysical systems for the quantification of trust. Success in this endeavor will lead to powerful new characterizations of resilience, and stronger performance guarantees, than if we view problems of security on the communication and networking side separately from the robotics and multi-agent user side. In order to address the key challenges outlined above, we must reach across field lines, and pave the way to a new era of intelligent, and secure, physical networked
systems – or \ul{Trust-Centered Cyberphysical Systems}.\\

%Research must catch up to this technological opportunity however. Currently, many approaches to security focus on either the cyber side, such as with examining transmitted data to detect anomalies \my{\cite{}}, or the physical side, where specialized hardware or authentication keys are employed \my{\cite{}}. Cyberphysical systems provide new opportunities for detecting and mitigating attacks that exploit both the cyber \emph{and} physical sides to quantify trust. Further, the quantification of trust leads to powerful opportunities for characterizing new concepts of resilience. This also allows us to arrive at stronger performance guarantees than if we view problems of security on the communication and networking side separately from the robotics and multi-agent user side.\\

%\vspace{-0.01in}
\noindent \emph{\textbf{Key Challenge 5: Powerful methods of detecting and mitigating the performance impact of attacks in cyberphysical systems requires exploitation of \emph{both} the cyber \emph{and} physical sides of cyberphysical systems.}}\\
\vspace{-0.02in}

%\noindent \emph{\textbf{Claim 6: A quantification of performance impact and \emph{resilience} for multi-agent CPS requires a tight integration between the wireless \textcolor{magenta}{networking instead of communication?} communication side and control side for these systems.}}\\

  \begin{figure}
  \centering
   \vspace{-0.3in}
    \includegraphics[scale=0.27]{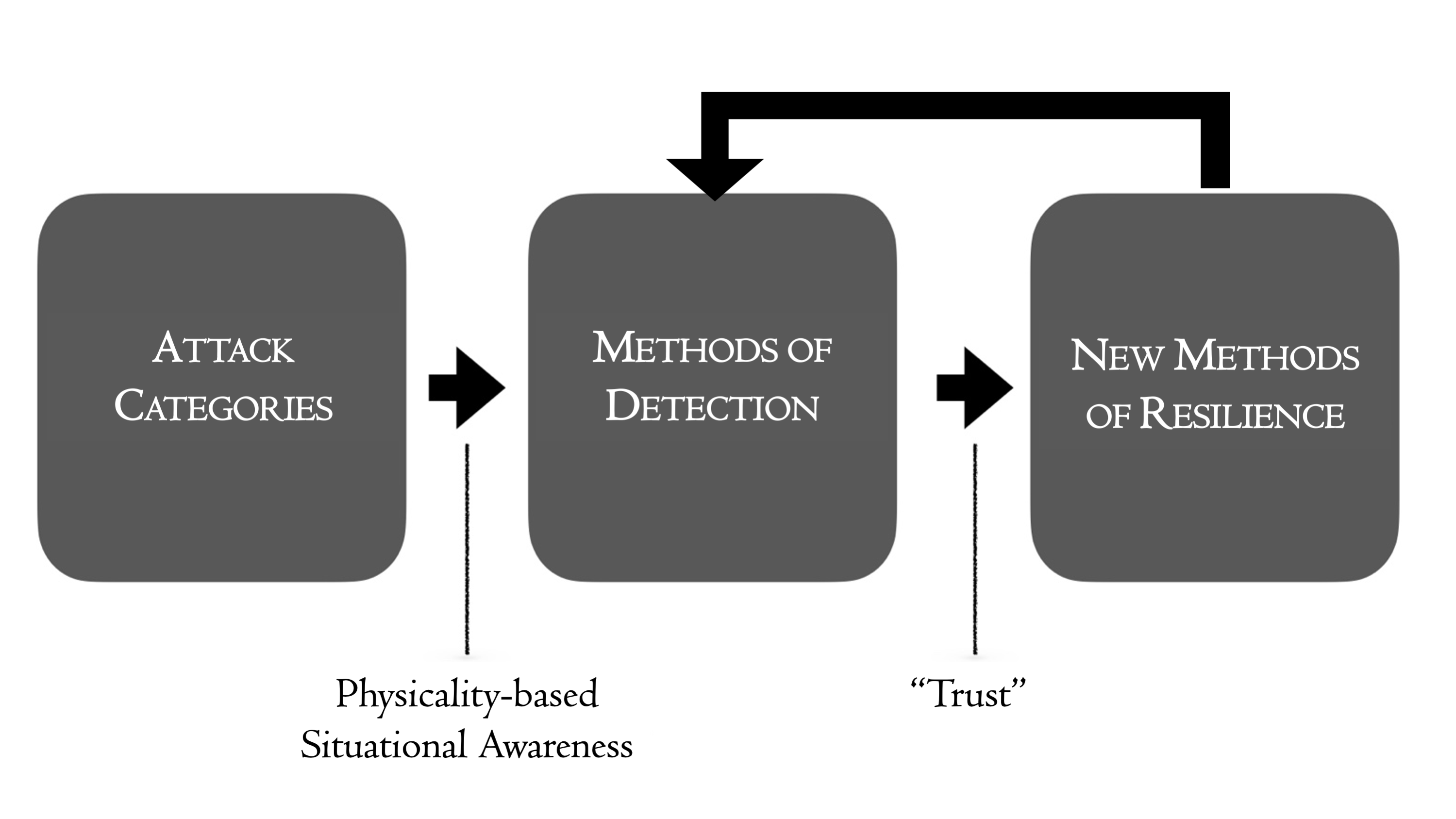}
    \vspace{-0.3in}
  \caption{Survey paper section outline and interconnections. We discuss attack categories in Section \ref{sec:threats}, methods of detection in Section \ref{sec:network}, and new methods of resilience in Section \ref{sec:performance}.}
  \vspace{-0.2in}
  \label{fig:outline}
  \end{figure}
  
In this survey paper we focus on work that has begun to quantify trust and performance impacts in multi-agent cyberphysical systems (subsequently referred to as CPS). We argue that the outlook is very positive in this direction and we identify a need for continued work in this area -- namely, of providing a unified framework all the way from the quantification of trust at the link-level in a network, to the quantification and characterization of performance impact for the multi-agent use-cases of these networks. The objective of this survey article is to highlight the potential of, and current progress in, this direction of research.

\subsection{Related Survey Papers}

The topic of detecting and combating malicious activity in distributed networks has been studied and surveyed \textit{independently} in multiple communities. Notably, detecting adversarial agents based on channel characteristics and general context awareness in the form of radio frequency (RF) inputs, localization, sensing, and hardware and physical layer authentication methods are surveyed in \cite{Xu_survey_Comm_2016,L_Hanzo_survey_2016_a,L_Hanzo_survey_2016_b,Ersi_survey_Comm_Mag_2022,Ersi_survey_arXiv_2022,security_6G_survey_2023}.
The two surveys \cite{Ersi_survey_Comm_Mag_2022,Ersi_survey_arXiv_2022} provide the most recent report on context-aware security that can exploit new sensing capabilities available in the form of image, video, or sound. Additionally, the survey \cite{security_6G_survey_2023} provides the latest update on the security of edge networks in the coming wireless communication networks.
Similarly, the robotic community has investigated these topics too, see for example \cite{Kabiri2022ARO}.
On the other hand, the signal processing and control communities have dedicated multiple  survey papers, such as \cite{Vempaty_SSP_survey_2013,Chen_SSP_survey_2018,Yang_SSP_survey_2020,Sundaram_survey_control_2022,Ishii_control_survey_2022}, to combating adversarial activity in distributed systems based on the  \textit{data} agents share with one another and their knowledge on the connectivity graph. 

Additional related surveys are \cite{Resilient_self_organization_survey_2019,surveyonCyberNetworkSecurity,Prorok_survey_2021,Prorok_survey_2022} which focus on the resilience of self-organization/distributed 
and cyberphysical systems.  
In particular, \cite{Resilient_self_organization_survey_2019} presents the possibility to exploit situational awareness and contextual information regarding the system environment
to enhance resilience to cyber-attacks. Nonetheless, since at the time, the available theory was very limited, on the whole, it was not clear how to acquire and exploit the physical aspect of the problem, including, situational awareness to enhance resilience.  
Similarly, the survey paper \cite{surveyonCyberNetworkSecurity} includes a discussion on the role of the network in cyberphysical systems, but the communication world and physical world are not integrated broadly and seamlessly~\cite{surveyonCyberNetworkSecurity}. Instead, the scope of integration has been mainly limited to system control, coding and encryption, and event detection and communication aspects such as packet switching, routing and denial of service attacks. 

The recent survey papers \cite{Prorok_survey_2021,Prorok_survey_2022} touch upon resilience to attacks utilizing side information regarding the trustworthiness of agents that is acquired using the physical aspect of the network, nonetheless, this is not the main focus of these works. 

This survey showcases the recent advances to bring the theory of distributed control together with the physical world, which provides valuable context. Specifically, it aims to leverage the physically-based context-aware adversarial detection to create resilient distributed systems that are highly robust to adversarial activity. Recent results show that this approach carries great promise and provides superior resilience guarantees and graceful degradation of these guarantees when the number of adversaries in the system increases, for example.

\vspace{-0.1in}
\subsection{Organization of the Survey Paper and Connection to Key Challenges}
% \begin{wrapfigure}[11]{r}{0.65\textwidth}

% \end{wrapfigure}
We divide this survey article into three major interconnected sections. In Section~\ref{sec:threats} we expound upon Key Challenges 1 and 2, where we outline both novel and existing threats that are particularly relevant for future cyberphysical networks. Section~\ref{sec:network} outlines the reasons why traditional notions of security are not enough to address these threats (Key Challenge 1), and makes the case for new opportunities for security brought forth by 6G and future networking technologies, expanding on Key Challenges 4 and 5. These include a range of new authentication methods that rely on the physicality of the network, such as RF fingerprinting and Physically Unclonable Functions, as well as the role of contextual awareness in detecting and identifying attacks. In Section~\ref{sec:performance} we survey novel trust-enabled resilience guarantees that have recently emerged for various networked multi-agent system applications, by exploiting new concepts of security and sensing brought forth by future networks (Key Challenges 3 and 5). Lastly, in Section \ref{sec:open_prob_conclusion}, we discuss the role, and the future potential, of  %using \emph{trust values} 
exploiting and integrating the new sensing and contextual physical information
for arriving at stronger, provable performance guarantees for these systems.

We note that many of the concepts, particularly in Section~\ref{sec:threats} and Section~\ref{sec:performance}, extend broadly to wireless networks. Section~\ref{sec:network} however, focuses mostly on how advances in next generation \emph{cellular networks} will bring new tools to the table that will allow for the measurement and quantification of trust over the network. Sensing over wireless channels has been a topic of intense research over the past decade~\cite{yasamin_throughWall,yasaminOccupancy,swarun_RFsensing,deepak_localization,adib2013see,gilUbicarse,Xiong:Securearray:2013,arrayTrack}, and demonstrates the great promise of endowing cellular networks with similar capabilities. The idea that sensing and contextual awareness capabilities can extend to cellular networks is of great importance since cellular networks can provide infrastructure support for systems operating outside of local wireless networks -- such as autonomous vehicles and delivery drones. As such, we survey the literature for both cellular and general wireless networks in this article.

A major goal in writing this survey article is the merging of ideas and concepts from the several disparate fields of wireless communication, control, and distributed multirobot systems. This goal is motivated by the fact that designing and utilizing future wireless networks in a way that provides resilience to various future applications requires an understanding of concepts across these disparate fields. Thus, our objective is to provide contextualization of this problem within all of the aforementioned fields, while also providing interconnections between them.
%\mynote{TODO: check the section layout and descriptions one last time after we finalize all the other sections.}

\section{Attack Models: New Foes Join Old Foes}
\label{sec:threats}
%\my{TODO: add a short summary for the content and layout of this section.}
%\my{Caution: Note the main difference between Byzantine and malicious agents. In the general case, I call misbehaving agents adversarial agents. Additionally, we should distinguish between decentralized and distributed schemes, or add a remark about the slight abuse of terminology.}

% \begin{wrapfigure}[12]{r}{0.5\textwidth}
\begin{figure}
  \centering
  % \vspace{-0.1in}
    \includegraphics[trim={8cm 2cm 8cm 3cm},clip,scale=0.4]{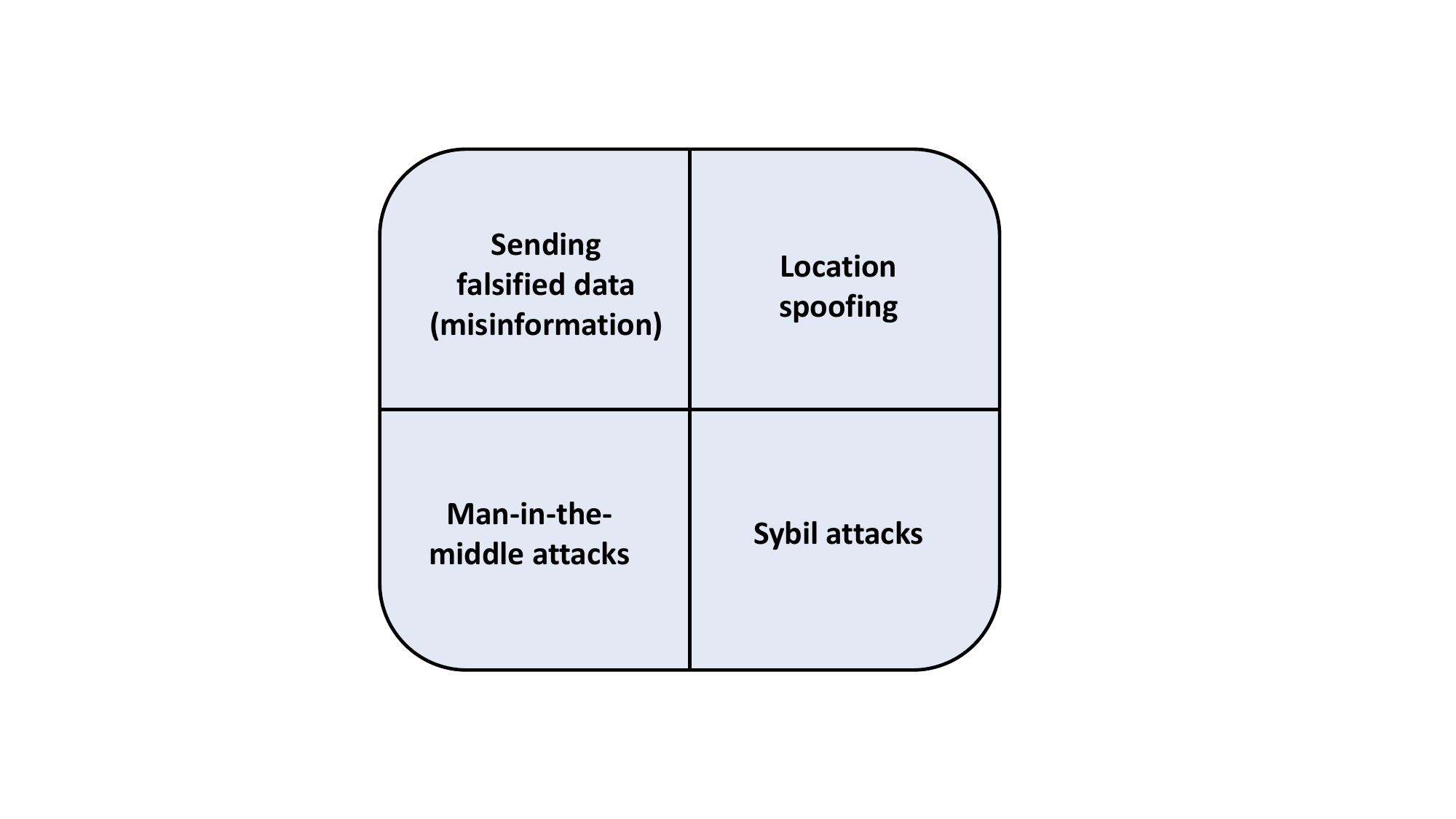}
    % \vspace{-0.1in}
  \caption{We focus on these four major attacks on agents' honest cooperation in conveying their outlook on a common task. As we discuss in this section, these attacks can be integrated to evade detection and increase the impact of the attack on the overall system objective.} 
  \label{fig:threats}
  \end{figure}
% \end{wrapfigure}

Security for networked systems is a large topic with a very rich history of research spanning cryptography~\cite{stinson2018cryptography}, authentication~\cite{Authentication_review_2021}, and specialization of hardware and software~\cite{hardware_software_security_2016,CR_based_high_layer_auth,Survey_PLA_Wireless_2021} among other research areas~\cite{Security_information_centric_2015}. However, as we head into a future where the nature of the networks themselves are evolving to include highly heterogeneous and autonomous agents such as drones and multi-robot systems, we must reconsider our definition of \emph{threats} and characterization of \emph{mitigation strategies}. We must ask, \emph{are threats to multi-robot systems the same as those for traditional networked systems? Or are there new threats and new tools for mitigating their impact that we should consider?} Whereas for traditional networked systems, questions of resilience primarily centered on link failures~\cite{DistOptAngelia,Angelia_time_varying1,Semi_dec_FL_intermittent_2022,cloudClusterTWC}, grid failures~\cite{grid_failure_1,grid_failure_2}, eavesdroppers~\cite{AM_WPLS_2014,VHP_WPLS_2017,YP_WPLS_2021,MH_WPLS_2021}, and jammers~\cite{Jamming_survey_2022,Jamming_survey_2015,Jamming_survey_2013,Jamming_survey_2011,Jamming_survey_2009,Jamming_survey_2008a,Jamming_survey_2008b}, malicious agents in tomorrow's networks will aim to disrupt the functioning of the agents utilizing the network, 
as is recently observed in \cite{surveyonCyberNetworkSecurity}. 

\begin{svblock}{}
This section is  dedicated to exploring attack models that are aimed at disrupting the operation of multi-agent systems that rely on the trusted cooperation of the participating agents in achieving a global task or objective.
\end{svblock}

There are several aspects of emerging cyberphysical networks that make them unique, requiring a re-examination of the question of resilience. These systems are 1) largely distributed, or decentralized, or a hybrid of these two as in a cloud-cluster architecture~\cite{cloudClusterTWC}, 2) highly heterogeneous in terms of resources and requirements, for example with regards to computation and tolerance of latency, and 3) often equipped with sensing, computing, and actuation onboard. This combination of attributes both redefines the possible attack surfaces and vulnerabilities of these systems, and lends new opportunities for identification and mitigation of threats. 

The emergence of new distributed and decentralized networks, and their dependence on truthful and correct agent information,  motivates examining their vulnerabilities to malicious activity more carefully.
Multiple research communities have proposed and developed attack models that target distributed systems, including the communication, signal processing, control, security (computer science), and robotics communities. Each of these communities has made unique contributions to the understanding of the vulnerabilities of distributed systems that rely on inter-agent communication and collaboration.   

Our focus in this section of the survey paper is on threats that are particularly relevant to emerging networks, or embodied cyberphysical systems, i.e. \emph{\textbf{where the physicality of the network plays a prominent role, either in the attack or its mitigation or both}}. 

We categorize relevant threats into the four broad categories of i. \emph{misinformation}, where erroneous information is intentionally disseminated throughout the network to interrupt the agents' ability to collaborate or perform their intended task, ii. \emph{location spoofing}, where agents might misreport information about their state including position and heading, iii. \emph{man-in-the-middle attacks} where an adversarial node intercepts the communication between two legitimate nodes and acts as a concealed relay to eavesdrop, drop messages, alter or forge  communicated messages between the two legitimate nodes,
and iv. \emph{Sybil attacks} where agents emulate other agents in the network. In the literature, agents that are not truthful and/or cooperative are referred to as adversarial, malicious, or Byzantine. These agents intentionally disrupt the system by sending falsified information to their neighboring agents or to a central controller, as we discuss in Section \ref{sec:attack_misinformation}. 
In some systems, such as vehicular networks,  agents must be located within a reported perimeter to participate in the multi-agent collaboration. Thus, if adversarial agents cannot position themselves within their desired reported location, they must lie about their true location. We elaborate on this setup further in Section \ref{sec:attack_location_spoofing}. Distributed networks also rely on the communication channel between agents to be trustworthy, i.e., that the only changes a transmitted signal experiences during transmission stem from the communication channel characteristics, such as attenuation, additive noise, and interference from other transmissions. %Similarly, we assume that the routing mechanisms such as the wireless local area network.
However, in practice, this information-passing process can be useful to carry out attacks called man-in-the-middle attacks 
as we discuss in Section \ref{sec:MIM}.
Finally, another potentially devastating attack on distributed systems is the Sybil attack, where an adversarial agent creates multiple fictitious entities to have a prominent influence on the network beyond what is possible for a single agent. We elaborate on this attack in  Section \ref{sec:attack_sybil}.
We note that in many scenarios, a combination of more than one type of attack can be used by an adversarial agent. We give two examples of such integrated attacks in Section \ref{sec:combined_attacks}.
\begin{figure*}
    \centering
\includegraphics[trim={0cm 1cm 0cm 2cm},clip,scale=0.45]{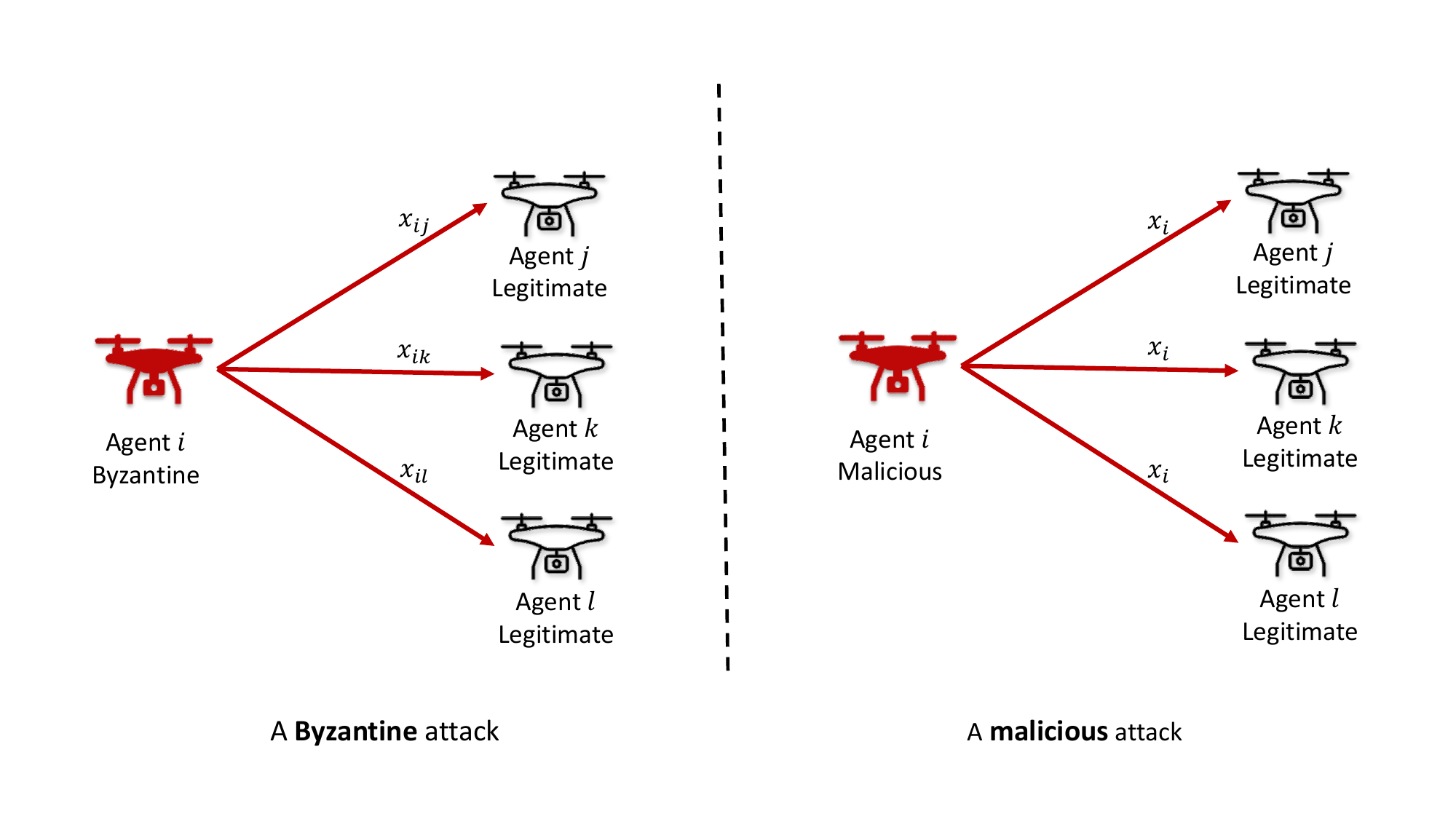}
    \caption{A Byzantine attack (left) vs. a malicious attack (right). In a Byzantine attack an adversarial agent can send different falsified data values to its neighbors whereas in a malicious attack the adversarial can only send identical falsified data values to all of its neighbors, for example via broadcasting. }
    %\my{TODO: revise the figures if we change the terminology. The same goes for the example subsections.}
    \label{fig:Byzantine_vs_malicious}
\end{figure*}

For the purposes of completeness and to compare and contrast our objectives in this survey paper, we include in the following paragraph a description of other research directions in the area of security and resilience for traditional networks that are out of scope for this work.\\

\noindent \textbf{\underline{What this survey is \emph{not} about:}} There are many attacks on a network that can compromise the authenticity of data and/or reliability of the links. In past decades security vulnerabilities such as data leakage and jamming were extensively studied. In a data leakage attack, an unauthorized eavesdropper wiretaps the channel between an authorized transmitter and receiver to learn the information that is sent. Information leakage can be exploited by malicious transmitters, allowing them to devise devastating attacks based on the leaked information.  This reduces the overall trustworthiness of the network.
Nonetheless, information leakage is a passive attack where the eavesdropper does not directly affect the authorized transmitter or receiver. Hence this type of attack is not within the main focus of this survey. We refer the reader to the relevant surveys on this topic in \cite{AM_WPLS_2014,VHP_WPLS_2017,YP_WPLS_2021,MH_WPLS_2021}.  
A jamming attack is a direct attack on the receiver where a jammer transmits \textit{interfering} signals to reduce the signal-to-noise ratio (SINR)
%\an{reduce the signal-to-noise ratio -- AN: isn't it to increase the ratio?} 
at an authorized receiver. Similar to data leakage, a jamming attack generally does not aim to input specific data into the system but rather affects the receiver's performance by reducing its SINR. Combating jamming is a prolific area of research, which develops anti-jamming strategies such waveform design robust to jamming, frequency hopping and spread spectrum, as well as subtraction of the jamming signal. Most jamming attacks cause a denial of service that does not inject harmful data into the system, thus they are not the main focus of this survey; we refer interested readers to multiple surveys that have been written on jamming over the years \cite{Jamming_survey_2022,Jamming_survey_2015,Jamming_survey_2013,Jamming_survey_2011,Jamming_survey_2009,Jamming_survey_2008a,Jamming_survey_2008b}. Nonetheless, a special type of jamming attack, known as \textit{deception jamming}, injects falsified information into the system in order to fool the receiver into believing the validity of the information it receives, see for example \cite{deceptive_jamming_2022,deceptive_jamming_2020,deceptive_jamming_2019,deceptive_jamming_2018,deceptive_jamming_2018_spie,False_data_injection_jamming_2018}. We do not consider such attacks in this survey paper, since our focus is on threats that are particularly relevant to networks where the \emph{physicality} of the system, by way of sensing, actuation, or onboard intelligence, plays a critical role in their attack surface and attack mitigation.

\subsection{Misinformation in Distributed Systems}\label{sec:attack_misinformation}

In this section we discuss a threat to performance in cyberphysical systems that emerges from the presence of invalid and untrustworthy data in the system, either due to unintentional misreporting (e.g. due to faulty sensors) or intentional manipulation of the data by one or multiple agents in the system.  The prevalence and potential impact of this threat is exacerbated by the decentralization or distributed nature of sensing and computation in the network. 

Distributed systems include multiple devices that aim at reaching a common goal using local information. In this survey, we refer to these devices as agents. Each agent is equipped with computing and communication capabilities. In many cases, such as sensor networks, the agents can also sense their surrounding environment.  
Generally, distributed systems can be further categorized into two groups, centralized and decentralized. The centralized settings include a central server, also known as a cloud or a parameter server, which orchestrates the operation of the network by communicating with each of the participating agents. In a decentralized setup this central server does not exist, instead, data flows in the network through peer-to-peer communication, and decisions are made locally. 

\begin{svblock}{Distributed systems vulnerability to misinformation}
Distributed systems depend on the reliable cooperation of the agents in following the distributed protocols. This makes them vulnerable to erroneous information that is either faulty or adversarial.
\end{svblock}

In distributed settings, erroneous inputs are often categorized as faulty or adversarial. Faulty information consists of innocent errors that are caused by equipment or software malfunction. These errors are not intentional, and are not strategically chosen to maximize the negative effect on system behavior or to detection. In contrast, adversarial inputs are strategically chosen by adversarial agents to harm the system's correct operation and prevent it from reaching its designated goal. For example, if the system goal is to reach a consensus in a decentralized system or agree on a minimizing point in distributed optimization,  adversarial agents  will input values that prevent the legitimate agents from reaching a common consensus value~\cite{bulloUnreliable}, or take the agents' values away from their optimal point~\cite{Sundaram_dist_adv_2019}.

\begin{figure*}
    \centering
\includegraphics[trim={0cm 2.5cm 0cm 2cm},clip,scale=0.45]{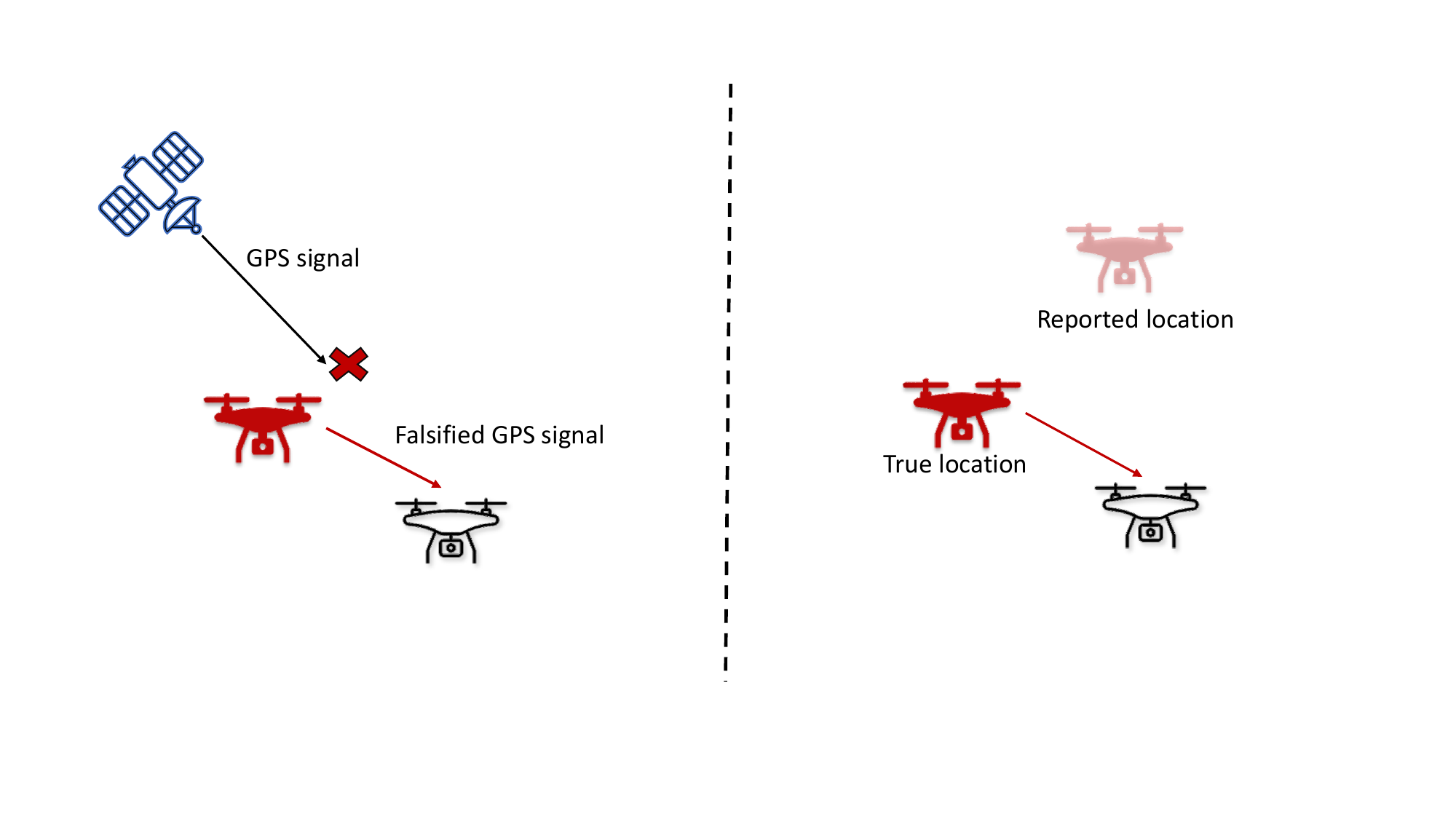}
    \caption{Location spoofing models. On the left figure, the attack is carried out with respect to the location of the \textit{received} agents which the attacker tries to prevent from recovering. On the right figure, the attack is carried out with respect to the location of the \textit{trasmitting} agent which it tries to hide from the receiving agent.}
    \label{fig:location_spoofing}
\end{figure*}

Adversarial inputs are often further categorized as \emph{Byzantine} or \emph{malicious}. We note that the distinction between adversarial, malicious, and Byzantine attacks does not seem to be standardized across different papers or research fields. For example, in the  computer science literature the term ``faulty agents"  may refer to adversarial agents \cite{Lamport_agreement_faults_1980}. Here we adopt the distinction proposed in \cite{Sundaram_Survey_resilience_graph_2022} to categorize erroneous inputs. In this section, we align to this convention and categorize adversarial inputs as depicted in Figure \ref{fig:Byzantine_vs_malicious}. In a Byzantine attack, an adversarial has the ability to send different falsified inputs to different agents, however, a malicious attacker can only falsify its input and broadcast the chosen value to all of its neighboring agents. Under this definition, Byzantine attacks can only occur in decentralized or semi-decentralized settings. Our focus in this survey will primarily be on intentional attacks such as malicious or Byzantine attacks, and less so on faulty agents.

\subsubsection{Byzantine Attacks}
The first works to consider  adversarial agents in decentralized systems are  \cite{Lamport_agreement_faults_1980} and \cite{Lamport_byzantine_general_1982,DOLEV198214} where Byzantine values are fed into a voting system with the main objective of preventing the system's consensus. The work \cite{BFT_arbitrary_values} extends this model to the case where the inputs of the Byzantine agents are arbitrary and not necessarily Boolean or belong to a finite set. This relaxation of the set of possible values Byzantine values can take is also  assumed in papers such as \cite{LeBlanc_byzantine_WMSR}, which are studied in the control theory literature.

\subsubsection{Malicious Attacks}

In malicious attacks, the malicious agents cannot send each of their legitimate neighbors different malicious inputs, instead, each malicious agent chooses a malicious value and then sends it to all of its neighbors, oftentimes referred to as a ``broadcast'' model. The malicious agents may collude in choosing their respective values. 

Malicious and Byzantine agents can choose to falsify or corrupt the  values they share with legitimate agents randomly \cite{False_data_injection_jamming_2018,Data_injection_attack_2019,Data_injections_2021}, add  artificial random noise \cite{Data_Injection_Attacks_2016,Data_Injection_2018,Data_Injection_Attacks_2022,Data_injection_attack_Urbashi_2022}, and introduce strategic bias \cite{Data_Injection_Attacks_2016,Data_Injection_2018,grid_attack_bias_2020,Data_Injection_Attacks_2022,Data_injection_attack_Urbashi_2022}. If they used data-based values, for example as in training ML models distributively, then they can manipulate their model updates directly, or poison their data and afterward follow the agreed protocol \cite{Data_poisoning_NIPS2017}. Thus, the latter case is only relevant to malicious attacks and not Byzantine attacks. All in all, Byzantine and malicious attacks have a critically disruptive effect on the reliability of distributed and decentralized networks, thus considerable efforts have been invested in making these networks more secure, robust, and resilient, as we discuss in Section \ref{sec:performance}.

%\textcolor{red}{$<--$ Can we continue this thread of reasoning? Do we want to say anything further about the impact/influence of this attack on the network?}

% Let us consider a decentralized or a distributed system with an iterative dynamic that yields the following agent values for each agent $i$ $x_{i}(t), t=1,\ldots$. 

% Let $x_m(t)$ be the input of malicious agent $m$ at time $t$, then several attack models have been proposed

% Attacks for distributed consensus and optimization includes 

% \paragraph{Bounded malicious inputs}
% Random corruption of data

% \paragraph{Malicious inputs with constant mean and additive noise}
% \begin{flalign}
%   x_m(t) = c_{\MM}+n_m(t)
% \end{flalign}
% where for each malicious agent $i$, the artificial noise additives $n_m(t), t=1,\ldots$ are \my{IID} random variables with a bounded variance.
% \paragraph{Biasing attack}

% Attacks for ML models:

% Data poisoning and model poisoning attacks

\subsection{Location Spoofing}\label{sec:attack_location_spoofing}

Device location reporting is integrated into many distributed systems such as IoT devices, vehicular networks, and robotic networks that utilize geospatial tagging and location-based services. We refer to \emph{location spoofing} as an attack where the location of an agent is falsified, either by the agent itself or by a third-party intercepting agent, with the intention of hiding or misreporting accurate location information (see Figure~\ref{fig:location_spoofing}).
\begin{svblock}{Distributed systems vulnerability to location spoofing}
    Truthful location acquisition and reporting are crucial components of location-based services. 
\end{svblock} 
Naturally, localization and location reporting are key requirements for navigation and collision avoidance in vehicular and robotic networks. Localization can be acquired by agents using multiple methods such as GPS, cellular antennas, and WiFi. Currently, GPS is the most prevalent positioning technique outdoors, nonetheless, it cannot typically be used indoors. On the other hand, cellular and WiFi localization techniques are effective in indoor scenarios.

In what follows we discuss location spoofing attacks in distributed systems through the general lens of IoT and distributed networks. We then provide an additional discussion about location spookfing in the special case of vehicular, robotic, and drone networks in Section~\ref{sec:threatCaseStudy}.

Location spoofing attacks can be categorized into two groups as depicted in Figure \ref{fig:location_spoofing}. The first attack type is carried out on the receiver's GPS-acquired location where a legitimate agent aims at locating its own position using, for example, GPS, whereas an attacker aims to feed the legitimate agent a falsified location by transmitting a signal that circumvents the true location \cite{GPS_spoofing_2011_requirements,GNSS_spoofing_2016}. The second attack aims to spoof, i.e., falsify the location of the transmitting agent. This can be a devastating attack on location-based services such as navigation, where non-existing traffic congestion can mislead navigation applications by making certain roads appear congested as described in~\cite{spoof_location_vehicula_2014,traffic_management_2015,Security_GPS_INS_location_tracking_2019,Road_traffic+poisoning_navigation_2022,advHypTestingICRA}.    

Multiple works, such as \cite{GPS_spoofing_2011_requirements,Location_spoofing_2013_beamforming}, have considered requirements for successful spoofing attacks.
Specifically, the work \cite{GPS_spoofing_2011_requirements} considers a satellite-lock takeover on devices that aim to recover their location. 
 Additionally, the work \cite{Location_spoofing_2013_beamforming} considers the requirements for location concealing and spoofing by an agent that does not wish to report its true location. The main difference is that in location concealing, the falsified location can be arbitrarily chosen with the objective of hiding the agents' true location, whereas in location spoofing, the falsified localization information is strategically chosen to advance an attackers' goals. 

\subsubsection{IoT}
Location spoofing attacks can target a myriad of malicious applications.
While the most obvious example of targeted networks and applications are vehicular networks and navigation, there are many more targeted location-based services that the general population may not be aware of. This includes the retail sector  
\cite{Location_spoofing_2017_IoT} where customers can be directed to points of interest such as nearby stores, public transportation stations, and tourist attractions based on their location.
Vendors can also utilize users' locations to customize their advertisements for the relevant locations and manage waiting times for services. Location spoofing can also impact healthcare services such as patient monitoring and urgent care, tracking the utilization of healthcare equipment in hospitals, and even pandemic monitoring and prevention \cite{IoT_pandemic_2021}.

\subsubsection{Vehicular, robotics, and drone networks}
\label{sec:threatCaseStudy}
Robotic networks, drone networks, and alarmingly, autonomous driving networks rely on localization for various tasks. Among these tasks are coverage (i.e., for surveillance and monitoring), rendezvous, navigation, collision avoidance, and aerial relay placement for communication purposes~\cite{FBconnectivityLab, GoogleProjectLoon, ProjectLoon_Nature}. While these networks are intended to be part of our future infrastructure, relying on them without protecting public safety from location spoofing can be dangerous owing to their vulnerability to location spoofing as is discussed in many works, including \cite{UAV_cyber_2013,Lidars_illusions_2017,Distance_spoofing_2019,Spooing_vehicular_radar_2021}. Specifically, one of the earlier works which call attention to the threats that attacks on military drone networks can cause by GPS spoofing is \cite{UAV_cyber_2013}.
The works \cite{Lidars_illusions_2017,Distance_spoofing_2019,Spooing_vehicular_radar_2021} show how sensors such as Lidars can be manipulated to misjudge the distance between two autonomous vehicles which can result in collisions. Finally, the works \cite{spoof_location_vehicula_2014,traffic_management_2015,Security_GPS_INS_location_tracking_2019,Road_traffic+poisoning_navigation_2022,moserADSB_Mobicom,jansen2021trustADSB} demonstrate how location spoofing can harm key components of managing vehicular networks such as traffic monitoring.

\subsection{Man-in-the-Middle attacks}\label{sec:MIM}

\begin{figure}
    \centering
\includegraphics[trim={5cm 3cm 5cm 5cm},clip,scale=0.45]{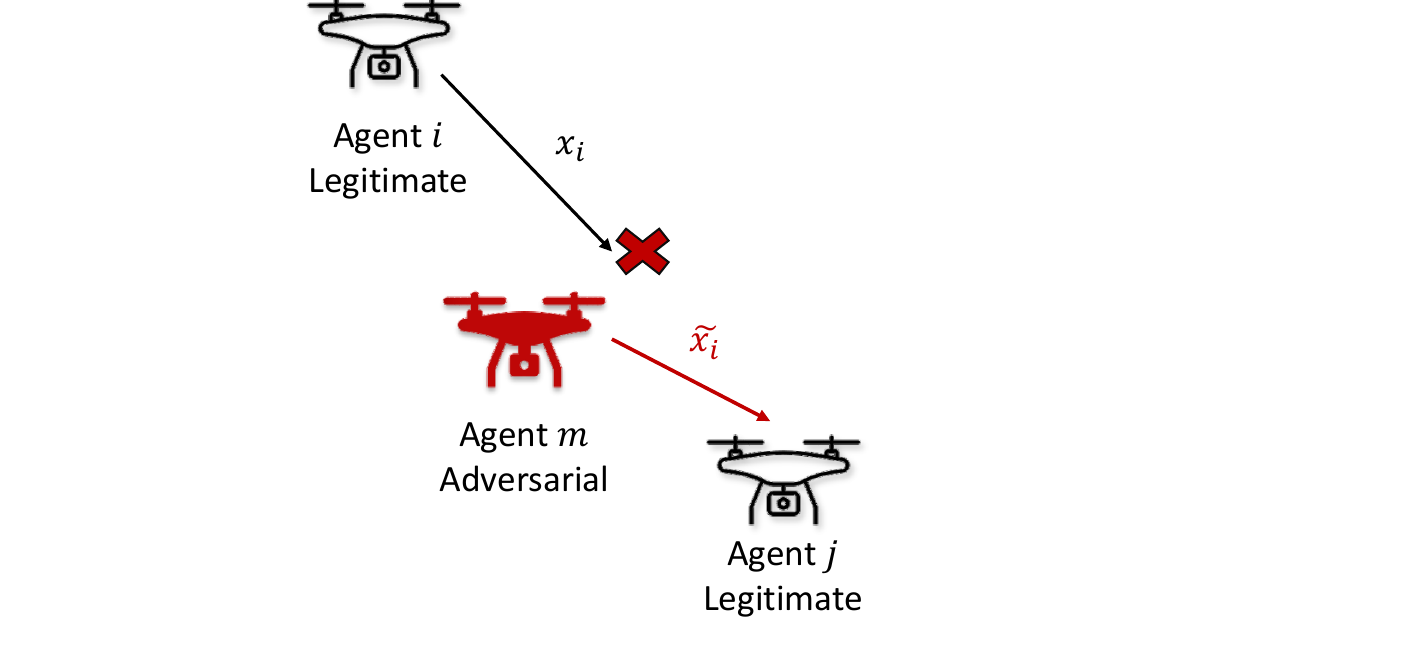}
% trim={<left> <lower> <right> <upper>}
    \caption{A man-in-the-middle attack where a malicious agent relays the transmission $x_i$ of a legitimate agent $i$ intended to agent $j$ without their knowledge. Under the MiM attack, the malicious agent sends agent $j$ the  message $\tilde{x}_i$, instead. This message can be a modification of $x_i$ or a replacement for it. The malicious agent can also choose not to relay the message $x_i$ and cause a denial-of-service.}
    \label{fig:Sybil}
\end{figure}

Distributed systems are also vulnerable to man-in-the-middle (MiM) attacks where an adversarial agent relays the communications between  unsuspecting legitimate end nodes. 
The end nodes, e.g., a transmitter and receiver that want to directly communicate with one another,   are  unaware that instead of communicating directly, the communication between them is covertly done by an adversarial node that relays the messages between them. This adversarial node can eavesdrop on the messages it relays, modify them, replace them completely, or drop some messages altogether. 

\begin{svblock}{Distributed systems vulnerability to MiM attacks}
    Trustworthy and secured communication links are the fundamental   conditions for the ability to utilize distributed optimization systems safely and securely. MiM attacks threaten this endeavor by breaching this underlying requirement for safe and secured communication means to share truthful information reliably.
\end{svblock}

MiM attacks are  well-investigated attacks, consequently, multiple surveys have been written on this subject, including \cite{MiM_Survey_2016,MiM_Survey_WiFi_2022}. Furthermore,  unlicensed spectrum and public links, such as WiFi \cite{Attacks_phone_WIFI2018} and Bluetooth \cite{Attacks_IoT_bluetooth_2018}, are especially vulnerable to MiM attacks  since there is no central entity that can provide a centralized security infrastructure. Having said that, even communication links in the  GSM and the 4G LTE cellular systems can suffer from MiM attacks, as discussed in \cite{MiM_Survey_2016}. These attacks can be carried out in multiple ways as elaborated in \cite{MiM_Survey_2016}. This includes spoofing based-attacks where the MiM attacker intercepts a message from a legitimate node to another and alters it by spoofing the Address Resolution Protocol (ARP) and modifying the desired local ARP cache table of legitimate nodes so data will be routed to and from the adversarial node as if it were the legitimate source or destination node \cite{chen2010proof,yang2012man}.  Similar attack types have been studied for modern vehicles whose components are becoming more computerized (``smart'') and distributed, connected to each other through a central command unit such as the Controller Area Network (CAN). This poses vulnerabilities for intercepting signals from critical vehicle components~\cite{sensorNetworksVehs}.

MiM attacks can also target radar systems as the work \cite{SensingAttack22} suggests. While the methods that are presented in \cite{SensingAttack22} enhance privacy by generating fictitious entities that are to be discovered by radar sensing, these methods can also be adopted by adversaries to fool radar sensing systems by creating manipulating signals to be sensed by legitimate receivers. This can alter key measures of interest such as the number of people in a certain room, their moving patterns, and even their breathing patterns. 

Finally, we note that though at first look location-spoofing of a GPS signal and MiM attacks seem similar, in practice, there is no need to apply the more complicated-to-implement MiM attack to spoof a GPS signal. Instead, the adversarial attacker can send the wrong GPS coordinate to the targeted legitimate node by sending a more powerful signal than the truthful one that is sent by the GPS (see \cite{bhoi2014vehicular}).

\subsection{Sybil Attacks}\label{sec:attack_sybil}

In distributed systems, local data is fused to reach or optimize a global objective, either through the help of a central controller or decentrally.  In honest setups, i.e., when all agents truthfully follow the predefined protocol, the larger the number of agents with similar inputs, the more likely the resulting global system's objective aligns with their outlook. This is clearly evident in a voting system that, given the binary opinion of agents, chooses the one in the majority\cite{thomas1979majority,garcia1985assign}. As a result, an attacker can gain significant influence on the overall system behavior by creating many fictitious identities which often report similar measurements and opinions. This attack is known as the Sybil attack \cite{Sybil_attack_2002}.

\begin{svblock}{Sybil attacks for distributed systems}
    In distributed systems, a potentially devastating form of attack, called a Sybil attack \cite{Sybil_attack_2002}, has been devised where an adversarial creates many fictitious identities to gain substantial effect on the overall system's behavior.
\end{svblock}

\begin{figure}
    \centering
\includegraphics[trim={5cm 7cm 5cm 5cm},clip,scale=0.45]{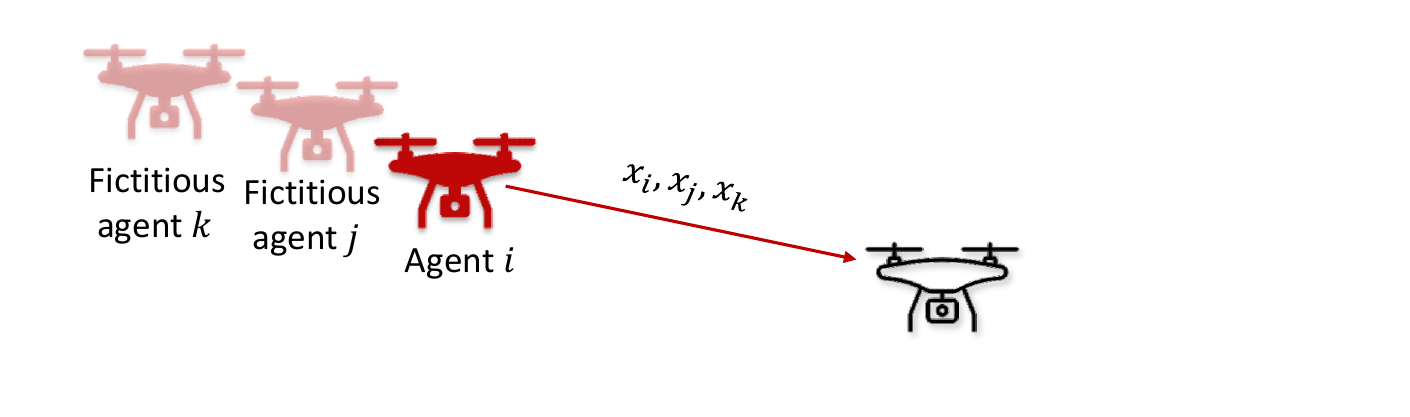}
% trim={<left> <lower> <right> <upper>}
    \caption{A Sybil attack where multiple fictitious adversarial identities are created or stolen to hijack a distributed system.}
    \label{fig:Sybil}
\end{figure}

Sybil attacks can devastatingly affect many types of distributed systems, from general sensors and IoT networks \cite{Sybil_sensors_2004,Sybil_IoT_2014,Sybil_forwarding_routing,Security_IoT_Sybil,low_power_lossy_IoT,Sybil_Analytical_IoT}, mapping \cite{Sybil_mapping_2018} 
\cite{Sybil_Google_maps_2022}, vehicular networks \cite{Sybil_Vanet_2022,Sybil_Vanet_book
} robotic systems \cite{GilAURO} to social networks \cite{Sybil_social}.

The most direct form of a Sybil attack is to inject false opinions or measurements to neighboring legitimate users, or a central server, in order to conceal the true state of the measured phenomenon and overall optimal system objective. This could be the estimated state of congestion on a road, recommendations in reputation systems, detected hypothesis, and others as discussed in the works~\cite{Sybil_sensors_2004,Security_IoT_Sybil,Sybil_mapping_2018,Sybil_Google_maps_2022,Sybil_Vanet_book,GilAURO,Sybil_social,Sybil_Vanet_2022}. 
A Sybil attack can also be carried out in somewhat indirect yet effective ways as described in \cite{Sybil_forwarding_routing,low_power_lossy_IoT}. In particular in \cite{Sybil_forwarding_routing}  Sybil nodes drop packages intended for legitimate users, thus causing denial-of-service errors. Furthermore, Sybil attacks can also be used   to increase the chance of eavesdropping on legitimate messages  via routing in ad-hoc decentralized networks.
%where the routing protocol adapts to the network's participating nodes and topology. %\sg{[SG: I'm not sure I understood the last example. Is there a relevant citation that we can include here?]}  \my{Thanks! I'll clarify this point.}
Additionally, in \cite{low_power_lossy_IoT} the Sybil attack causes another type of denial of service error which is inflicted by causing legitimate users to restart the routing protocol frequently to drain their batteries.

\subsection{Integrated Attacks}\label{sec:combined_attacks}

The attacks we describe above can be combined to create an effective and devastating multi-faceted attack. Next, we give two examples of combined attacks. The first attack targets navigation systems, whereas the second attack targets coverage deployments in robotic systems.

\subsubsection{Navigation Systems as a Case Study}

To showcase the harmful effect of misinformation, location spoofing, and Sybil attacks on vital everyday application services, we now discuss how such attacks on navigation systems that the general population heavily rely on might be designed. In particular, we show how the three attack types we discuss in this section complement one another to create a powerful attack on navigation systems. To this end, we further examine the reliance on navigation systems such as Google Maps or Waze on democratic reporting by users who agree to share with the application their location and can report special events such as traffic accidents. To distort the navigation application a \textit{Sybil} attack \cite{Sybil_Google_maps_2022,Sybil_Vanet_book,Sybil_Vanet_2022} can be effectively used by creating many fictitious identities which can \textit{spoof} the location of vehicles \cite{Location_spoofing_2017_navigation,GPS_spoofing_book}
 that the users  share with the navigation application and \textit{misinform} the application about  events such as traffic jams or a newly found road using the Sybil nodes. Thus, targeting the desired malfunctioning behavior of the attacked section of the network such as blocking small roads to avoid fictitious traffic jams, and entering dangerous unpaved roads.
 
\subsubsection{Integrated Attacks on Coverage in Robotic Systems}
\begin{figure}
    \centering
    \includegraphics[scale=0.28]{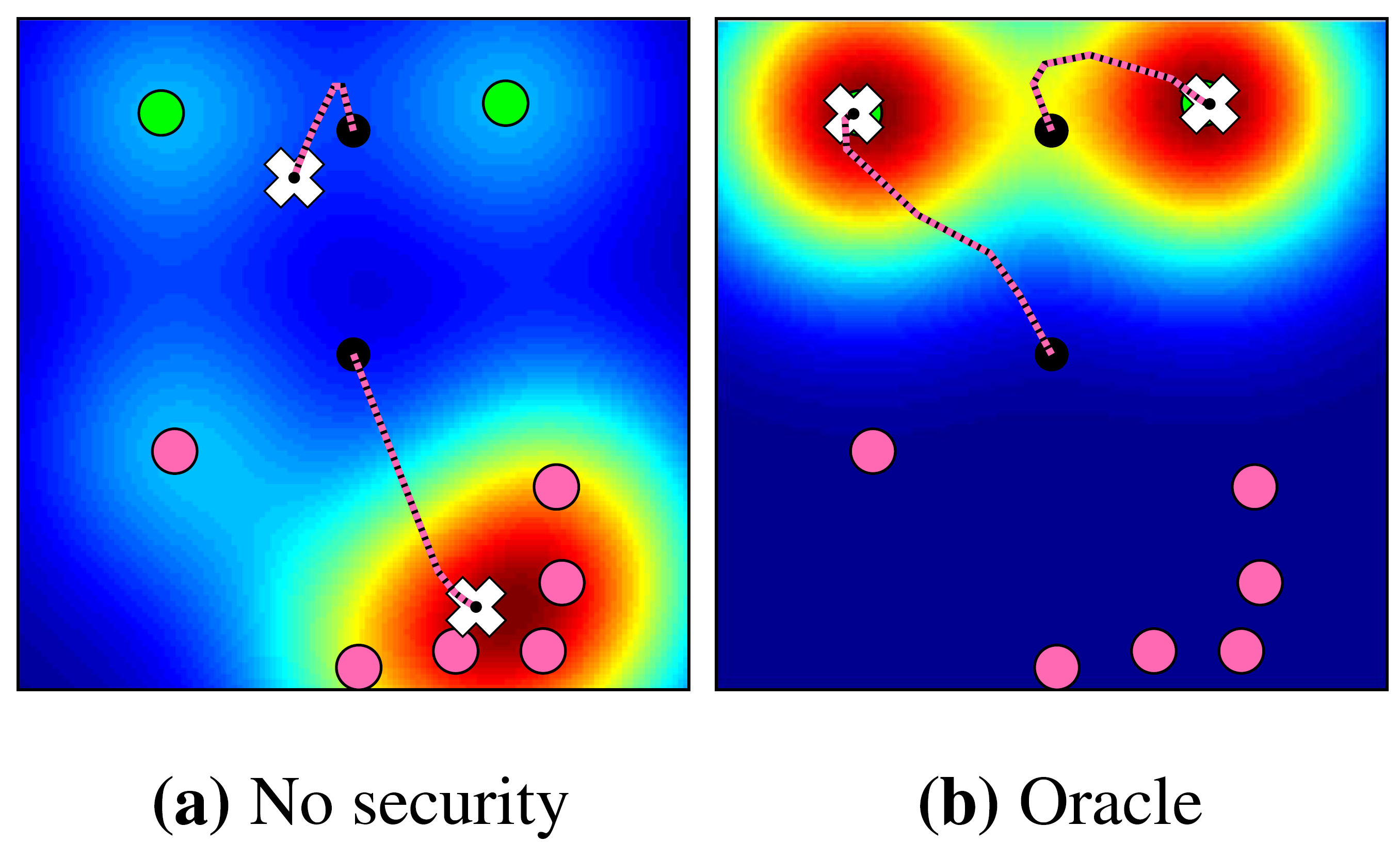}
    \caption{A Sybil attack on  the optimal positioning of two aerial servers \cite{GilAURO}. The two black points depict the starting positions of the two aerial servers before optimizing their location with respect to the deployed robots \cite{bulloCoverage}. The two  white $x$'s denotes the resulting locations of the aerial servers. The green circles denote legitimate robots and the red circles denote the robot identities created by the Sybil attack. Finally, subfigure (a) shows how the attacking robots cause one of the aerial servers to move into their vicinity instead of serving only the legitimate robots as depicted in subfigure (b) where the aerial servers ignore the attacking robots.}  
    \label{fig:GilAURO_coverage}
\end{figure}

An additional integrated attack aims at disrupting the optimal positioning   of aerial servers, aka drones, that are deployed to  provide communication and coordination of robots. To this end, the aerial servers are deployed to minimize their distance to client robots at their reported positions \cite{bulloCoverage,GilAURO}. The work \cite{GilAURO} considers an  attack that integrates both a Sybil attack and location spoofing. This is imperative to deploy a Sybil attack since all robots (legitimate or fictitious) must report a location. Furthermore, if the aerial servers also act as a fusion center that receives sensed information from the deployed robots, the adversarial robots can also send falsified sensed measurements. In Figure \ref{fig:GilAURO_coverage}, which is originally appeared in \cite{GilAURO}, we can see the effect of the Sybil attack on the deployment of two aerial servers. The figure on the left, i.e., Figure \ref{fig:GilAURO_coverage}(a), exhibits the distorted positions of the aerial servers, whereas the figure on the right, i.e., Figure \ref{fig:GilAURO_coverage}(b), depicts the optimal positioning of the aerial server that ignores the existence of adversarial robots. 
Finally, we further refer the reader to Section \ref{sec:rputation_based_application} Paragraph d) where we elaborate on the coverage attack and how to exploit the physical aspects of the problem, which we discuss in the next section, to mitigate it.

%\section{Trustworthiness and Resilience of Future Networks: the Role of the Physical Layer \textcolor{red}{Ersi: the text needs to be cleaned from repetitions}}
\section{Trustworthiness and Resilience of Future Networks: the Role of the Physical Layer}
\label{sec:network}

%\mynote{The title doesn't fully match the content of the section since we also discuss higher layers and not only the PHY. We also don't focus here on resilience.}

\begin{figure*}[t!]
    \centering
\includegraphics[scale=0.425]{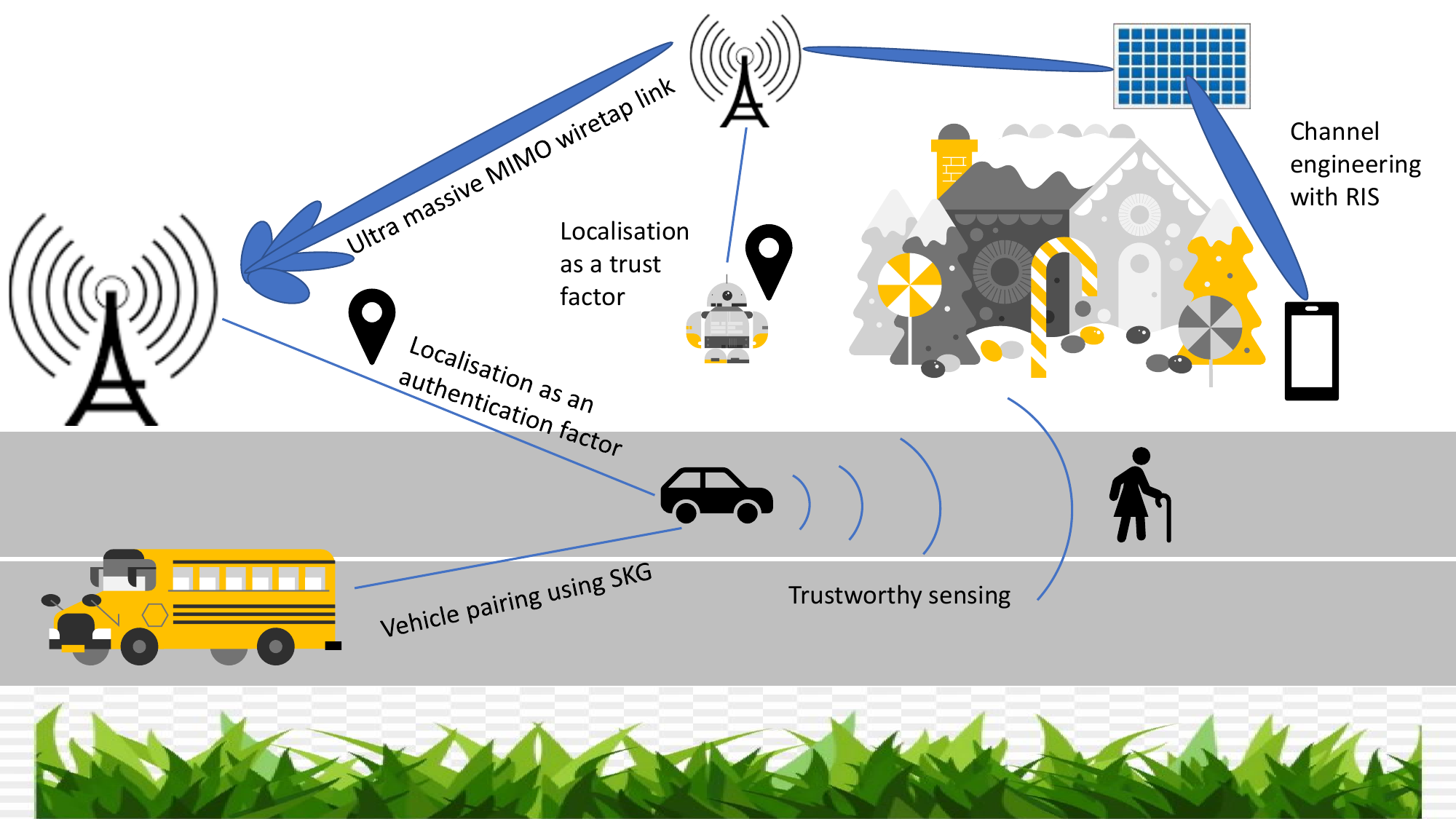}
    \caption{A broader look at trustworthiness. }
    \label{fig:broader_trustworthiness}
\end{figure*}

The anticipated fusion of the physical and cyber worlds marks the beginning of a new era in which the physical properties of interconnected systems contribute to trust building \cite{Nokia}. In this setting, which we capture in Fig.~\ref{fig:broader_trustworthiness}, the overall behavior of the devices, agents, and systems involved, as well as semantics and context, come into play in building trust and evaluating trustworthiness. This is true whether the network connecting these systems is a cellular network or a general wireless network. In this section we will focus heavily on advancements in sensing over cellular networks and expected features in future networks such as 6G. We note in passing that many of these concepts apply to general wireless networks as well. Cellular networks provide exciting opportunities to bring sensing more ubiquitously across devices, for example drones or autonomous vehicles operating in urban environments; thus, we focus mostly on cellular network advances in this section.

Cellular communication is carried out under technological standards that regulate how communication is performed. 
Over time technological advances have been developed to allow high communication rates, reduced power consumption, security, etc. Consequently, cellular communication standards have also adapted over time to take in these technological advances. The evolution of celullar communication is marked by its generation, namely, 1G, 2G, 3G, 4G and the current 5G, where each generation introduced a new improvement with respect to the older ones \cite{Security_5G_Beyond_2019}. In terms of security, the two prominent evolutions have been part of the 3G and 5G designs. 3G introduced a new security method with respect to 2G that allowed users to authenticate the  network they establish a connection with to prevent its impersonation. 5G has had a wider security goal and aimed to establish \textit{``Secure by Design"} principles \cite{5G_secure_by_design,5Gsecurity_survey}. To that end, it aims to provide  mutual authentication between the sender and receiver that does not rely on extrinsic security measures by removing the outdated underlying assumption of additional safety mechanisms such as physical link security (i.e., no tapping), and hardware or application security. Unarguably, the security protocols of 5G systems are  a significant improvement with respect to 4G. However, in 6G these advances will not suffice \cite{security_6G_survey_2023}. To name but a few open issues, the overly aggressive latency constraints can be a limiting factor for public key encryption-based authentication handshakes in machine-to-machine type communication requiring fast communication, i.e.,  high communication rates. For example, those involving side links between autonomous vehicles, drones and robots. 6G trustworthiness is defined by the Next G Alliance as ``confidence in the ability of the 6G system to
perform as expected in the face of environmental disturbances, impairments, errors, faults, and
attacks'', while resilience refers to ``the network’s ability to meet a diverse set of service objectives and to be able to identify, anticipate, detect, and respond to the evolution of state of the network'', which means that the network should be both robust and adaptive \cite{NextGAlliance}. 

 Thus, it is essential to consider the resilience and trustworthiness of future networks, and in order to achieve this we must view this problem holistically, involving all layers of the communication network, starting from the physical layer and moving up. From this perspective, we can posit new methods of security and resilience as complementary or even alternative measures to traditional methods.

To initiate our discussion we first describe the \emph{physical layer} (PHY) and both its traditional and future role in wireless networking. To gain universality, the communication process is abstracted into layers with dedicated roles \cite{Zimmermann80}. At the base is the PHY which is responsible for the transmission and reception of electromagnetic and  optical signals. In addition, sensing capabilities will be integrated with communication signals in the near future, to enable verticals like autonomous driving and smart manufacturing. The PHY is  responsible for encoding the intended messages into codewords, subsequently converted (i.e., modulated) to transmitted electromagnetic or optical signals; at the receiver side, the inverse operations (i.e., demodulation and subsequent decoding) on the received signals take place.
% In a minor abuse of terminology, in this work we  refer as PHY to both the hardware, the signals and the propagation medium, as the interface between the physical world and the communication device.
Above the PHY are multiple layers such as the network and presentation layers that are responsible for tasks such as compression/decompression,  cryptography-based encryption/decryption, and routing.

In this section, we pay particular attention to developments involving the role of the PHY of the communication process -- which is poised to play a critical part in improving resilience and trustworthiness of future networks. First, however, we explain the challenges in providing security for emerging cyberphysical systems and why traditional methods of security might not be enough.

\subsection{Trustworthiness for Cyberphysical Systems and why Traditional Security Methods are not Enough}

In this section we identify \emph{new and necessary concepts of trust} for future cyberphysical systems, and the limitations of traditional methods of network security in providing this trust. 

First focusing on the device level, trust of the  sensing,  communication links, and  processing, needs to be ensured. At a very abstract level, we need to be able to trust: 1) The communication links that carry the data exchanged and authenticate agents and devices, providing confidentiality, integrity, authentication and availability guarantees \cite{TrustBeamforming};
2) The sensing, such as radar, RF, camera, and lidar, that collects raw or processed data and  drives actuation \cite{TrustSensing}. Of particular importance in 6G is high precision localization which will be discussed further in this section;
3) The computing and processing platforms for learning and optimization at different parts of the network such as on-device, edge, and core network~\cite{trustComputing};
4) The Artificial Intelligence (AI), optimization, and control algorithms that determine the behavior of the autonomous agents, devices, and systems based on the received data and sensing inputs \cite{TrustAI}. 
 5) The virtualization of the network with respect to communication and computation, which are transparent to users and critical switches in the network. And finally, 6) in many safety-critical applications we must be able to trust the underlying embedded systems, which rely on software-hardware integration designed to achieve targeted tasks. 
 
 We discuss each of these in more detail below, emphasizing their unique challenges from a networking perspective.

\subsubsection{Trustworthy sensing} Additional challenges arise in the trustworthiness of the sensing layer with respect to how to  render it resilient to unsolicited sensing, denial of service and MiM attacks. %The interplay between privacy and \textcolor{red}{spoofing (in the attack section we used the terminology misinformation for injecting false data. Spoofing was used for lying about identity/location.)} of sensed data resonates with rate -- distortion \textcolor{red}{why? also, should we explain what is rate--distortion? Stephanie, do you think the robotic community would know this term?} approaches to privacy. 
This is likely to emerge as an important research topic, given that radar sensing in particular has already been incorporated in  numerous privacy-sensitive applications, such as smart home devices (e.g., by Google, Amazon, etc.), and has been embraced by automotive and chip-set industries. 

\subsubsection{Trustworthy computing} 
%\textcolor{red}{what do we mean by trustworthy computation? Is it computation or estimation? Ersi: it refers to trustworthy execution platform}
Trustworthy computing is achieved by developing technologies to make computing more secure and private through  hardware enhancements and associated software modifications -- a key aspect has to do with decentralization. 
So far the main focus of security in these systems has been dedicated to avoiding a central point of failure by developing distributed systems, and to preserving privacy and overcoming leakage of user information. Among the distributed technologies currently explored to achieve trustworthy computation are
distributed ledger technologies \cite{CarbonNeutralDL}, federated learning \cite{IEEE_FL}, crowd-sourcing \cite{crowd-sourcing}, private computation \cite{PrivateComputation}, and private information retrieval \cite{PrivateIR}. 
From a hardware standpoint, composability and isolation are key features for a trustworthy hardware layer \cite{TrustworthyHardware}, achieved through a modular approach that enhances scalability as well. As an example, unlike traditional multiprocessor architectures in which components are not isolated from each other and therefore can communicate with all other components, utilize input/output interfaces, and access each memory, the design of composable, microkernel-based operating systems with a tile-based hardware architecture and communication control have appeared in recent years \cite{Sebastian22, Stefan22}.

\subsubsection{Trustworthy AI}
While it has been proposed to use deep learning for anomaly and intruder detection as a security measure, machine learning algorithms themselves can be a target for security attacks.
Significant research efforts have focused on the trustworthiness of AI algorithms, including interpretability, explainability and fairness (unbiased AI) \cite{Thesis2015}. To this end, there is a vast literature on the use of coalitional game theory (Shapley values) or other explainable AI (XAI) approaches \cite{Belle21}, 
interpretability approaches \cite{ExplainableAI}, and the use of evidence theory \cite{NIST21}. At the same time, reputation based and crowd vetting approaches have been widely investigated towards this end, see for example, \cite{STAR}, \cite{TRO_crowdvetting}.  
Nevertheless, there is a fundamental need to connect this vast literature with the behavior of the networked agents and their context in trust building, thereby closing the loop between network and control. To this end we motivate the discussion of the role of context in building trust, and demonstrate how it can be utilized in closing the security loop between network and control in Section~\ref{sec:performance} of this survey paper.

\subsubsection{Network virtualization}
Security is also imperative for trustworthy network virtualization where the computation and communication resources are not centrally available but are spread in the network.
For security, managing and accessing the network resources are done in a transparent way with respect to the users who can only access them virtually,  without having full knowledge regarding their physical location, resources, etc. 6G will interconnect heterogeneous systems by exploiting virtualization -- in the form of network function virtualization (NFV), software-defined networking (SDN) \cite{NVF}, and open radio access network (ORAN) \cite{ORAN}, among others. In more detail, 6G mobile networks will allow  running distributed multi-tenancy workloads with stringent latency, throughput, and energy footprint requirements. Generally,  there has been a lot of attention in recent year to exploit deep learning methods to detect anomalies  \cite{Shone18}, \cite{Gupta18}. Furthermore, specific approaches for distributed anomaly detection in SDN wireless sensor networks~\cite{Nunez20,Nunez22} have also emerged, demonstrating that it is possible in large-scale IoT networks to identify compromised or faulty sensors by monitoring hardware behavior (memory usage, power consumption, Tx / Rx times, etc.).

\subsubsection{Embedded systems} A related aspect that concerns fault injection attacks represents concrete threats to embedded  systems, which are computer systems that are designed to perform specific tasks reliably. Since there are used in many  safety-critical systems implemented, for example, in the aerospace industry, it is crucial to secure them against attacks. As opposed to non-malicious fault detection \cite{FaulteDetection}, a fault injection attack can leverage faults inside the microarchitecture to compromise security \cite{FaultInjection}. For example, an attacker can inject errors during the execution of a cryptographic algorithm and exploit the faulty results to extract information about the secret key in embedded systems \cite{FaultInjectionAES}. This closes the loop back to trustworthy computing,  and execution platforms, resistant to fault injection \cite{AES_Fault}.

% First, we survey in subsection \ref{subsec:Authentication_Trsust_Building} upper and lower-layers authentication methods for trust building in wireless networks. \mynote{Then, we further examine in Section XX  the current and future role of the PHY in trust building.} \my{ Finally, we discuss, in Section XX, the current and future 6G efforts to build a broader and systematic outlook of trustworthiness with robotic network support.}

%A broader look at 6G trustworthiness is given in subsection XX.
%We then focus on two new networks that are utilized and serve AI models and interconnect multiple agents, namely, the sixth generation (6G) of mobile network, and robotics networks. 
%\textcolor{red}{Ersi: robotic networks are part of 6G so I don't think it is necessary to treat them separately.}

%\my{In recent years, enhancing security has been brought forward by the 6G of mobile network standards, which provide guidelines for hardware and service providers to comply with. This motivated many research works to look for new ways to ensure the security of mobile networks and improve existing ones. In particular, 6G will be the first \textcolor{red}{AI native generation}, i.e., multiple operation will rely on the use of AI including for reliable transmission and localization, sensing, etc.. 6G will interconnect intelligent and autonomous cyberphysical systems (robots, vehicles, platoons), will bring to life digital twins of physical objects and the metaverse. }

%\my{TODO: add a short introduction for robotic/autonomous networks - Subsection D.}

After having outlined several new and unique challenges on the horizon for current and emerging networks, we give special consideration to widespread traditional methods of providing trust and security in networks; \emph{authentication}, and its future challenges and drawbacks.

\subsection{Authentication Alone is not Enough for Trust Building }\label{subsec:Authentication_Trsust_Building}
A cornerstone operation in trust building is authentication, which drives access control and permissions. In the literature, authentication can refer to either: i) message integrity authentication that can be used to verify that non-modification has occurred to the data exchanged (e.g., provided using digital signatures); %Digital signatures or message authentication codes are used most often to provide message authentication, while key-agreement techniques might also be used to provide this service.
or, ii) source authentication to verify the identity that created the information, such as the user or device; here we focus on the latter. Key approaches for source authentication include tag-based methods, and challenge-response (CR) based approaches that ensure users prove their legitimacy by requiring them to successfully complete one or a series of challenges. 
%\end{itemize}
In addition, blockchain based approaches have also been proposed for identity management. 

In the following, we first examine user authentication and device authentication as a means of trust building and briefly review current approaches, revealing their limitations with respect to trustworthy authentication of autonomous cyberphysical systems and deployment in large-scale low-end IoT. After a brief review of existing approaches, we move to discussing alternative, recently proposed solutions, that exploit the physical properties of either the communication channel or the device hardware itself. 

\subsubsection{Higher-layer authentication methods and their limitations}
As previously mentioned, two key approaches are commonly used for  authentication, exploiting either tags that are used to annotate messages, e.g., digital certificates signed with the private key of a certificate authority (CA) and verified with the CA's public key, or, CR-based identification, most commonly using passwords \cite{PKI}. 

\textit{\textbf{Tag-based authentication approaches -- }} These approaches (see for example \cite{tag_based_high_layer_auth}) typically leverage public key encryption and are suitable for high-end systems with more computational capabilities such as autonomous vehicles, and some unmanned aerial vehicles. Whilst the National Institute of Standards and Technology (NIST) has recently approved the first post-quantum algorithms to replace standard public key encryption (PKE), namely, the CRYSTALS-Kyber Algorithm \cite{CRYSTALS_Kyber_2018}, there are still open issues that need to be addressed with respect to %Most importantly,  substantial key sizes (1024 bits) are required to reach a security level equivalent to AES-256, which in turn raises concerns regarding the latency and the computational load induced. Furthermore, there are scalability issues that might be encountered in the deployment in a massive number of devices. 
large scale or public key infrastructureless deployment. 
Two prominent examples of large scale infrastructureless deployments concern smart-cities, namely: 1) massive numbers of low-end IoT devices (e.g., environmental sensors) requiring authentication, and 2) device-to-device (D2D) sidelinks in vehicle-to-everything (V2X) networks operating outside the base station coverage. 

Moreover, 
%we note that 
autonomous cyberphysical systems such as autonomous vehicles will act without a human-in-the-loop; as a result, their {trustworthy} authentication cannot be holistically encapsulated in the guarantees provided by authentication handshakes, irrespective of the technologies used. Behavioral aspects of such agents need to be taken into account and new authentication methods need to be leveraged, for example, based on positioning \cite{ChortiOJVT}, or more generally, on physical layer approaches \cite{Paul08}.

\begin{svblock}{}
Future trustworthy authentication will necessarily build on trust in the algorithms that control and generate the agent's behavior. Although a device might be successfully authenticated, when it comes to autonomous operation this alone does not guarantee trustworthiness. 
\end{svblock}

%\mynote{We give examples for such trust-building algorithms in Section XX.}

%\textcolor{red}{Before exploring this issue in further detail, we first highlight a number of other limiting issues that also arise in terms of communication KPIs of autonomous cyberphysical systems, focusing on latency, complexity and scalability.}

%\subsubsubsection{Tag-based higher-layer authentication approaches}

\textit{\textbf{Challenge-Response-based higher-layer authentication approaches --}} 
These typically consist of two stages, namely an enrollment stage where  a user sends a request to log into a system, and a release (authentication) stage where the system challenges the user to authenticate itself and lets it log in if the user completes the challenge successfully.
Higher-layer CR-based protocols (see \cite{CR_based_high_layer_auth}) have become in recent years an important mechanism for the authentication of low-end devices with limited computational power, such as wireless sensors or radio-frequency identification (RFID) tags. Typically, CR authentication requires the use of an authenticated encryption algorithm and relies on encrypting a message including the device ID, timestamp and other information with a hash of a shared secret (e.g., password or biometrics). 

There are two key weaknesses with respect to such approaches. First, keys are often chosen to have low-entropy, to facilitate memorization. More broadly, they can be stolen, guessed, or, even shared by the users themselves. Secondly, in terms of computational complexity, even the baseline operations of encryption and decryption can be difficult to implement on low-end IoT devices with constrained power supplies and die size. Although many other approaches based on alternative algorithms such as \cite{LPN} have been  proposed,  two key issues remain: (i) devices need to be able to securely store secrets (passwords); and (ii) they should be able to do so  even if they have a minimum computational and power supply. 

\textit{\textbf{Blockchain-based authentication approaches -- }} These rely on a distributed ledger system that allows for transactions among distributed agents without the need for a trusted central authority \cite{Blockchain_survey_2019,Blockchain_based_security_2021}. It is a natural technology for distributed machine learning systems for which the privacy of local data needs to be preserved and in which mutual trust has not necessarily been established.  

Thus, blockchain can play an important role in distributed settings. Indeed it has been considered, for example, as a mechanism for serverless federated learning \cite{Li22}.  Nonetheless, blockchain does have disadvantages, and paramount among these in the context of wireless networks are high energy consumption and long latencies.  While generally secure, it is also susceptible to some types of attacks, including Sybil and denial-of-service attacks \cite{Sybil_attack_2002,Jamming_survey_2008b}. %\textcolor{red}{Ersi: shall we mention that blockchain is energy hungry? }

\begin{figure*}
    \centering
\includegraphics[trim={1cm 2.25cm 1cm 4cm},clip,scale=0.45]{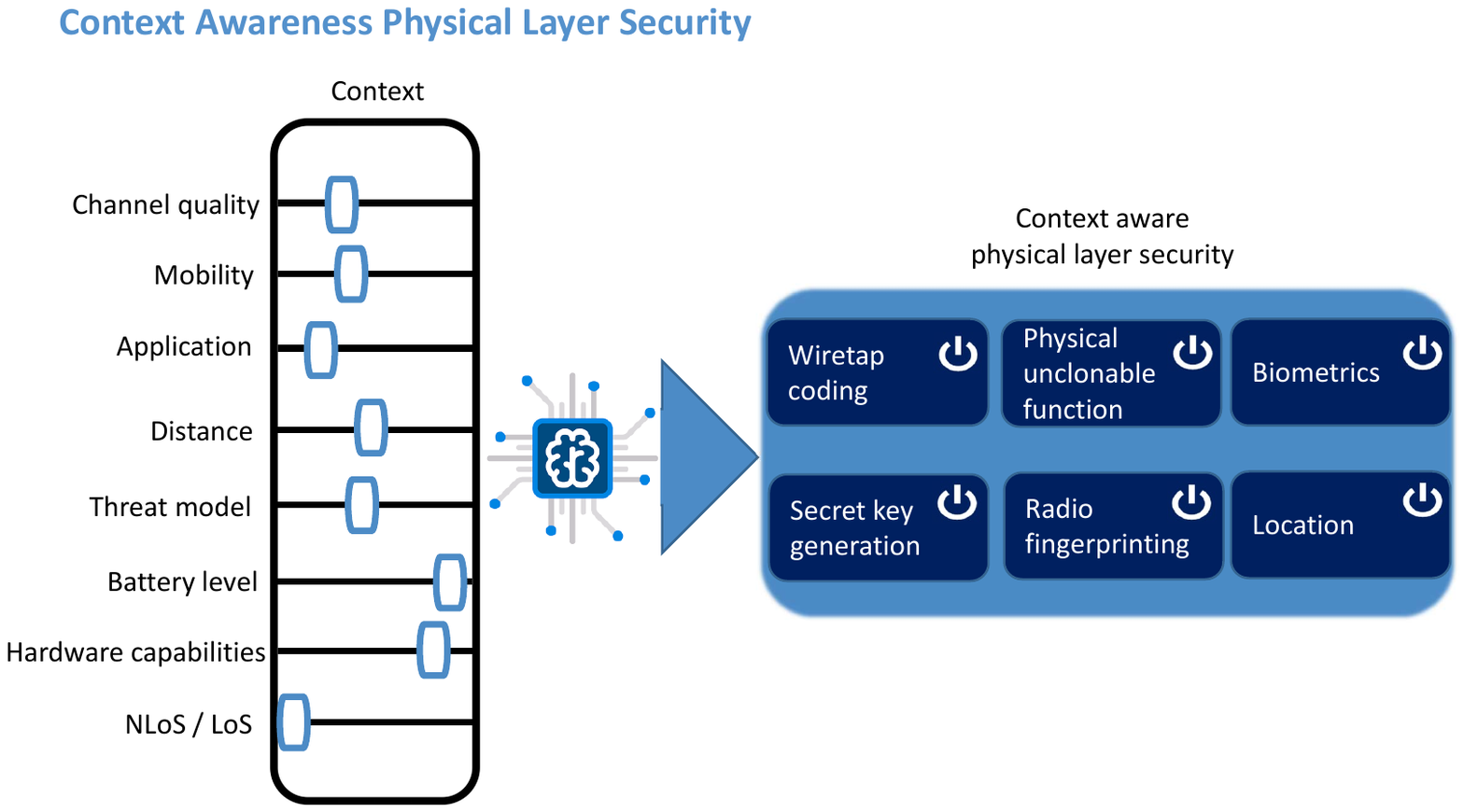}
    \caption{Context awareness physical layer (PHY) security. }
    \label{fig:Context_aware_PLS}
\end{figure*}

\subsection{The Role of Physical Layer Security in Future Generations of Wireless}

Cryptography-based security relies on high-complexity cryptographic algorithms presuming that the resources that are required to break them are unattainable for many attackers.
With myriads of low-end IoT wireless sensors, the high algorithmic complexity of current security algorithms can be an issue. The severity of lacking such low-complexity IoT friendly security schemes is manifested in the low protection levels of IoT data; we note in passing  that according to an industry report \cite{IoT_99_claim} $98\%$ of the IoT traffic is unencrypted. 
Furthermore, as a result of the anticipated quantum revolution, even the current high-complexity encryption may not be secure in the quantum era and future-proof quantum-resistant security
systems should be developed \cite{Quantum_comp_security_2018}. 
These call for developing potentially low-complexity/quantum resistance security solutions that holistically reexamine previously unused or underutilized physical dimensions of the systems to avoid solely relying on high-complexity algorithms as security measures  \cite{Survey_PLA_Wireless_2021}.

In this section we survey how a new take on the role of the physical layer in future generations of wireless networks can provide new concepts of trust. In particular, we review current and future advances in \emph{Physical Layer Security} (PLS).

To overcome several outlined challenges related to trust, Physical Layer Solutions (PLS), which exploit the characteristics of physical
phenomena to provide security, could complement conventional upper-layer security schemes to provide authentication (discussed in the previous subsection), confidentiality (keyless transmissions) and key generation and distribution \cite{Shakiba-Herfeh2021}. PLS is considered a scalable approach to
strengthening the overall trust and resilience of 6G, with a particular focus on constrained devices. Among the possible solutions currently considered are hybrid crypto-PLS approaches, e.g., by using PHY generated resumption keys in zero-round-trip time (0-RTT) protocols \cite{Mitev22}. %Another reason for considering PLS has to do with scalability. \textcolor{red}{Quantum-resistant}, i.e., unbreakable by a quantum computer, \textit{and} lightweight authentication and key distribution for massive IoT, persists as an open challenge at present, as the recently standardized post-quantum cryptographic algorithms are still computationally expensive for low-end devices. To address all of these issues, PLS technologies emerge as complementary schemes to standard cryptography \cite{Shakiba-Herfeh2021, Tomasin22, Chorti23}

\subsubsection {Emerging hardware and physical-layer authentication methods}

\begin{figure*}
  \centering
  \vspace{-0.1in}
    \includegraphics[scale=0.15]{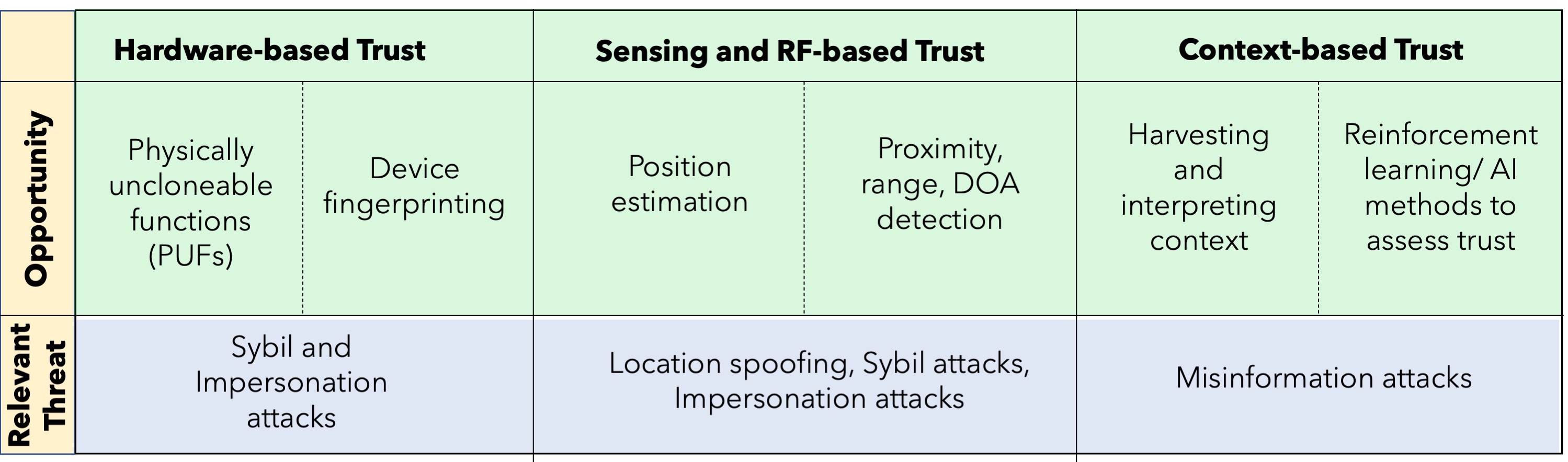}
    \vspace{-0.1in}
  \caption{Examples for new opportunities for security and resilience in future networks. 
%  \my{TODO: update this figure to include MiM attacks.}
  }
  \label{fig:opportunities}
\end{figure*}

This subsection reviews recent trends in hardware and physical layer authentication methods. 
It additionally discusses leveraging the context of operation and communication as a key factor enabling building trust and evaluating in real-time the trustworthiness of the algorithms that run autonomous cyber-physical systems. Fig.~\ref{fig:Context_aware_PLS} captures examples for  physical contexts that can be extracted and leveraged to provide additional security measures
The main focus of this survey paper is on the exploitation of these methods, which can be thought of as distinct or supplementary to some of the more traditional higher-layer authentication methods described in the previous section. Furthermore, Fig.~\ref{fig:opportunities} shows what threats these new opportunities can provide security and resilience to.

%\mynote{The difference between higher layer CR authentication and PHY CR authentication is not very clear at first read.  I believe that since the higher layer CR comes first it should include more explanation regarding what is CR authentication. Then when we discuss the PHY CR authentication we should highlight how it differs from the higher layer CR authentication. }

\textit{\textbf{Hardware-based authentication methods}} comprise hardware-based CR protocols.  Recall that, typically,  CR-based methods  consist of two stages, namely an enrollment stage and a release (authentication) stage. The enrollment stage occurs offline. In this stage, the unique hardware-related characteristics of the node, or the user to be authenticated are measured, this can include silicone imperfections, and phase noise that stems from the crystal that is used for frequency generator, among other things. Hashed versions of these measurements along with related helper (side) data are stored on the verifier side in a database. In the release stage, new measurements are taken and sent to the verifier; the latter uses the helper information to regenerate the hash of the initial measurement, in which case the authentication is successful. A major advantage of such approaches is that the device to be authenticated {does not need to store any pre-shared secrets} and responds to challenges by performing measurements on its own hardware.  %The role of the helper information is critical as it allows to correct for discrepancies between different measurements due to noise (in any actual system,  deriving the same exact outcome from two consecutive measurements is impossible). %Therefore, in order to assist the verifier to reproduce the stored measurement results a helper data is also stored in the database in the enrollment stage. Therefore, users base on new measurements and the helper data reproduce the information stored in the database. 
%Error correcting codes from the family of Slepian Wolf encoders are typically used in these systems; as an example, if the implementation is based on linear block codes, the helper information is typically in the form of the syndrome of the initial measurement.

A prominent example of hardware-based authentication relies on \textbf{physical unclonable functions} (PUF).
 The concept a physical unclonable function was first introduced in \cite{Pappu02}. The idea is that integrated circuits (ICs) have unique characteristics in their physical microstructure due to inevitable and unpredictable perturbations during the fabrication process. These unique characteristics are {unclonable} and can be considered as signatures of the ICs. A PUF needs to be {unclonable}, whereby, given the exact fabrication procedure, it should be infeasible to reproduce the exact same physical microstructure. Recent advances in the understanding of PUFs have fueled increased interest in these authentication techniques ~\cite{PUFsgunlu2019code,PUFschen2018secret,PUFsOptimalitySummary}. For PUFs, this includes analyses regarding strong and
weak PUFs, and the proposal of methods for generating and reconstructing secret keys from noisy PUF outputs, for example, based on Wyner-Ziv coding~\cite{PUFsgunlu2019code}. There are currently
numerous commercial products based on PUFs and increased research interest in the design of
fuzzy extractors \cite{fuzzy} which are aimed at reducing leakage of secret keys due to correlations in the
PUF outputs \cite{securePUFs}.

Additional ways to harness hardware to authenticate nodes, utilize  unique transmitter and receiver characteristics such as I-Q imbalances, nonlinearities, phase noise, and antenna signatures, among others. However, unlike the case of PUF-based authentication, currently, there are no strong guarantees regarding the unclonability of a device's fingerprint and such approaches can be useful for soft-authentication.
%\subsection{Anomaly Detection (Direction of Arrival, }
%\subsection{Physical-layer methods: RF- and sensing-based}

\textit{\textbf{Positioning and RF-fingerprinting-based authentication}} leverages the intrinsic properties of the
wireless channel between two devices. 
% \sg{Wireless channels consist of two parts, the first  large-scale fading related to the agents' position via path loss and shadowing, and stochastic small-scale multipath fading owing to interference and variations in the wireless channel.}
%\my{Stephanie, I revised the following paragraph though I saved the old text. Feel free to change it :)}
Wireless channels are characterized by large-scale and small-scale multipath propagation effects, aka fading, of the transmitted radio wave \cite{Goldsmith}. 
Large-scale fading is caused by path loss and shadowing that are related to the transmitter and receiver positions. As such, it is slow to change and is characterized by a deterministic factor.
The signals sent from the transmitter to the receiver also suffer from reflections, diffraction, and scattering which induce small-scale multipath fading.
This causes fast variations in the power of the received signal due to the arrival of the received signal through multiple paths with different time delays and phase shifts which are captured by statistical terms.  
An additional key distinction between deterministic fading and stochastic fading \cite{Goldsmith} is that the former can be used to develop location-dependent RF-fingerprints and/or positioning estimates \cite{Senigagliesi2020}, while the latter can be exploited as a source of shared randomness for the distillation of symmetric keys \cite{Srinivasan_Joint_localization_2022}. Furthermore, high precision positioning and tracking will be enabled by the extensive use of sensing (e.g., radar sensing), which in turn will allow for integrity and consistency checks of estimated positions. 
Similarly, channel charting inspired approaches for Channel State Information (CSI) pre-processing can enhance the separability of wireless fingerprints, as shown in \cite{Srinivisan21}. 
As we discussed in Section \ref{sec:threats}, many attacks forge the attacker's location. For example, since two transmitters cannot be at the same physical location in a location-based service a Sybil attacker must report false locations for the identities it creates. Since the RF and radar fingerprints are affected by the positioning of the transmitter and receiver, they can be exploited for positioning verification and authentication and assist in detecting untrustworthy transmitters.

\begin{svblock}{Fusion for accuracy of sensing-based functionalities}
 Exploiting both RF and sensing inputs (e.g., radar data), it is possible to substantially increase the accuracy of proximity, range, AoA and DoA estimation~\cite{radar}.   
\end{svblock}

%\subsubsection{Proximity, Range, Direction of Arrival}
%Exploiting both RF and sensing inputs (e.g., radar data), it is possible to increase substantially the accuracy of proximity / range / AoA and DoA estimation. More precisely, ++++

\subsection{Feasibility of Physical Layer Security in the 6G Era} 

Despite its great potential, translating PLS theory to practice remains elusive, with only a handful of PLS products  currently available for device authentication using PUFs and fingerprinting. Despite the lag, 6G will bring about several enabling technologies which render PLS feasible. 
We discuss them next in further detail and delay the discussion of the  broader outlook they enable and its challenges for Section \ref{sec:security_broader}.

% \begin{itemize}
% \item 
\subsubsection{Channel engineering and controllability}
Channel engineering and channel controllability, which  shape and control the channels transmitted signals propagate through, are key elements in 6G, enhancing the
first steps taken in 5G with the introduction of millimeter waves (mmWave). Channel
engineering and controllability are under study, especially in the form of pencil-sharp
beamforming in ultra-massive multiple input multiple output (MIMO) systems at mmWave and
THz bands, allowing to make the case for wiretap channels \cite{wiretap_wyner1975wire}
 without relying on assumptions
regarding the position or number of antennas of passive attackers. For wireless channels, the knowledge of local geometry and of the propagation environment provides important context information for the probabilistic secrecy characterization of a transmission. In this direction, secrecy
maps, where the channel characteristic between any two physical points in a map can be learned from measurements \cite{Utkovski21,secrecyMaps_utkovski2019learning}, and related concepts can be of great interest. Another recent advancement
concerns the use of reconfigurable intelligent surfaces (RIS), see ~\cite{RISelMarco2020reconfigurable,RISelmossallamy2020reconfigurable,RISliu2021reconfigurable,RISelMarco2022reconfigurable}, which are programmable surfaces that can control the phase, polarization, amplitude, and eventually frequency \cite{RIS_frequency_2021}  of impinging signals, effectively allowing to steer the reflection of the transmitted signal. The use of RIS systems, drones~\cite{GoogleProjectLoon,FBconnectivityLab}, and multi-hop wireless networks have been studied to either improve the received signal
quality or used in challenge-response authentication protocols \cite{Tomasin22}.

% \item 
\subsubsection{Device positioning}
%\mynote{I moved this paragraph to here to create a nice flow. First, we discuss positioning. Then we add contextual awareness via camera, reader, etc. Then we discuss the rest of the points.}
In 6G, positioning will be a default service. As a result, positioning integrity is of
paramount importance, and angle of arrival (AoA) \cite{Xiong:Securearray:2013, Abdelaziz2019, Xu2022} in conjunction with ranging, camera depth
estimations, and other sensing methods, are studied to render positioning unspoofable. Several protocols have already
been proposed that incorporate positioning information as a second soft authentication factor \cite{Mitev22},
and Sybil cyberattacks in robotic systems have been detected using AoA \cite{Location_spoofing_2016_geo_spatial,GilAURO}. Overall, positioning
will be an important parameter for evaluating the trustworthiness of autonomous agents in 6G, building on leveraging context \cite{Nokia2}.

\begin{svblock}{Standardization of positioning services in 6G}
 In 6G positioning is expected to be a default service~\cite{Hexa-X-vision} and will be an important parameter for evaluating the trustworthiness of autonomous agents in future networks.    
\end{svblock}

% \item 
\subsubsection{Additional contextual information}
The introduction of sensing in 6G will provide contextual information, such as supporting the use of cameras and radar. in addition to the  wireless antennas, to extend the notion of RF fingerprint to include additional measures for situational awareness.  
The sensing capabilities will allow receiving nodes to better assess the wireless propagation environment and enable adaptive PLS solutions to secure against adversaries and untrustworthy nodes. This includes but is not limited to leveraging  the additional sensing received inputs to improve the estimation of  line-of-sight conditions and localization of transmitting sources (including jammers and impersonating nodes) and potential eavesdroppers \cite{Chorti22}.
%\sg{for example ranging capabilities,}
% line-of sight \sg{which is useful in deciding when wiretap coding is applicable, and others}
An additional and closely related contextual information that complements physical knowledge is the concept of semantics, i.e., meaning, and context-enabled PLS in which PLS may
further benefit from \cite{Chorti22}. In particular, the meaning and importance (semantics) of exchanged
messages is of much interest as it allows us to assess the value of information and make
decisions regarding the required security level.

% \textcolor{red}{OLD TEXT:
% The introduction of sensing in 6G will provide  contextual information (including ranging
% capabilities) e.g., to identify the presence of jammers, line-of sight (e.g., to decide when wiretap coding is applicable), etc. Sensing capabilities will allow to better assess the propagation environment and enable adaptive PLS solutions. Closely
% related to the latter is the concept of semantics and context enabled PLS in which PLS may
% further benefit by context awareness \cite{Chorti22}. Furthermore, the meaning / importance (semantics) of exchanged
% messages is also of interest as it allows to assess the value of information and make
% decisions regarding the required security level.}

% \my{How does the following item differs from the RF fingerprint one?}
%sgnote{From this paragraph it is not clear how the contextual information provided in 6G will help to detect the presence of jammers and line-of-sight. Is this related to the ranging capabilities? Also, can we explain why is line-of-sight needed for wiretap coding?}
%\mynote{I rewrote the paragraph to make it clearer. Let's verify with Ersi that this is what she how visioned it.}

% \item
\subsubsection{Secret key generation and communication}
A full palette of countermeasures to address both passive (eavesdropping) and active
attacks (man-in-the-middle, jamming) against secret key generation (SKG) systems have recently been proposed, see for example \cite{Mitev21}. At the same time, the development of preprocessing techniques for the isolation of entropy-rich
reciprocal components of the observed Channel State Information (CSI), which captures the stochastic fading effect~\cite{Goldsmith} of wireless signal propagation in the environment, can allow efficient implementation of key generation schemes \cite{Srinivisan21}. 

Furthermore, the secrecy and  security of communicating via keyless or SKG codes depend on the codelength of the communicated codewords and keys which are finite in any practical setup.   
The analysis of the finite block-length for secrecy and reconciliation encoders allows for attaining a clearer understanding of the trade-off between codelength, error packet
rate, information leakage, and secrecy, as is discussed, for example, in \cite{SKG_length_2015,Yang19,Herfeh21}.

In the next section we show how these enablers in 6G can be used to provide valuable contextual awareness and thereby trust, for future emerging networks.

\subsection{A Broader Look at Trustworthiness in 6G and Beyond}\label{sec:security_broader}

In this section, we attempt to provide a holistic view of how trust can be built in 6G by leveraging the physical layer and contextual information and the challenges that await.

\subsubsection{The role of context awareness and semantics in trusting  autonomous systems}
Context and semantics can serve as a base for autonomous controls and cyberphysical systems that react accordingly to their physical and cyber realities. Context is a construct that goes beyond the pure description of the environment in terms of \textit{\textbf{where}}, \textit{\textbf{when}}, \textit{\textbf{what}} and \textit{\textbf{who}} (known as the four Ws) based description of context. Therefore, context-awareness \cite{Chorti22} becomes closely related to situational awareness, i.e., the process of perception of elements in the current situation, the comprehension of the current situation and the projection of future status \cite{Schilit94}.
%With respect to semantics, early approaches in the direction of deﬁning a formal framework for semantic communications have appeared in [Bao11], in essence extending the standard Shannon model of reliable communication of binary sequences to reliable communication of semantic messages. The concept of semantic entropy was introduced and employed in semantic source and channel coding statements. 
As 6G is dubbed as the first AI-native\footnote{Edge and on-device intelligence will enable real-time operation, without a human-in-the-loop, of autonomous agents in 6G, thus rendering it an ``AI-native'' generation.} wireless generation, it seems natural to invest on AI to interpret semantics and context in future systems. %, using for example of attention mechanisms and NLP based tools. 
More broadly, new means of reaching situational awareness by harvesting and interpreting context are envisaged. %related aspects encompass high-precision localization -- projected to reach centimetre level -- sensing from multiple sources that will allow obtaining an understanding of the type of nodes, in addition to the time of communication and the age of information, the type of data exchanged, etc.   

\begin{svblock}{Context extracted from physics-centered information}
Context can be extracted jointly from physical and cyber sources, i.e., from radio frequency (RF) inputs, hardware and sensors, as well as network attributes~\cite{Chorti22}.   
\end{svblock}

Incorporating context awareness in trust building can allow handling more efficiently aspects related to identifying risk or threat level and required security level, particularly for applications such as autonomous driving, robots, e-Health, etc. As a first step, one can consider the first of the four Ws, \textit{where}, i.e., the positioning and tracking of autonomic devices, with respect to their expected \textit{normal} behavior. More broadly, context should be extracted jointly from physical and cyber sources, i.e., from RF inputs, hardware and sensors as well as network attributes, pointing to the need for fusion of cross-layer attributes, potentially through the use of multi-modal fusion.

\subsubsection{Remaining challenges for extracting trust in future networks} We briefly mention that the vision of using the enabling technologies in sensing, positioning, and channel engineering that are expected in next-generation networking to derive trust is not free of its own set of challenges. For example, in sensing, there exist key questions of privacy and reliability. In an interesting result in a recent paper \cite{SensingAttack22}, it has been demonstrated that is it possible to mount MiM attacks such as ``injecting'' fake humans in a given sensed environment that uses standard radar waveforms used for positioning and tracking, namely frequency-modulated continuous wave (FMCW) \cite{fmcw}. It is noteworthy that the authors developed the injection as an approach for enhancing privacy, i.e., protecting against unsolicited radar sensing. However, this work also raises concerns regarding the trustworthiness of the sensing. In decentralized computing, the virtualization of the radio, open radio access network (ORAN), towards an open-source software, heterogeneous hardware, and interoperable interfaces brings about the need for novel security architecture designs, e.g., see \cite{Stefan22}. As an example, in a recent report from the European Union on security risks related to ORAN, misconfiguration of networks and low-quality products are highlighted as amplified risks due to the open-source nature of ORAN, while at the other end of the spectrum, dependency on (potentially untrusted) suppliers is lowered as a risk \cite{EU_ORAN}. To this end, multiple works have investigated in recent years methods to secure distributed, and specifically virtual networks, via anomaly detection which has been extensively investigated in recent years. 

Nevertheless, \emph{the ability to infer context from the sensing, computing, and channel engineering capabilities expected from future generations of wireless technologies, will be invaluable for paving the way to trust-centered cyberphysical systems.} Context can be used to better assess, and perhaps quantify, the legitimacy (i.e., trustworthiness) of a link or of a user (agent). This yields important implications for security in these networks, which is the focus of the next section of this survey paper.

\section{Trust-enabled Resilience Guarantees for Multi-agent Systems} 
\label{sec:performance}

The focus of this section is on use cases for networks supporting multi-agent tasks. In particular, we focus on the \emph{achievable performance} of multi-agent systems in the face of untrustworthy agents on the network. In this section we describe in more detail several concrete trust-based performance guarantees that have been derived in the literature for many key  applications which we review in Section~\ref{sec:intro} and in \an{Fig.~\ref{fig:trustSchematic}}. Specifically, we discuss applications in the domains of \emph{Democratized Sensing} and \emph{Democratized Computing} ranging from distributed consensus, to distributed optimization, coverage, distributed model predictive control, federated learning, and others. We categorize our discussion by \emph{trust type}, where different notions of trust have led to different achievable performance guarantees for multi-agent systems.  In particular we consider four different notions of trust that have been successfully applied in multi-agent systems:
\begin{enumerate}
    \item \ul{Stochastic trust:} This type of trust refers to the case where the likelihood of a link in the network being untrustworthy can be quantified. Often times additional information from sensors onboard the agents, embedded signatures in the transmitted signals, or orthogonal data such as agent positions are used to derive these likelihood probabilities.
    \item \ul{Reputation-based trust:} This type of trust often arises from comparing data transmitted by different agents against each other, or against a known model of nominal behavior. 
    \item \ul{Network measures of trust and robustness:} Unlike the previous two measures of trust that we consider here, this metric is not an explicit quantification of trust between agents or over particular links, but rather a topology-specific measure that quantifies \emph{redundancy of information} in the network. 
\end{enumerate}
For the most part, the works surveyed in this section use a representation of the multiagent network as a graph $\mathbb{G}=(\mathbb{V},\mathbb{E})$, where $\mathbb{V}=\{1,\hdots,n\}$ denotes the set of node indices for the agents and $\mathbb{E}\subset\mathbb{V}\times\mathbb{V}$ 
denotes the set of edges. The notation $\{i,j\}$ is used to denote the edge connecting agents $i$ and $j$. The set of \emph{neighbors} of node $i$ is denoted by $\mathcal{N}_i= \{j\in\mathbb{V}\mid \{i,j\}\in\mathbb{E}\}$.  The malicious or non-cooperative agents are represented by a subset of nodes $\MM$, $\MM\subset \mathbb{V}$. These are either \emph{malicious} or \emph{malfunctioning} and, thus, do not reliably cooperate in the coordination protocols. The set $\MM$ is often assumed to be unknown, with the cardinality $|\MM|$. When referring to an agent that is not malicious or malfunctioning we use the term \textit{legitimate}, and the set of legitimate agents is denoted $\mathcal{L}$, $\mathcal{L}\subset \mathbb{V}$,
with the cardinality $|\LL|$ and $\mathbb{V}=\LL\cup\MM$.

\begin{table*}[ht]
    \centering
    \begin{tabular}{p{0.3\linewidth} | p{0.3\linewidth} | p{0.3\linewidth}}
      \textbf{Problem Class} & \textbf{Trust Type} & \textbf{Resilience-related Performance Guarantee} \\ \hline
      Resilient distributed average consensus & Stochastic trust & Convergence to average consensus value~\cite{TRO_foundations,hadjicostisCDC22} \\ \hline
      Resilient distributed optimization & Stochastic trust & Convergence to true optimal value over legitimate agents almost surely~\cite{distOptCDC}\\ 
    \hline
    Machine learning with faulty reported gradients for the loss function & Reputation-based trust & Convergence with faulty gradient estimates~\cite{zeno}\\
    \hline
    Distributed Model Predictive Control with attackers & Reputation-based trust & Agents can identify and mitigate the influence of adversarial neighbors in the context of a distributed MPC algorithm~\cite{ananduta2020resilient}\\
    \hline
    Coverage and sensing with unreliable agents & Reputation-based trust & Convergence to performance-weighted configuration~\cite{schwagerAdaptiveSensingCoverage}\\
    \hline
     Resilient consensus from an information theoretic perspective & Network measures of robustness & Generic identification of misbehaving agents~\cite{bulloUnreliable}\\
     \hline
     Weighted Mean-Subsequence-Reduced (WMSR) & Network measures of robustness & Approximate agreement for the $f$-local malicious adversary case~\cite{SundaramWMSR}\\
     \hline
    \end{tabular}
    \caption{Example performance guarantees for different trust types.}
    \label{tab:perfGuarantees}
\end{table*}

We organize the rest of this section by trust-type and we survey several results from the literature on performance guarantees for each of these trust types.
\subsection{Stochastic Interpretations of Trust and Novel Resilience Guarantees}
% \begin{wrapfigure}[11]{r}{0.45\textwidth}
\begin{figure}
  \centering
  % \vspace{-0.1in}
    \includegraphics[scale=0.1]{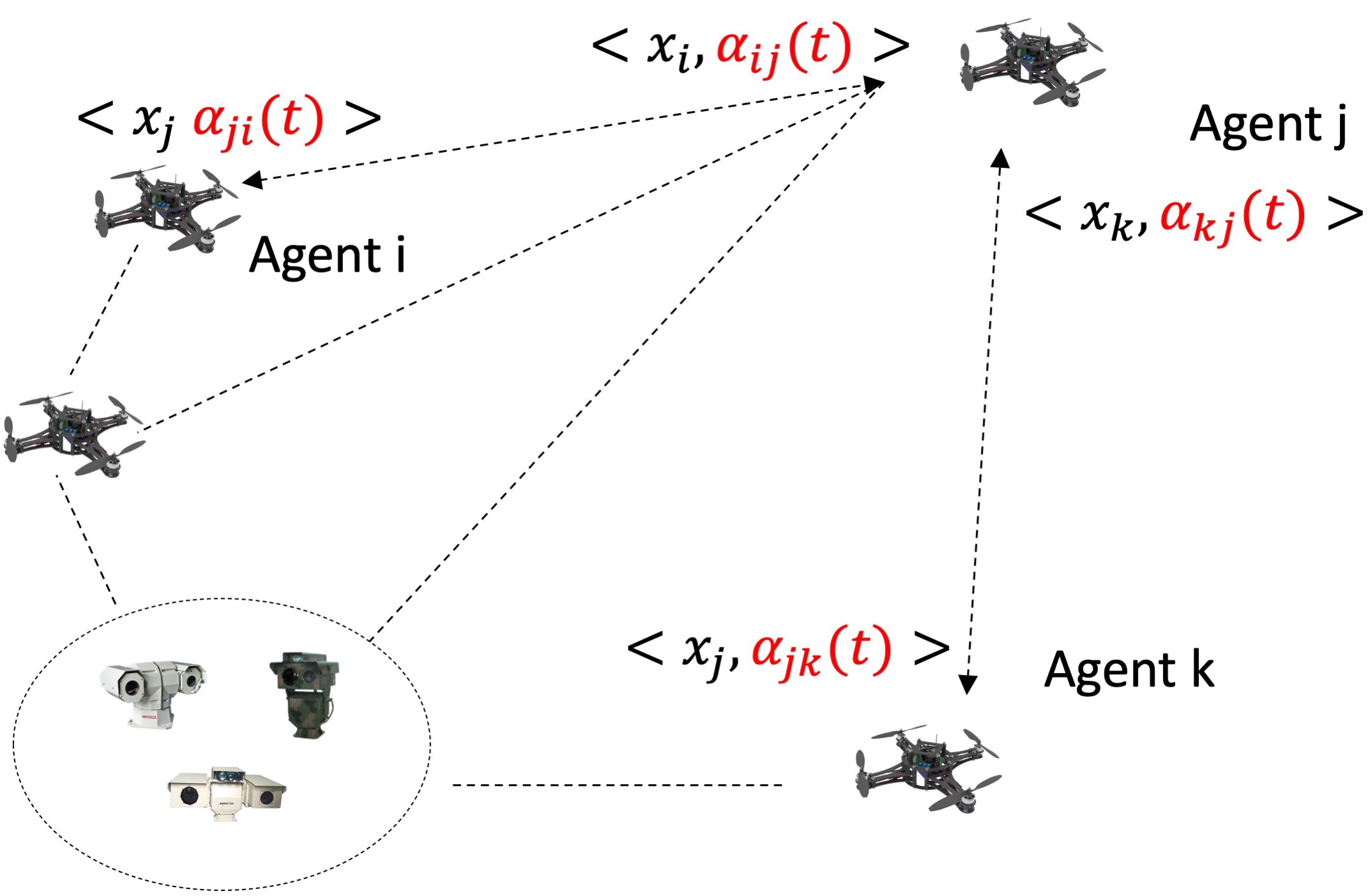}
    % \vspace{-0.3in}
  \caption{Multi-agent cyberphysical system where trust values are derived for links in the network.} %\my{This plot does not match how we define $\alpha_{ij}$ in our TRO and TAC papers.}}
  \label{fig:multiAgent}
  \end{figure}
% \end{wrapfigure}
In this section we will summarize work employing stochastic values of trust that lead to new, stronger, guarantees on resilience compared to classical long-standing results in resilience; particularly with regards to the proportion of malicious agents that can be tolerated.  We will focus on results pertaining to adversarial consensus, distributed optimization, and learning problems.

The work discussed here considers the case where additional information exists in the network that can be exploited to infer trust for a particular link. For example many of the \emph{physical} channels of information discussed in Section~\ref{sec:network} of this paper can be used as sources for such measures of trust. These can be derived from sensors onboard the agents, embedded signatures in the transmitted signals between agents, agent positions or behaviors, and/or a combination of these. Examples of works deriving and using side ``trust'' metrics are growing in recent literature literature~\cite{panagouTAC22,GilAURO,schwagerAdaptiveSensingCoverage}. An abstraction that can be used to capture the presence of this additional information in the network is modeling this as a \emph{stochastic value of trust}. Here, a higher stochastic trust value measured on a particular link in the network indicates a higher likelihood that the link is trustworthy and vice-versa. Along these lines, the works in this section consider the availability of a stochastic trust value, $\alpha_{ij}$, for each link in the network $(i,j)\in \mathbb{E}$. In particular

\begin{svblock}{Stochastic values of trust}
 This section considers the existence of a random variable $\alpha_{ij}(t)\in(0,1)$ taking values closer to 1 with higher probability in the case that the link $(i,j)\in \mathcal{L}$ and closer to 0 with higher probability in the case that the link $(i,j)\in \mathcal{M}$. We refer to these $\alpha_{ij}(t)$ values as \emph{stochastic values of trust}.    
\end{svblock}

% \noindent \framebox[1.01\textwidth]{
% 	\parbox{\textwidth}{\underline{Stochastic values of trust:} This section considers the existence of a random variable $\alpha_{ij}(t)\in(0,1)$ taking values closer to 1 with higher probability in the case that the link $(i,j)\in \mathcal{L}$ and closer to 0 with higher probability in the case that the link $(i,j)\in \mathcal{M}$. We refer to these $\alpha_{ij}(t)$ values as \emph{stochastic values of trust}.}
% }\\ 

Stochastic values of trust can be derived empirically or analytically. The paper~\cite{GilAURO} shows that $\alpha_{ij}(t)$ values can be extracted from processing the wireless signals transmitted between communicating agents by examining their physical properties for signatures of uniqueness. Furthermore, the work \cite{GilAURO} characterizes some analytical properties of the $\alpha_{ij}(t)$ derived from angle-of-arrival (AOA) fingerprints, showing that in expectation they can be bounded closer to 1 for the legitimate link case and closer to 0 for the malicious link case for a Sybil Attack. From a performance standpoint, provable performance guarantees can be attained if certain characterizations of the $\alpha_{ij}(t)$, such as bounds on their expectation, and/or variance, can be shown analytically. In this section we will focus on these types of characterizations and resulting performance guarantees in resilience for many problems from consensus, to distributed optimization, and others that we briefly survey here.

We first review characterizations of the stochastic values of trust that have been derived in the literature, specifically, we consider three possible classes of characterizations depending on what is known for the problem at hand: i) bounds on expectation, ii) bounds on variance, and iii) known distribution. These characterizations have been used to arrive at different performance guarantees that will be surveyed here. Next we will better describe these different characterizations, and we will provide examples of what types of resilient performance guarantees have been derived for distributed consensus, optimization, and learning problems. 

The first characterization relies on bounds on the expectation of the $\alpha_{ij}(t)$, i.e., this is a first-moment characterization. These constitute mild assumptions on the characterization of the $\alpha_{ij}(t)$ leading to significantly stronger results for resilience than the classical Byzantine adversary problem~\cite{DOLEV198214}. Here, a characterization on the expectation of the $\alpha_{ij}(t)$ is one where there exists some $0<\epsilon<1/2$ such that
	\begin{align}
  \label{eq:alphaExpectations1}
	    &E[\alpha_{ij}(t)|i\in\LL,j\in\LL]>1-\epsilon,\\
	    &E[\alpha_{ij}(t)|i\in\LL,j\in\MM]<\epsilon.
     \label{eq:alphaExpectations2}
	\end{align}

% \noindent \framebox[1.01\textwidth]{
% 	\parbox{\textwidth}{\underline{Bounds on Expectation of the Stochastic Values of Trust:} If bounds on the expectation of the stochastic values of trust hold then there exists some $\epsilon>0$ where
% 	\begin{align}
% 	    &E[\alpha_{ij}(t)|i\in\LL,j\in\LL]>1-\epsilon\\
% 	    &E[\alpha_{ij}(t)|i\in\LL,j\in\MM]<\epsilon
% 	\end{align}}
% }\\ 

% \begin{wrapfigure}[11]{r}{0.4\textwidth}
\begin{figure}
  \centering
  % \vspace{-0.2in}
    \includegraphics[scale=0.1]{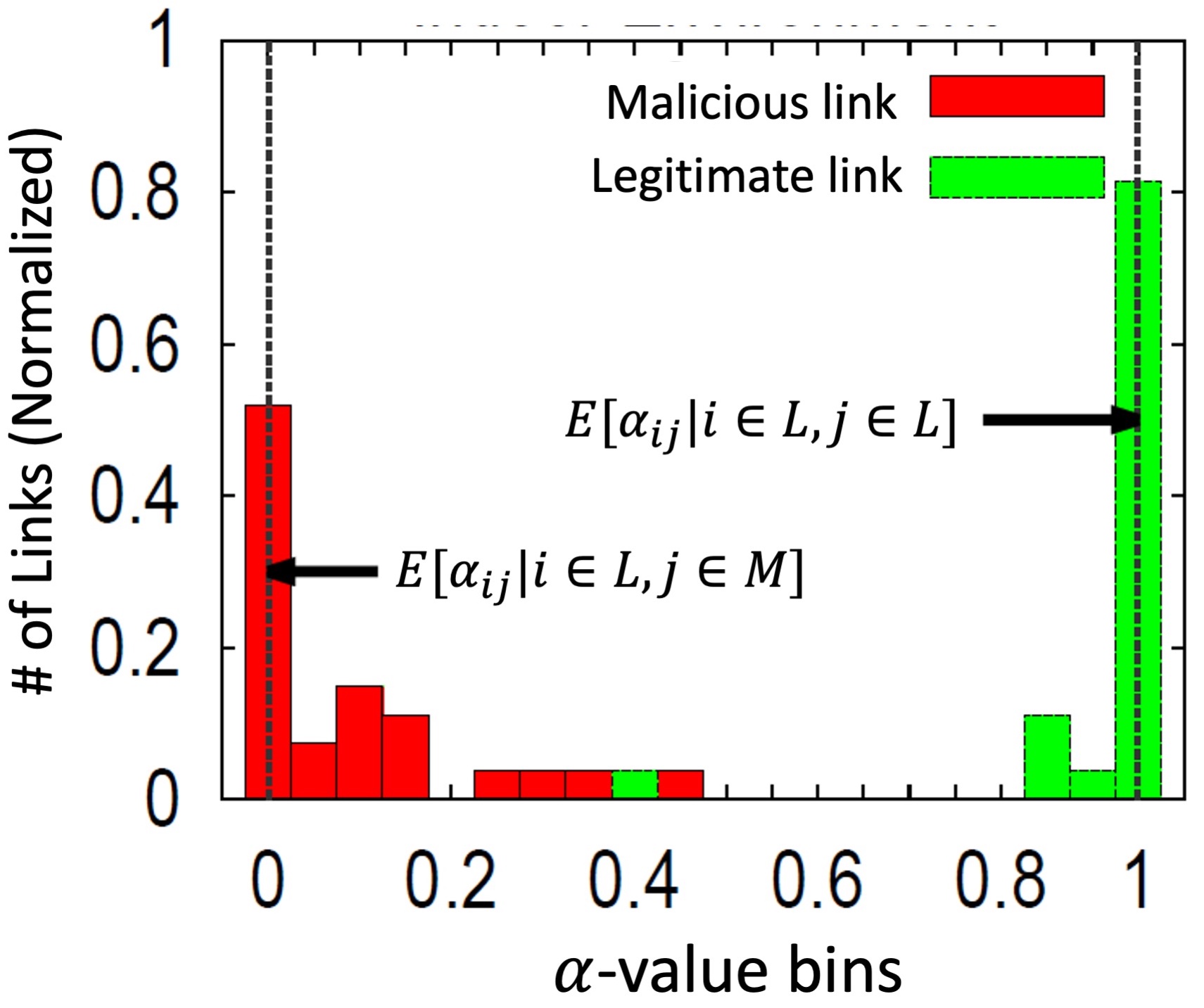}
    % \vspace{-0.1in}
  \caption{Histogram of $\alpha$-values for malicious and legitimate edges. This histogram demonstrates actual experimental evidence of the conditions expressed in Eqs.~\eqref{eq:alphaExpectations1} and \eqref{eq:alphaExpectations2}.}
  \label{fig:empiricalExpectationAlpha}
% \end{wrapfigure}
\end{figure}
\begin{figure}
  \centering
  %\vspace{-0.2in}
    \includegraphics[scale=0.1]{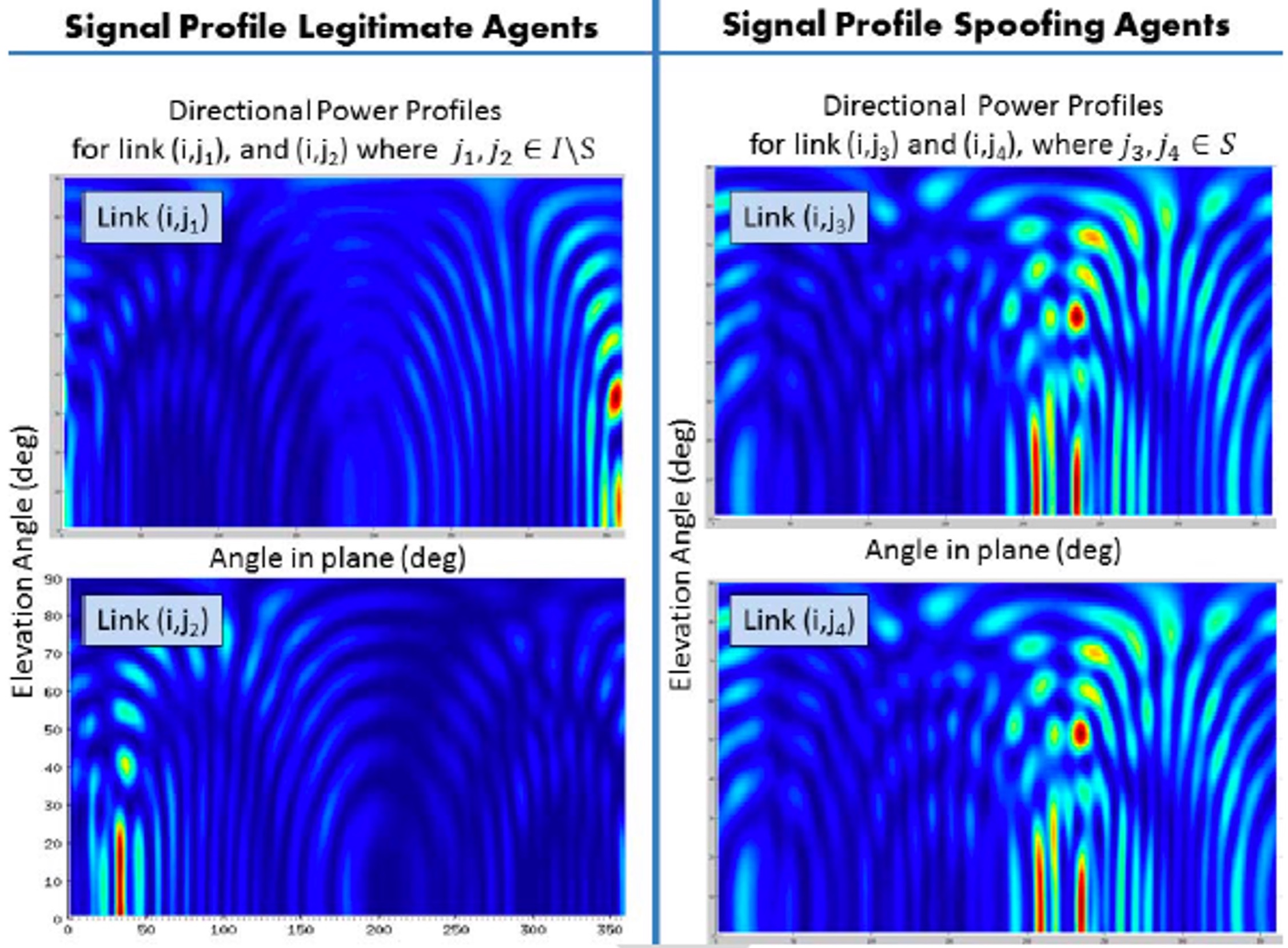}
    \vspace{-0.1in}
  \caption{Angle-of-Arrival wireless fingerprint.}
  \label{fig:AOA_fingerprints}
\end{figure}

This is a very mild assumption and it is to say that the trust values are \emph{informative} so long as the $\epsilon$ is bounded away from 1/2 (the case of no information).  An example of such a characterization holding is derived in the paper~\cite{GilAURO} for the case of a Sybil Attack where wireless AOA fingerprints are used to extract the $\alpha_{ij}(t)$ values. In that work, the $\alpha_{ij}$ values capture the uniqueness of the measured transmission AOA (see Fig.~\eqref{fig:AOA_fingerprints}).   Further, experiments on an actual multi-robot system subjected to a Sybil Attack over the wireless medium empirically show agreement with this bound on the expectation of the stochastic trust values (see Fig.~\ref{fig:empiricalExpectationAlpha}).
% \subsubsection{Characterizations of Trust}

Another statistical characterization of the $\alpha_{ij}(t)$ can rely on the second-moments, i.e., the variance of these random variables, and assume that they are exactly or approximately known. In this case, stronger results on convergence rate for many distributed adversarial multi-agent problems can be derived. This characterization of the $\alpha_{ij}(t)$ requires that their variance is known such that
	\begin{align}
	    &\text{Var}(\alpha_{ij}(t))=\sigma_\LL^2 \text{ for } t\geq 0,\ i\in \LL,\ j\in \mathcal{N}_i\cap \LL,\\
	    &\text{Var}(\alpha_{ij}(t))=\sigma_\MM^2 \text{ for } t\geq 0,\ i\in \LL,\ j\in \mathcal{N}_i\cap \MM,
	\end{align}
	\noindent for $\sigma_\LL,\sigma_\MM>0$.    

% \noindent \framebox[1.01\textwidth]{
% 	\parbox{\textwidth}{\underline{Known Variance of the Stochastic Values of Trust:} This characterization of the $\alpha_{ij}(t)$ requires that their variance is known such that
% 	\begin{align}
% 	    &Var(\alpha_{ij}(t))=\sigma_\LL^2 \text{ for } t\geq 0,\ i\in \LL,\ j\in \mathcal{N}_i\cap \LL\\
% 	    &Var(\alpha_{ij}(t))=\sigma_\MM^2 \text{ for } t\geq 0,\ i\in \LL,\ j\in \mathcal{N}_i\cap \MM
% 	\end{align}
% 	\noindent for $\sigma_\LL,\sigma_\MM>0$.}
% }\\ 

An even more informative statistical characterization of the stochastic values of trust $\alpha_{ij}(t)$ would be when the distribution of these random variables is known. We suspect that this would lead to stronger resilient performance guarantees for more general classes of distributed multi-agent problems. One such example can be found in~\cite{advHypTestingICRA} for the case of adversarial hypothesis testing, although this seems to largely be an open problem and we do not include many such results here. Nonetheless, we note that in the case of a Gaussian distribution, the first and second moments are sufficient to fully characterize the continuous Gaussian distribution.

The next section summarizes several works that use these different characterizations involving stochastic observations of trust, to arrive at resilient performance guarantees for different distributed multi-agent problems. To highlight the unique contribution of each group of works we summarize them in Table \ref{tab:perfGuarantees}.

\subsubsection{Performance Guarantees for Resilient Multi-Agent Algorithms}

Using the characterizations outlined in the previous section, we now survey several results in resilience that have been derived in the literature for multi-agent systems. We consider results for three types of distributed multi-agent problems: those of 1) consensus, 2) distributed optimization, and 3) learning trust of other agents in the network. We define these problems below and summarize their known performance guarantees when stochastic observations of trust are available.

\paragraph{Resilient Distributed Average Consensus}The distributed average consensus problem concerns the problem of agreement where all legitimate agents in the graph $\mathbb{G}$ wish to compute the average of the initial legitimate agents' values. In other words, \textit{legitimate} agents update their value according to the following update equation for all $t\ge T_0-1$, for some arbitrary $T_0> 0$:

\begin{align}
\label{eq:consensusProtocol}
    x_i(t+1)=w_{ii}(t)x_i(t)+\sum_{j\in\mathcal{N}_i}w_{ij}(t)x_j(t),
\end{align}
where $x_i(t)\in\mathbb{R}$, while the weights $w_{ij}(t)$ are nonnegative and sum to 1, i.e.,
$w_{ii}(t)>0$, $w_{ij}(t)\ge 0$ for $j\in {\cal N}_i$ and $w_{ii}(t)+\sum_{j\in{\cal N}_i}w_{ij}(t)=1$. The process is initiated at some time $T_0\ge0$ with the agents' initial values $x_i(T_0-1)=x_i(0)$ for all $i\in\mathbb{V}$. A malicious agent in this context is one that does not follow the update rule in~\eqref{eq:consensusProtocol} for updating its value $x_i(t)$ at some time $t\ge T_0-1$. 

The work in~\cite{TRO_foundations} shows that when the weights $w_{ij}(t)$ are chosen in such a way where they incorporate stochastic values of trust then one can arrive at strong performance guarantees for distributed consensus even in the case of adversarial activity. Stochastic values of trust are incorporated here by assigning equal weight to all neighbors values $x_j(t)$ that are part of their \emph{trusted neighborhood} as defined using a history of trust values $\alpha_{ij}(t)$, and zero weight otherwise. In this case bounds on the expectation of the stochastic values of trust are assumed. Notably, \emph{convergence can be recovered even in the case where the number of malicious agents $|\MM|$ is arbitrarily large} -- a contrast to classical results in resilience where the proportion of malicious agents is usually bounded below half of the network~\cite{DOLEV198214,controlTheoretic_Security_Bullo}. This convergence result is derived as Theorem 1 in~\cite{TRO_foundations}:

\begin{svblock}{Convergence of Resilient Distributed Consensus:}
For the case of the adversarial distributed average consensus problem, when weights are chosen using stochastic observations of trust, then the consensus protocol in Eq.~\eqref{eq:consensusProtocol} can be shown to converge almost surely to a value in the convex hull of the initial values of both the legitimate and malicious agents. This holds independently of the number $|\MM|$ of malicious agents in the network~\cite{TRO_foundations}.   
\end{svblock}

% \noindent \framebox[1.01\textwidth]{
% 	\parbox{\textwidth}{\underline{Convergence of Resilient Distributed Consensus (Theorem 1 in~\cite{TRO_foundations}):} For the case of the adversarial distributed average consensus problem, the consensus protocol in Equation~\eqref{eq:consensusProtocol} converges almost surely, i.e. $\lim_{t\rightarrow \infty} x_{\LL}(T_0,t)=z(T_0)\boldsymbol{1}$ where $\boldsymbol{1}$ is a vector of ones of the appropriate dimension, $x_{\LL}(T_0,t))$ is the value of legitimate agents in the set $\LL$ after running consensus to time $t$ starting at some arbitrary $T_0\geq 0$,  and the value $z(T_0)$ is in the convex hull of the initial values $x_i(T_0), i=1,\ldots,n,$ of both the legitimate and malicious agents. This holds independently of the number $|\MM|$ of malicious agents in the network.} 
% }\\

Furthermore, the deviation of the converged value, denoted by $z(T_0)$, from the converged value in the nominal case where there are no malicious agents, can be characterized as a function of the amount of time $T_0$ that agents are allowed to gather stochastic trust values of their neighbors before starting the consensus protocol (also derived in~\cite{TRO_foundations}). The paper in~\cite{hadjicostisCDC22} includes a convergence result to the true initial average (with no deviation) in the case where stochastic values of trust can be utilized. Finally, for the case where there is a known variance for the stochastic values of trust, tighter convergence rates can be characterized using the results in~\cite{TRO_foundations}. 

\paragraph{Resilient Distributed Optimization}Another important problem of interest is that of distributed optimization where the legitimate agents aim at optimizing a common objective and malicious agents try to impair the computation by injecting harmful data. The aim of the legitimate agents is to distributively minimize the sum of their objective functions $f_i:\mathbb{R}^d\rightarrow\mathbb{R}$ over the constraint set $\XX\subset\mathbb{R}^d$, i.e.:
\begin{flalign}\label{eq:dist_opt_obj}
   x_{\LL}^{\star}=\arg\min_{x\in\XX} f_{\mathcal{L}}(x),\hbox{ with }
   f_{\mathcal{L}}(x)=\frac{1}{|\LL|}\sum_{i\in\mathcal{L}}f_{i}(x).
     \end{flalign}
By choosing a local update rule and exchanging some information with their neighbors, the legitimate agents want to determine an optimal solution $x^{\star}$ to Eq.~\eqref{eq:dist_opt_obj}. In contrast, the malicious agents aim to either lead the legitimate agents to a common non-optimal value  $x$ such that $f_{\mathcal{L}}(x)>f_{\mathcal{L}}(x^{\star})$, or prevent the convergence of an optimization method employed by the legitimate agents. A common distributed optimization protocol for this problem takes the form:
\begin{flalign}\label{eq:legitimate_update_opt_gradient_selfish}
    &c_{i}(t)=w_{ii}(t) x_i(t)+ \sum_{j\in\mathcal{N}_i}w_{ij}(t)x_j(t),\nonumber\\
    &y_{i}(t)=c_i(t)-\gamma(t)\nabla f_i(c_i(t)),\nonumber\\
    &x_i(t+1) = \Pi_{\mathcal{X}}\left(y_i(t)\right),
    \end{flalign}
\an{where} $\Pi_{\mathcal{X}}$ is a projection operator onto the set $\mathcal{X}$, $c_i(t)$ is a consensus term, $y_i(t)$ is a gradient update, and $x_i(t+1)$ is the resulting updated iterate. For every legitimate agent $i\in\LL$, \an{the scalar} $\gamma_i(t)>0$ is the stepsize of agent $i$ at time $t$ and $\Ni$ is the set of neighbors of agent $i$ in the communication graph. The set $\Ni$ is composed of both legitimate and malicious neighbors of agent $i\in\LL$, while the weights $w_{ij}(t),j\in\Ni\cup\{i\},$ are nonnegative and sum to 1. 

% \begin{figure}
%   \centering
%   \vspace{-0.2in}
%     \includegraphics[scale=0.2]{distOptAlg.jpg}
%     \vspace{-0.1in}
%   \caption{Resilient distributed optimization algorithm.}
%   \label{alg:agent_i_dynamic}
% \end{figure}

% \begin{algorithm}[t!]
% \caption{Resilient distributed optimization for agent $i\in\LL$.}\label{alg:agent_i_dynamic}
% \begin{algorithmic}
% \State \textbf{Inputs:} $T$, $T_0$, $\NN_i$, $x_i(0)$, $\nabla f_i(\cdot)$, $\gamma(\cdot)$\\
% \an{\bf Angelia: the role of $T_0$ is unclear \my{We should change the notation of (11) to reflect that.}}
% \State \textbf{Outputs:} $x_i(T)$.
% \State Set $\beta_{ij}(t)=0$ for all $j\in\NN_i$
% \For{$t=0,\ldots,T-1$}
% \State Set $\NN_i(t) = \{j\in\NN_i:\:\beta_{ij}(t)\geq0\}$
% \State Set $d_i(t) = \max\{|\NN_i(t)|+1\}$
% \State Send $x_i(t)$ and $d_i(t)$ to neighbors
% \For{$j\in\NN_i$}
% \State Receive $x_j(t)$ and $d_j(t)$
% \State Extract $\alpha_{ij}(t)$
% \State Set  $\beta_{ij}(t+1)=\sum_{k=0}^{t} \left(\alpha_{ij}(k)-0.5\right)$
% \vspace{0.1cm}
% \State Set $w_{ij}(t)=\frac{\mathds{1}_{\{t\geq T_0\}}\mathds{1}_{\{\beta_{ij}(t)\geq 0\}}}{2\max\{d_i(t),d_j(t)\}}$ % \Comment{This is a comment}
% \EndFor
% \State Set $w_{ii}(t)=1-\sum_{m\in\NN_i}w_{im}(t)$
% \State Set $x_i(t+1)$ according to the dynamic \eqref{eq:legitimate_update_opt_gradient_selfish_trusted}
% \EndFor
% \end{algorithmic}
% \end{algorithm}

In~\cite{distOptCDC,distOpt_arXiv} it is shown that if the weights in the update Equation~\eqref{eq:legitimate_update_opt_gradient_selfish} are chosen using the stochastic observations of trust, then the following two key results hold. First, that \emph{convergence to the true optimum point $x_\LL^*$ holds in the almost sure sense and in the r-th mean for every $r\geq 1$} and secondly, that \emph{expected convergence rate can be established in the mean-squared sense where that rate is exponential in the observation time $T_0$.}

\begin{svblock}{Convergence of Resilient Distributed Optimization to the Optimal Value Almost Surely ~\cite{TRO_foundations}}
\an{For every legitimate agent $i\in\LL$, the iterates $x_{i}(t)$ generated by} the distributed optimization protocol in~\eqref{eq:legitimate_update_opt_gradient_selfish}, with weights defined using stochastic values of trust~\cite{TRO_foundations}, converge almost surely to $x_{\LL}^{\star}$ f\an{or every $T_0\geq 0$}.    
\end{svblock}

% \noindent \framebox[1.01\textwidth]{
% 	\parbox{\textwidth}{\ul{Convergence of Resilient Distributed Optimization to the Optimal Value Almost Surely (Theorem 2 in~\cite{distOptCDC}):} $x_{i}(t)$ converges a.s. to $\my{x_{\LL}^{\star}}$ for every $i\in\LL$ and $T_0\geq 0$.}
% }\\

%\begin{svblock}{Convergence of Resilient Distributed Optimization to the Optimal Value in Mean (Theorem 3 in~\cite{distOpt_arXiv})}
%For every $T_0\geq0$, $x_{i}(t)$ converges in the $r$-th mean to $x_{\LL}^{\star}$ for every $i\in\LL$ and $r\geq1$. That is, 
%\[\lim_{t\rightarrow\infty}\EEop\left[\|x_i(t)-%x_{\LL}^{\star}\|^r\right]=0,\]
%for all $r\geq1$.    
%\end{svblock}

% \noindent \framebox[1.01\textwidth]{
% \parbox{\textwidth}{\ul{Convergence of Resilient Distributed Optimization to the Optimal Value in Mean (Theorem 3 in~\cite{distOptCDC}):} For every $T_0\geq0$, \my{$x_{i}(t)$} converges in the \an{$r$-th} mean to $\my{x_{\LL}^{\star}}$ for every \my{$i\in\LL$ and} $r\geq1$. That is, 
% \[\lim_{t\rightarrow\infty}\EEop\left[\|x_i(t)-\my{x_{\LL}^{\star}}\|^r\right]=0,\]
% for all $r\geq1$.} 
% }\\
	
Notably, for the distributed optimization case, convergence to the nominal value with zero deviation is possible. Also derived in~\cite{distOptCDC,distOpt_arXiv} is a result on the convergence rate for the resilient distributed optimization problem that can be decreased by tuning the observation window $T_0$. 
%decreases exponentially with larger observation window $T_0$. 
This is presented as Theorem 5 in~\cite{distOpt_arXiv}.
% \begin{wrapfigure}[16]{r}{0.4\textwidth}
%   \centering
%   \vspace{-0.4in}
%     \includegraphics[scale=0.15]{CrowdVetting.jpg}
%     \vspace{-0.1in}
%   \caption{Opinion passing between neighbors using the CrowdVetting concept.}
%   \label{fig:crowdvetting}
% \end{wrapfigure}

\paragraph{Learning Trust}Finally, another important problem for resilient multi-agent systems is identifying trust of each agents' neighbors as quickly as possible. Previous results have shown that a subset of trusted neighbors' opinions can be leveraged to accelerate learning trust of the agents' extended network in a finite number of rounds of communication. This problem is called the problem of \emph{Crowd Vetting} and has been applied in the context of a Sybil Attack in~\cite{TRO_crowdvetting}. It was shown in~\cite{TRO_crowdvetting} that the Crowd Vetting Algorithm allows for agents to produce accurate opinions of trust of one another, with probability $1-\delta$ for $0<\delta<1$, within a finite number of rounds characterized by some $r^*>0$. Results for learning trust of neighboring agents over \emph{directed} graphs have been presented in~\cite{l4dc}.

We will conclude this subsection by noting that the use of additional side information from the network, \an{which } is used as inter-agent trust inputs, results in provably stronger resilience.\\

\noindent \emph{\textbf{Remark:} The availability of stochastic values of trust allows for convergence even in the case where \emph{more than half of the network connectivity is comprised of malicious links}. This is an improvement over the well-known classical resilience results for Byzantine systems~\cite{DOLEV198214}.}\\

In the next sections we will survey other multi-agent problems where trust plays an important role in attaining different performance guarantees in the resilient setting.

%Some relevant works to summarize here include~\cite{TRO_foundations, ICRA2019_switching, TRO_crowdvetting, distOptCDC}

\subsection{Reputation-based Interpretations of Trust and Related Performance Guarantees}\label{sec:rputation_based_application}

In this section we review the quantification of trust as a \emph{reputation-based scoring}. Several works, both in democratized sensing and democratized computing, have identified different ways of quantifying trust based on \emph{reputation}.  An agent's reputation is usually derived from a \emph{history of performance} of that agent as observed by other agents in the network. The concept of reputation-based trust has been used in several different problem settings from crowdsensing~\cite{STAR}, to federated learning~\cite{zeno,krum}, multi-agent coverage~\cite{schwagerAdaptiveSensingCoverage}, blockchain~\cite{blockchainPoor}, and multi-agent control~\cite{panagouTrustbasedCBF,ananduta2020resilient,maestre2021scenario}. In this section we will survey a few specific examples of reputation-based trust, namely \emph{social credit}, \emph{sensing quality}, \emph{performance scoring}, and associated performance guarantees for these types of trust metrics. 

\subsubsection{Definitions of Reputation based on Data and Sensing}
We divide different metrics of reputation by whether they are \emph{data-based} meaning that they use a history of data or values transmitted by a particular agent to assess trust, or \emph{sensor-based} meaning that they are based on behavioral observations of the agent often obtained through physical sensing such as with cameras. 

\paragraph{Blockchain and Crowdsensing} Several works have explored the option of using a collected history of transmitted data values for each neighboring agent and comparing those values against internal models of nominal behavior in order to establish measures of trust. For example in the case of utility-optimal crowdsensing,~\cite{STAR} introduces the concept of \emph{social trust assisted reciprocity} (STAR) and \emph{social credit}. In the case of computing in edge networks, blockchain-based trust-mechanisms have been employed to choose the miner of the blockchain and ultimately suppress selfish edge attacks~\cite{blockchainPoor}. For example the paper in~\cite{blockchainPoor} uses Bayesian inference from~\cite{vehTrust} to compute a service reputation:

\begin{svblock}{Service Reputation~\cite{blockchainPoor}}
Each edge device in a network is assigned a true service reputation $\tilde{R}_j$ based on the record credibility $c_{ij}\in\{0,1\}$, and the individual service record $R_{ij}$, where $\tilde{R}_j$ is computed as
	\begin{align}
	    \tilde{R}_j=\sum_{i=1}^Mc_{ij}R_{ij},
	\end{align}
	\an{where} the individual service record $R_{ij}$ is computed using Bayesian inference according to~\cite{vehTrust} in a way that captures the credibility of a agent based on a history of their truthful reporting reliability (cf.\ Eq.(2) in~\cite{vehTrust}).    
\end{svblock}

% \noindent \framebox[1.01\textwidth]{
% 	\parbox{\textwidth}{\underline{Service Reputation~\cite{blockchainPoor}:} Each edge device in a network is assigned a true service reputation $\tilde{R}_j$ based on the record credibility $c_{ij}\in\{0,1\}$, and the individual service record $R_{ij}$, where $\tilde{R}_j$ is computed as
% 	\begin{align}
% 	    \tilde{R}_j=\sum_{i=1}^Mc_{ij}R_{ij}
% 	\end{align}
% 	and the individual service record is computed using Bayesian inference according to~\cite{vehTrust} in a way that captures the credibility of a agent based on their truthful historical reporting reliability (cf. Eqn.(2) in~\cite{vehTrust}}.
% }\\ 

This service reputation is then included as part of the state, which also captures the volume of data to be processed, the offloading rate of the mobile device $i$ to edge device $j$, and a record of the payment received at the last iteration. By defining a state to include a history of service reputations for each edge device $1\leq j\leq N$, the work in~\cite{blockchainPoor} designs a reinforcement learning-based edge CPU allocation algorithm to optimize the edge CPU resource allocation policy. A novel performance guarantee for this scheme is derived as Theorem~2 in~\cite{blockchainPoor} whereby the edge device with the best channel condition is able to optimize its utility after a sufficient number of iterations.

\paragraph{Machine Learning}
For the case of machine learning, rankings based on whether an agent's behavior is suspicious according to nominal behavior models have been employed to reject values reported by certain nodes that are deemed to be suspicious or faulty~\cite{zeno,krum}. For example, the paper~\cite{zeno} considers the optimization problem of
\begin{align}
\label{eq:MLProb}
    \min_{x\in\mathbb{R}^d}F(x),
\end{align}
where $F(x)=\mathbb{E}_{z\sim\mathcal{D}}[f(x,z)]$ and $z$ is sampled from the unknown distribution $\mathcal{D}$, and $d$ is the dimension.  The objective here is to find the minimizer $x^*$ of $F(x)$, which is assumed to exist, by using a distributed optimization method where each agent (or worker) provides a gradient that can be combined to find the next iterate $x(t+1)$: 
\begin{align}
    x(t+1)=x(t)-\gamma(t)\text{Aggr}(\{g_i(x(t)):i\in[m]\}),
\end{align}
where Aggr$(\cdot)$ is an aggregation rule such as the average, and $m$ is the number of workers. \an{The gradient $g_i(\cdot)$} is of the local empirical loss $F_i(x)=\frac{1}{n}\sum_{j=1}^n f(x;z^{i,j}), \forall i\in [m]$ where $z^{i,j}$ is the $j$th sampled data on the $i$th agent, and $n$ is the number of sampled i.i.d. data points from the distribution $\mathcal{D}$. Using some assumptions about the expected nature of the gradients, namely that 1) the gradient should improve the loss function and 2) the magnitude of the update should not be too large, the authors of~\cite{zeno} suggest the incorporation of a \emph{score metric}. Suggested updates from untrusted agents that have low scores are then eliminated when computing the new aggregate gradient that will be used to update the value of agent $i$'s iterate. Under the assumption that there are $q$ faulty workers computing gradient estimators, and where $q<m$ for a total number of $m$ workers, the authors in~\cite{zeno} provide a convergence guarantee for the problem in~\eqref{eq:MLProb} when the Score Metric described above is used to eliminate some agents from the update iteration. A central claim to this class of works is that by incorporating the reputation of different agents contributing to a democratized computing task, performance guarantees can still be met even with a higher number (sometimes arbitrarily so) of malicious agents than legitimate agents.

% \noindent \framebox[1.01\textwidth]{
% 	\parbox{\textwidth}{\underline{Score Metric~\cite{zeno}:} A score metric using each agent's reported gradient can be computed as
% 	\begin{align}
% 	    Score_{\gamma,\rho}(u,x)=f_r(x)-f_r(x-\gamma u)-\rho\|u\|^2
% 	\end{align}
% 	Where $f_r(x)=\frac{1}{n_r}\sum_{i=1}^n_rf(x;z_i)$ where the $z_i$'s are i.i.d. samples drawn from $\mathcal{D}$ and $n_r$ is the batch size of $f_r(\cdot)$ and $\mathbb{E}[f_r(x)]=F(x)$. Here $u$ is the update (gradient estimator), $x$ is the current parameter, $\gamma$ is the learning rate, and $\rho>0$ is a constant weight.}
% }\\ 

\paragraph{Distributed Model Predictive Control}

The problem of distributed Model Predictive Control with adversarial attacks has also been shown to benefit from incorporating metrics of trust for arriving at more robust performance~\cite{ananduta2020resilient, maestre2021scenario}. In this problem a system state and control input for agent $i$ is given as $x_i$ and $u_i$ respectively, $x=[x_i]_{i\in[1,M]},\mathbf{u}=[\mathbf{u}_i]_{i\in[1,M]}$ are vectors of all agent states and control inputs, $M$ is the total number of agents, $\mathbf{u}_i=[u_i^T(0),u_i^T(1),\hdots,u_i^T(N_h-1)]^T$ is the sequence of global inputs for a prediction horizon of $N_h$ time steps, and a cost defined over the global state and input trajectories is defined as~\cite{DMPC_Maestre}:
\begin{align}
    \mathbf{J}(x,\mathbf{u})=\sum_{i\in[1,M]}\sum_{n=0}^{N_h-1}\left(\|x_i(n)-x_{\text{ref},i}\|^2_{Q_i}+\|u_i(n)\|^2_{R_i}\right),
\end{align}

\noindent where $x_{\text{ref},i}$ is the state reference for agent $i$, and $Q_i\geq 0$ and $R_i>0$ are weighting matrices.  The centralized Model Predictive Control (MPC) problem is given by the following:
\begin{align}
    &\min_\mathbf{u} J(x,\mathbf{u})\\
    &\text{s.t.}\ C_\text{in}\mathbf{u}\leq c_\text{in}\\
    &\hspace{0.6cm}C_\text{eq}\mathbf{u}=c_\text{eq},
\end{align}

\noindent where $C_\text{in}$, $C_\text{eq}$, $c_\text{in}$, $c_\text{eq}$ are matrices and vectors respectively, defining affine constraints on the optimization variable $\mathbf{u}$. Distributed versions of MPC attain the optimal centralized MPC solution through iterative methods, some of which are discussed in~\cite{DMPC_Maestre} for example. Works in this domain have explored the use of a trust mechanism that indicate agents with a higher probability of being attackers based on a constructed identification mechanism  that learns over time which agents are likely to be attackers. One example of such a trust mechanism is derived in~\cite{ananduta2020resilient}. Here, the trust mechanism derives from a Bayesian hypothesis testing approach where several different hypotheses are constructed to cover different behaviors stemming from insider attacks. A prior is applied that assigns uniform probability for all hypotheses ranging from no malicious neighbors, to one attacker, to all combinations where there are $>$1 attackers. For these cases a probability of attack is computed. The reference~\cite{maestre2021scenario} includes an example of computing such a probability of attack for a small five agent network.

% \begin{svblock}{Probability of Attack using Bayesian Methods~\cite{ananduta2020resilient}}
%     Assuming that the attack rate of malicious agents is given as $P_a$ and each agent communicates with a random subset of neighbors $S(k)$ at time $k$ where this subset has a number $s_m(k)$ of malicious agents in $S(k)$, then the probability of an attack can be computed as:
% 	\begin{align}\label{probAttackMPC}
% 	    P_S(k)=1-(1-P_a)^{s_m(k)}.
% 	\end{align}
% \end{svblock}

% \noindent \framebox[1.01\textwidth]{
% 	\parbox{\textwidth}{\underline{Probability of Attack using Bayesian Methods~\cite{ananduta2020resilient}:\label{probAttackMPC}} Assuming that the attack rate of malicious agents is given as $P_a$ and each agent communicates with a random subset of neighbors $S(k)$ at time $k$ where this subset has a number $s_m(k)$ of malicious agents in $S(k)$, then the probability of an attack can be computed as:
% 	\begin{align}
% 	    P_S(k)=1-(1-P_a)^{s_m(k)}
% 	\end{align}}
% }\\
From here the Model Predictive Control optimization problem of each agent uses the probability that its' neighbors are attackers based on the Probability of Attack using Bayesian Methods described above. The idea is that each agent learns over time which are the attackers in the system and excludes them from the optimization problem.

\paragraph{Coverage and Sensing}
The problem of coverage and distributed sensing is another important application of multi-agent networks where obtaining and integrating values of trust in the decision-making processes of the agents has resulted in resilient performance guarantees for these systems. The coverage problem concerns the distributed deployment of agents to different physical locations across an environment of interest.  It is assumed that the agents will be monitoring or performing some sensing task to cover \an{different regions} of the environment and, thus, this problem is often referred to as \emph{coverage}~\cite{bulloCoverage,schwager2017,pavoneRiskCoverage}. Assuming that the environment of interest can be expressed as a bounded, convex region $\mathcal{Q}\subset\mathbb{R}^2$, where points in $\mathcal{Q}$ are denoted by $q$ and positions of $n$ individual agents are denoted $p_i\in\mathcal{Q}$, $i\in\{1,\hdots,n\}$, the coverage problem seeks to find a positioning of the $n$ agents such that they minimize (in a local sense) the following cost function:
\begin{align}
\mathcal{H}_V(p_1,\hdots,p_n)=\sum_{i=1}^n \int_{V_i} \frac{1}{2}\|q-p_i\|^2\phi(q)dq,
    \label{eq:coverage}
\end{align}

\noindent where $V_i$ corresponds to the Voronoi cell defined over the environment that agent $i$ is responsible for covering
\begin{align}
    V_i=\{q\in\mathcal{Q}|\|q-p_i\|\leq \|q-p_j\|,\ \forall j\neq i\}.
\end{align}

\noindent \an{A} weighted version of this Voronoi cell can be described as $W_i$ where
\begin{align}
    W_i=\{q\in\mathcal{Q}|\|q-p_i\|^2-w_i\leq \|q-p_j\|^2-w_j,\ \forall j\neq i\}.
\end{align}

The authors in~\cite{schwagerAdaptiveSensingCoverage} derive a new coverage control law where differences in agent performance can be reflected in the converged coverage positions of the agents.  The differences in agent performance can arise from variations in sensing quality for example due to general sensor ``health.'' An agents' health value is represented by a scalar $h_i\in\mathbb{R}$, and this can be thought of as a trust variable wherein agents that have better performance can be trusted to cover a larger footprint within the environment $\mathcal{Q}$. Those agents that are less trusted should cover a proportionally smaller footprint.  To capture this, a measure of trust based on sensor health is presented. The authors of~\cite{schwagerAdaptiveSensingCoverage} show that by incorporating this performance-based trust metric, the agents will each converge to the center of their weighted Voronoi cell and that the difference in weights across partitions reflects the difference in ``sensor health'' between agents which is a key element of the inter-agent performance metric.
%\mynote{Here we can tie back to the integrated attack we described in the last part of the attack section and add the last part of the figure that originally appeared in your paper, Stephanie.}

% \noindent \framebox[1.01\textwidth]{
% 	\parbox{\textwidth}{\underline{Performance-based Trust~\cite{schwagerAdaptiveSensingCoverage}:} A measure of trust based on each agent $i$'s sensing performance is introduced in~\cite{schwagerAdaptiveSensingCoverage} as
% 	\begin{align}
% 	    \gamma_i(p_i,q,h_i)=-\alpha(\|q-p_i\|^2-h_i)
% 	\end{align}
% 	where $h_i\in\mathbb{R}$ is the sensor performance health for agent $i$, and $\alpha$ is a scaling factor.
% 	}
% }\\

\begin{svblock}{Convergence of Performance-Weighted Coverage}
By using a control law that incorporates trust based on sensor health $h_i$ for each agent $i$, agents converge to the centroids of their weighted Voronoi cells, and the relative weights of their Voronoi partitions reflect the relative differences in the agents' performance metrics such that $(w_i-w_j)\rightarrow(h_i-h_j)\ \forall i,j$~\cite{schwagerAdaptiveSensingCoverage}.
\end{svblock}

% \noindent \framebox[1.01\textwidth]{
% 	\parbox{\textwidth}{\underline{Convergence of Performance-Weighted Coverage[Theorem 1 in~\cite{schwagerAdaptiveSensingCoverage}]}: By using the control law in Eq.~\eqref{eq:CoverageControl} with performance-based weights as dictated by Eq.~\eqref{eq:covWeights}, agents converge to the centroids of their weighted Voronoi cells and the relative weights of their Voronoi partitions reflect the relative differences in the agents' performance metrics such that $(w_i-w_j)\rightarrow(h_i-h_j)\ \forall i,j$.} 
% }\\

In other words, the proportion of the environment that each agent is responsible for monitoring is directly related to the amount of performance-based trust assigned to that agent. In this way, the coverage problem has automatically adapted to variations in performance and trust of agents in the network in contrast to the original coverage problem that does not account for inter-agent trust.

%In this section we will summarize works in crowdsensing, coverage, and other ``democratized sensing'' tasks where agents share information or sensed quantities to arrive at an accurate global decision or estimation of a value of interest. For example the work in~\cite{STAR} quantifies the use of \emph{social credit} which leads to provably maximal utility optimization. Other papers to cite here include~\cite{schwager2017} among others.

%\subsection{Democratized Computing}
%In these cases a metric of \emph{reputation} are derived and exploited for capturing trust. Some application areas include MEC's and blockchain as well as federated learning. Some citations for this section include~\cite{blockchainPoor} among others.

\subsection{Network Measures}

An additional  class of metrics that have been used to arrive at novel resilience guarantees regarding notions of network connectivity. Many works have derived different characterizable types of resilience that can be attained over networks, categorized by different notions of connectivity. Examples are the works in~\cite{SundaramWMSR,controlTheoretic_Security_Bullo,networkResilience_Cortes,RSS_CBF}. We note that here, no explicit trust values or external reputation scores for agents are used, but rather graph properties, such as a minimum connectivity level, provide resilience to the underlying multi-agent tasks.  The applications that we will review here deal with multi-agent consensus. Although, we note that many of the concepts of connectivity discussed here that result in resilient consensus also generalize to other resilient versions of distributed sensing and distributed computation tasks such as adversarial hypothesis testing and others~\cite{Sundaram_Survey_resilience_graph_2022,SundaramHypTesting}.\\

\paragraph{Resilient Consensus from a System Theoretic Perspective}

The importance of the role of graph connectivity for attaining resilient consensus has been studied for many decades and many known results relating connectivity and resilient consensus date back to the seminal papers on Byzantine agreement problems~\cite{DOLEV198214, Lamport_agreement_faults_1980}. The authors in~\cite{controlTheoretic_Security_Bullo} take a system theoretic approach to studying the consensus problem that is stated as follows. Let $n,m,p \in \mathbb{N}$, let $A\in\mathbb{R}^{n\times n}, B\in\mathbb{R}^{n\times m},$ and $C\in\mathbb{R}^{p\times n}$. Then the triple $(A,B,C)$ can be used to describe a linear discrete time system of the form:
\begin{align}
\label{eq:sysTheoreticConsensus}
    x(t+1)&=Ax(t)+Bu(t),\\
    y(t)&=Cx(t).
\end{align}

The authors in~\cite{bulloUnreliable} show that the measure of \emph{$k$-connectivity} of the graph is a key metric for \emph{detecting} and \emph{identifying} $k-1$ misbehaving agents in the system.  Here, misbehaving agents are defined as any agent $j$ that provides a non-zero control input to the system, i.e. where $u_j(t)\neq 0$ for some $t\in\mathbb{N}$ in Eq.~\eqref{eq:sysTheoreticConsensus} (c.f. Definition~1 in~\cite{bulloUnreliable}). The consensus system with misbehaving agents represented by the set $K$, $K=\{i_1,i_2,\hdots\}\subseteq \mathbb{V}$ and $B_K=[e_{i_1}\ e_{i_2},\ldots]$, where $e_i$ is the $i$th vector of the canonical basis, is:
\begin{align}
    x(t+1)=Ax(t)+B_Ku_K(t).
    %\text{\my{Should the first plus sign be an equality one?}\sg{yes! Thank you}}
\end{align}

Here, the authors introduce the concept of \emph{identifiability} of misbehaving agents to be the \emph{possibility of distinguishing the misbehaviors} (inputs) \emph{of two distinct agents, or between two disjoint subsets of agents.} The $k$-connectivity of a graph plays a central role here, where a graph $\mathbb{G}$ is $k$-connected if the removal of at most $k$ vertices of $\mathbb{G}$ does not disconnect it.

% \noindent \framebox[1.01\textwidth]{
% 	\parbox{\textwidth}{\underline{$k$-connectivity of a graph $\mathbb{G}$}: A graph $\mathbb{G}$ is $k$-connected if the removal of at most $k$ vertices of $\mathbb{G}$ does not disconnect it.} 
% }\\

Given a $k$-connected network, under the control system theoretic approach offered in~\cite{bulloUnreliable}, it is proven that there exist identifiable inputs for every set of $k-1$ misbehaving agents.

\begin{svblock}{Generic identification of misbehaving agents (\an{Theorem~6.2} in~\cite{bulloUnreliable})}
Generically, there exist identifiable inputs for every set of $k-1$ misbehaving agents in $\mathbb{G}$.
\end{svblock}

% \noindent \framebox[1.01\textwidth]{
% 	\parbox{\textwidth}{\underline{Generic identification of misbehaving agents (Thm 6.2 in~\cite{bulloUnreliable})}: Generically, there exist identifiable inputs for every set of $k-1$ misbehaving agents in $\mathbb{G}$.} 
% }\\

Thus, it has been shown that the connectivity of the graph has important provable implications for the detection and identification of misbehaving agents in a multi-agent network performing consensus.  Another graph property that was shown to be important for achieving approximate consensus using only local information is that of \emph{network robustness}.\\

\paragraph{Network Robustness and Consensus using only Local Information}
Here, we briefly summarize the notion of \emph{network robustness} presented by the authors in~\cite{SundaramWMSR} for achieving \emph{approximate agreement}, or agreement to within a relatively small convex set contained in the convex hull of the initial values of legitimate agents. The authors present an algorithm called the \emph{Weighted Mean-Subsequence-Reduced} (W-MSR) algorithm where agents remove extreme (outlier) values with respect to its own value at each timestep before performing its individual update according to the standard consensus protocol.  For this algorithm the authors show that $k$-connectivty is no longer sufficient to ensure convergence. They address a new notion of network robustness called \emph{$(r,s)$-robustness} (\an{Definition~13} in~\cite{SundaramWMSR}). Intuitively, this notion of network robustness captures the notion that enough nodes in every pair of nonempty, disjoint sets $\mathcal{S}_1,\mathcal{S}_2\subset\mathbb{V}$ will have at least $r$ neighbors outside of their respective sets~\cite{SundaramWMSR}. Further, it is shown that this metric of network robustness leads to a convergent linear consensus protocol, in the asymptotic agreement sense, when a local outlier rejection rule is employed to remove untrusted agents before each agent updates their iterates. Similar metrics of robustness are employed to arrive at resilience in other multi-agent problems such as adversarial hypothesis testing~\cite{SundaramHypTesting,Sundaram_Survey_resilience_graph_2022}.

%%% SG Note: I am not sure if the commented text below fits in with the discussion on trust-based measures. Perhaps if this work were to show how to choose the \lambda based on inter-agent trust it would fit better. Perhaps we can discuss if it would fit here as is or what section it might fit better under?

%\subsection{\my{Competition-Based Measures}} 
%\my{An additional method to mitigate the presence of untrustworthy agents is carried out via the competition-based approach  proposed in \cite{competition_based_resilient_opt}. Here,  each legitimate agent $i$ utilizes the following Friedrick-Johnsen (FJ) dynamic to update its value:
%\begin{flalign}
%    x_i(t+1)=\lambda_ix_i(0)+(1-\lambda_i)\sum_{}w_{ij}x_j(t).
%\end{flalign}
%Optimizing the values $\lambda_i$ creates a competition between agent's initial value and that of the collaborative dynamic of a consensus system. In a reliable system
%$\lambda_i=0$ and agents best conduct is to fully trust their neighbors. In a less trustworthy setups, an agent may prefer to act selfishly in pursuit of avoiding being misled by malicious agents in the network. The 
%work \cite{competition_based_resilient_opt} gives insight on how to optimize the choice of $\lambda_i$ to minimize the expected mean squared consensus
%error where the randomness is taken with respect to the legitimate agents' initial values/opinions and the malicious agents inputs.}  

\subsection{Combining Trust Values and Network Topology}
Lastly, we consider characterizations of graph topologies that have been shown to play an important role for arriving at resilience performance guarantees when stochastic observations of trust are also available. The quantity \emph{$\tau$-triangularity} characterizes the number of shared neighbors for a pair of agents $i$ and $j$ and becomes an important characterization for learning trust of neighboring agents. Particularly, \an{
let $l_{ij}=|\mathcal{N}_i\cap\mathcal{N}_j\cap \LL |$. Then, a graph $\mathbb{G}$  is $\tau$-triangular if 
there is a $\tau>0$ such that $l_{ij}\geq \tau$ for all $i\in\LL$ and $j\in\mathbb{V}$.
}

%then $\mathbb{G}$ is $\tau$-triangular.    
% \noindent \framebox[1.01\textwidth]{
% 	\parbox{\textwidth}{\underline{$\tau$-triangular Networks:} Let $l_{ij}=|\mathcal{N}_i\cap\mathcal{N}_j\cap \LL |$. Then for a given graph $\mathbb{G}$ if $l_{ij}\geq \tau$ for all $i\in\LL$ and $j\in\mathbb{V}$ then $\mathbb{G}$ is $\tau$-triangular.} 
% }\\

% Figure~\ref{fig:tauTriangularity} demonstrates the concept of $\tau$-triangularity for a subgraph formed by the neighborhood around two agents.
% \begin{wrapfigure}[8]{r}{0.5\textwidth}
%   \centering
%   \vspace{-0.2in}
%     \includegraphics[scale=0.1]{tauSim.jpg}
%     \vspace{-0.1in}
%   \caption{Decreasing number of communication rounds with increasing $\tau$-triangularity of the network.}
%   \label{fig:tauSim}
% \end{wrapfigure}
% \begin{wrapfigure}[8]{r}{0.5\textwidth}
%   \centering
%   \vspace{-0.2in}
%     \includegraphics[scale=0.08]{tauTriangularity.jpg}
%     \vspace{-0.3in}
%   \caption{Concept of $\tau$-triangularity of a network.}
%   \label{fig:tauTriangularity}
% \end{wrapfigure}
%In these works different notions of network connectivity are derived that lead to different characterizable types of resilience (for example, observability, controllability). Some works to cite here include~\cite{SundaramWMSR,controlTheoretic_Security_Bullo,networkResilience_Cortes}.
An interesting inverse relationship exists between the $\tau$-triangularity of the network and the number of communication rounds needed to accurately identify malicious neighbors in the system. A simulation study on this relationship can be found in~\cite{TRO_crowdvetting}. 

\section{Open Problems and Concluding Remarks}\label{sec:open_prob_conclusion}
In the future, wireless and cellular networks will face greater demands in terms of security and will need to accommodate a significantly wider range of applications, user devices, and user heterogeneity compared to previous networks. This can be attributed to the diverse nature of the applications and the composition of these emerging networks. Unlike in the past, where personal device (cellphone and laptop) users comprised the majority of network users, the future will introduce highly varied physical networks consisting of drones, smart and connected vehicles, IoT devices, and more. These networks will require high data rates, exceptional energy efficiency, low latency, and robust security.

Furthermore, in future networks, malicious agents will extend beyond traditional attacks like eavesdropping, denial-of-service, and link-dropping. Instead, they will attempt to disrupt the network's functionality and compromise the performance of various tasks, such as Democratized Sensing and Democratized Computing tasks that are executed over the network. This survey paper highlights numerous threats, both new and classical, that are particularly relevant to this category of emerging cyber-physical networks. Additionally, it presents various new opportunities for securing these systems. These capitalize on anticipated features and capabilities of future 6G networks, that revolve around the physicality of these networks. Such opportunities include leveraging innovative sensing techniques, channel engineering, and AI-powered situational awareness, which will be available for the first time in the next generation of networks.

The pursuit of this vision requires progress in multiple research communities on several open questions that need to be addressed. The first open question is around finding more \emph{characterizable concepts of ``trust''} in multiagent networks. This will likely require employing tools from wireless networking and signal processing, and should build upon the new capabilities of sensing, perception, and channel engineering that are poised to be a critical part of future-generation networks. This may also require engineering cyberphysical multiagent systems to support, extract, and exploit trust values. Section~\ref{sec:network} of this survey paper outlines some important advances along this direction, however, more is needed to support the wide array of cyberphysical systems and applications on the horizon. A second open question is that of \emph{designing novel distributed multiagent algorithms that integrate and exploit trust values to create resilient algorithms} with provable resilience guarantees. While Section~\ref{sec:performance} outlines some important advances in this direction, more such results will be needed before these systems reach their potential as a core component of future societal infrastructure. Lastly, the third open question will support the first two and this is \emph{the introduction of a shared nomenclature to reduce barriers across the fields of communication, robotics, and control}. This will be necessary in order to make progress across field lines and bring to fruition powerful new concepts of trust that close the loop between the network and performance of the networked multiagent systems. We note in passing that an interesting question is whether machine learning can be brought to bear on these problems, harnessing the large volumes of data that are often collected for different cyberphysical networks.

We conclude this survey article by noting that the physicality of emerging cyberphysical networks can play a pivotal role in addressing the aforementioned open questions. This survey paper captures the concept that the physicality of future multi-agent networks will play a prominent role in the structure of these networks, the establishment of trust between agents, and the attainment of quantifiable performance guarantees for these systems. In contrast to today's networks, where the application layer is largely decoupled from the lower communication layers, performance guarantees for emerging cyberphysical networks should assume a central role in redefining security for these systems. This is because simple authentication alone is insufficient to ensure security; the behavior of agents must also be considered when making this determination. This calls for a stronger connection between the application layer and lower layers than ever before. From this perspective, the physicality of emerging networks shows promise for a new era of physics-based trust in these systems, enabling us to surpass classical limitations on resilience for distributed systems and traditional notions of security. In a sense, this calls for a new field of \emph{resilient multiagent networks} that is rooted in \emph{novel and characterizable concepts of trust} -- thus paving the way to achieving verifiable and practical resilience in the new age of cyberphysical systems. 

%\newpage
\bibliographystyle{IEEEtran}
\bibliography{survey_references}

\newpage 
\section{Biography Section}
\begin{IEEEbiography}
[{\includegraphics[width=1in,height=1.25in,clip,keepaspectratio]{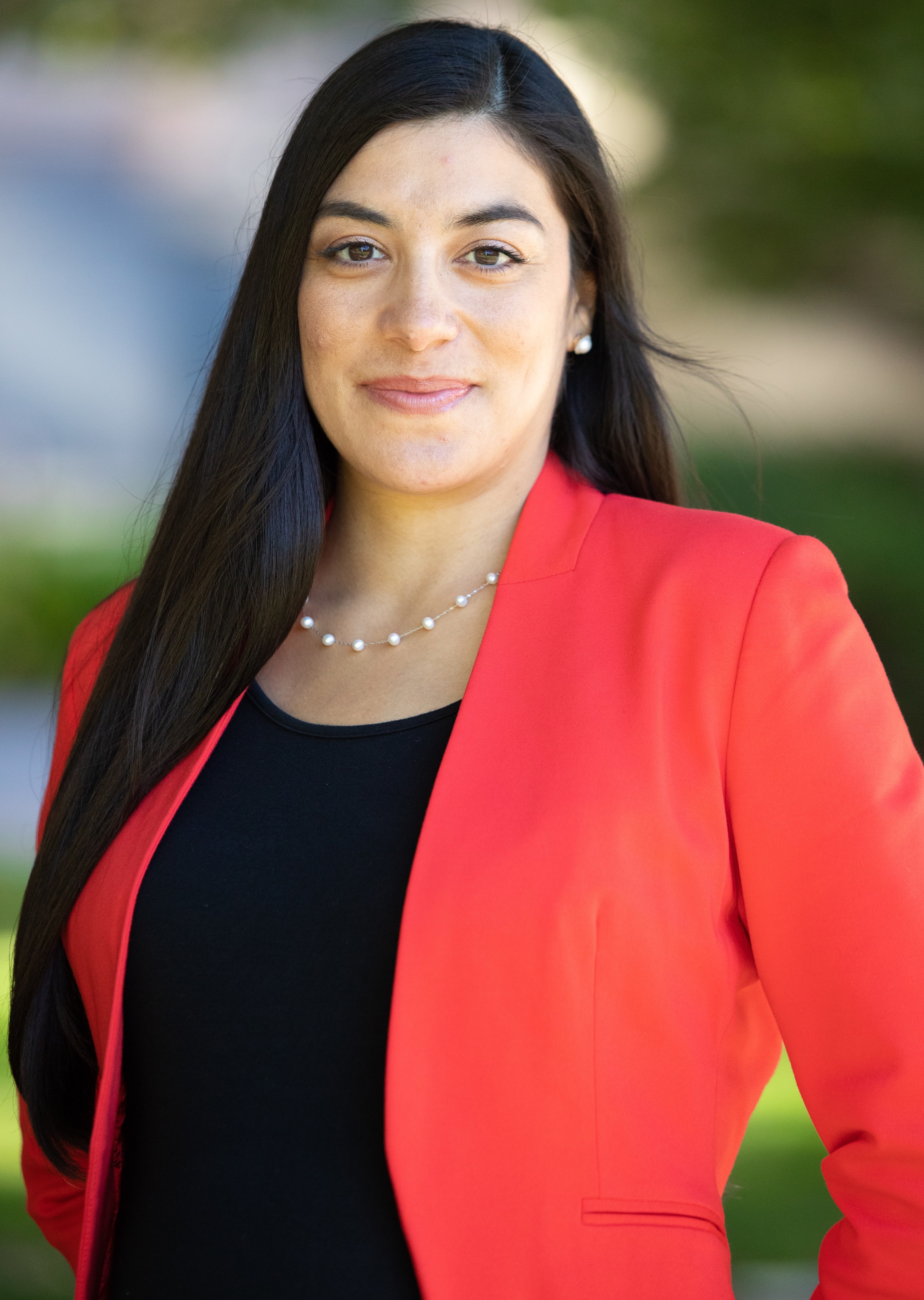}}]{Stephanie Gil}
is an Assistant Professor in the area of Computer Science at the School of Engineering and Applied Sciences (SEAS) at Harvard University where she directs the Robotics, Embedded Autonomy and Communication Theory (REACT) Lab. Prior, she has held a Visiting Assistant Professor position at Stanford University during the summer of 2019, and an Assistant Professorship at Arizona State University from 2018-2020. She completed her Ph.D. work (2014) and her M.S. work (2009) at the Massachusetts Institute of Technology (MIT), and she completed her B.S. at Cornell University in 2006. Her research focuses on multi-robot systems where she studies the impact of information exchange and communication on resilience and trusted coordination. She holds U.S. patents in adaptive heterogeneous networks for multi-robot systems and accurate indoor positioning using wireless signals. Her contributions have been recognized by several awards including the National Science Foundation CAREER award (2019), the Office of Naval Research Young Investigator award (2021), and she has also been selected as a 2020 Sloan Research Fellow for her work at the intersection of robotics and communication.
\end{IEEEbiography}
\begin{IEEEbiography}[{\includegraphics[width=1in,height=1.25in,clip,keepaspectratio]{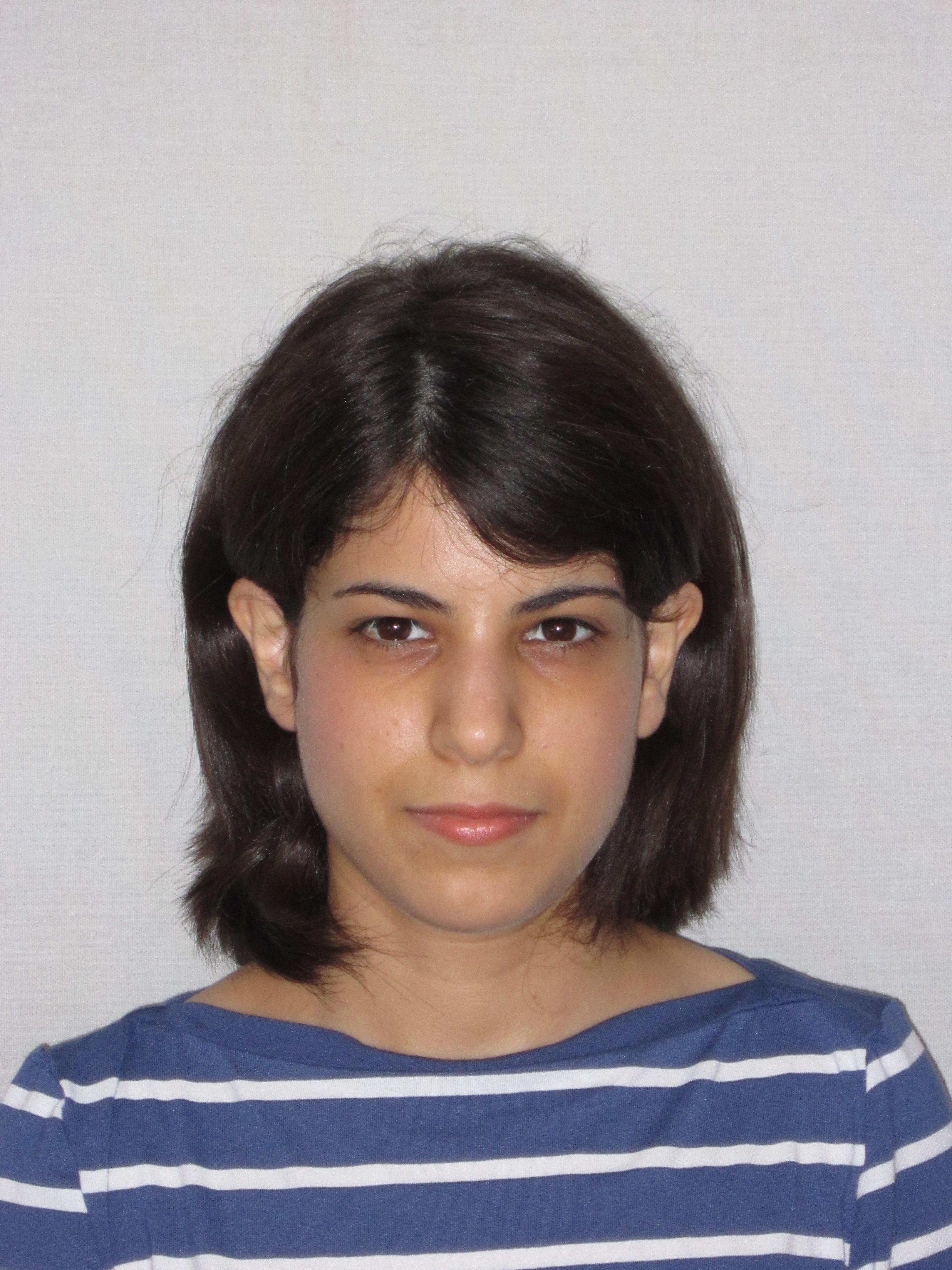}}]{Michal Yemini}
 is an assistant professor at Bar-Ilan University, Ramat-Gan, Israel. Prior to that, she was an associate research scholar at Princeton University, a postdoctoral researcher at Stanford University, Stanford, USA, and a visiting postdoctoral researcher at Princeton University. Her main research interests include distributed optimization, sequential decision-making, learning theory, information theory, and percolation theory.
She received the Eric and Wendy Schmidt Postdoctoral Award for Women in Mathematical and Computing Sciences, the Council of Higher Education's Postdoctoral Fellowships Program for Outstanding Women in Science, and the Bar-Ilan University's Postdoctoral Fellowship for Women. She obtained her BSc in computer engineering from the Technion-Israel Institute of Technology, Haifa, Israel, in 2011. In 2017 she received her PhD degree in the joint MSc-PhD program from the Faculty of Engineering, Bar-Ilan University, Ramat-Gan, Israel.
\end{IEEEbiography}
\begin{IEEEbiography}[{\includegraphics[width=1in,height=1.25in,clip,keepaspectratio]{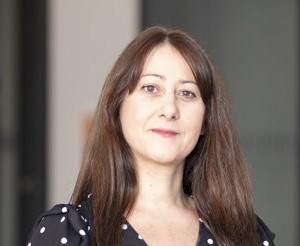}}]{Arsenia (Ersi) Chorti}
 is a Professor at the École Nationale Supérieure de l'Électronique et de ses Applications (ENSEA) at the ETIS Lab UMR 8051, Research Fellow of the Barkhausen Institut gGmbH and a Visiting Scholar at Princeton University. Her research spans the areas of wireless communications and wireless systems security for 5G and 6G, with a particular focus on physical layer security. Current research topics include : context aware security, multi-factor authentication protocols, 5G / 6G and IoT, anomaly detection, machine learning for communications, new multiple access techniques and scheduling. She is a Senior IEEE Member, member of the IEEE INGR on Security, Chair of the IEEE Focus Group on Physical Layer Security and a Member of various ITU Working Groups including on Security of the Metaverse. She has participated in the reduction of the ITU report M.2516-0 on Future technology trends of terrestrial International Mobile Telecommunications systems towards 2030 and beyond (sections on trustworthiness).
\end{IEEEbiography}

\begin{IEEEbiography}[{\includegraphics[width=1in,height=1.25in,clip,keepaspectratio]{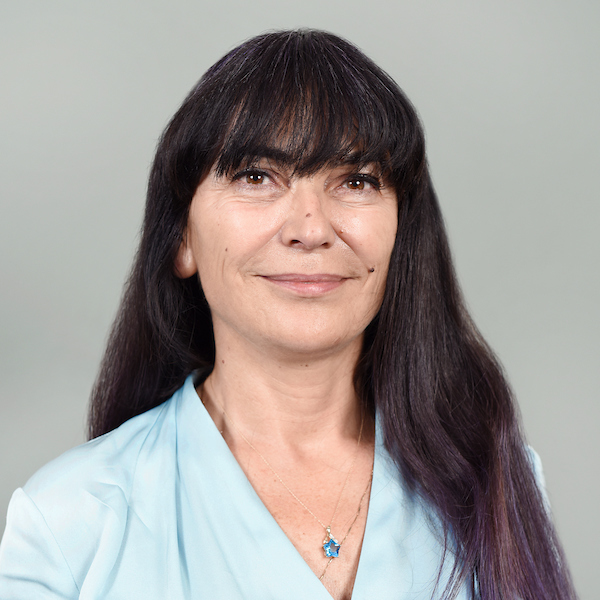}}]{Angelia Nedi\'c}
holds a Ph.D.\ from Moscow State University, Moscow, Russia, in Computational Mathematics and Mathematical Physics (1994), and a Ph.D.\ from Massachusetts Institute of Technology, Cambridge, USA in Electrical and Computer Science Engineering (2002). She has worked as a senior engineer in BAE Systems North America, Advanced Information Technology Division at Burlington, MA. Currently, she is a professor at the school of Electrical, Computer and Energy Engineering at Arizona State University at Tempe. Prior to joining Arizona State University, she has been a Willard Scholar faculty member at the University of Illinois at Urbana-Champaign. She is a recipient (jointly with her co-authors) of the Best Paper Award at the Winter Simulation Conference 2013 and the Best Paper Award at the International Symposium on Modeling and Optimization in Mobile, Ad Hoc and Wireless Networks (WiOpt) 2015.  Her general research interest is in optimization, large scale complex systems dynamics, variational inequalities and games. 
\end{IEEEbiography}
\vspace{-1in}
\begin{IEEEbiography}[{\includegraphics[width=1in,height=1.25in,clip,keepaspectratio]{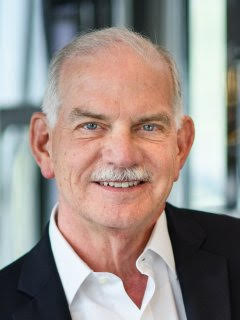}}]{H. Vincent Poor} (S’72, M’77, SM’82, F’87) received the Ph.D. degree in EECS from Princeton University in 1977.  From 1977 until 1990, he was on the faculty of the University of Illinois at Urbana-Champaign. Since 1990 he has been on the faculty at Princeton, where he is currently the Michael Henry Strater University Professor. During 2006 to 2016, he served as the dean of Princeton’s School of Engineering and Applied Science. He has also held visiting appointments at several other universities, including most recently at Berkeley and Cambridge. His research interests are in the areas of information theory, machine learning and network science, and their applications in wireless networks, energy systems and related fields. Among his publications in these areas is the recent book Machine Learning and Wireless Communications.  (Cambridge University Press, 2022). Dr. Poor is a member of the National Academy of Engineering and the National Academy of Sciences and is a foreign member of the Chinese Academy of Sciences, the Royal Society, and other national and international academies. He received the IEEE Alexander Graham Bell Medal in 2017. 
\end{IEEEbiography}

\vspace{-1in}
\begin{IEEEbiography}[{\includegraphics[width=1in,height=1.25in,clip,keepaspectratio]{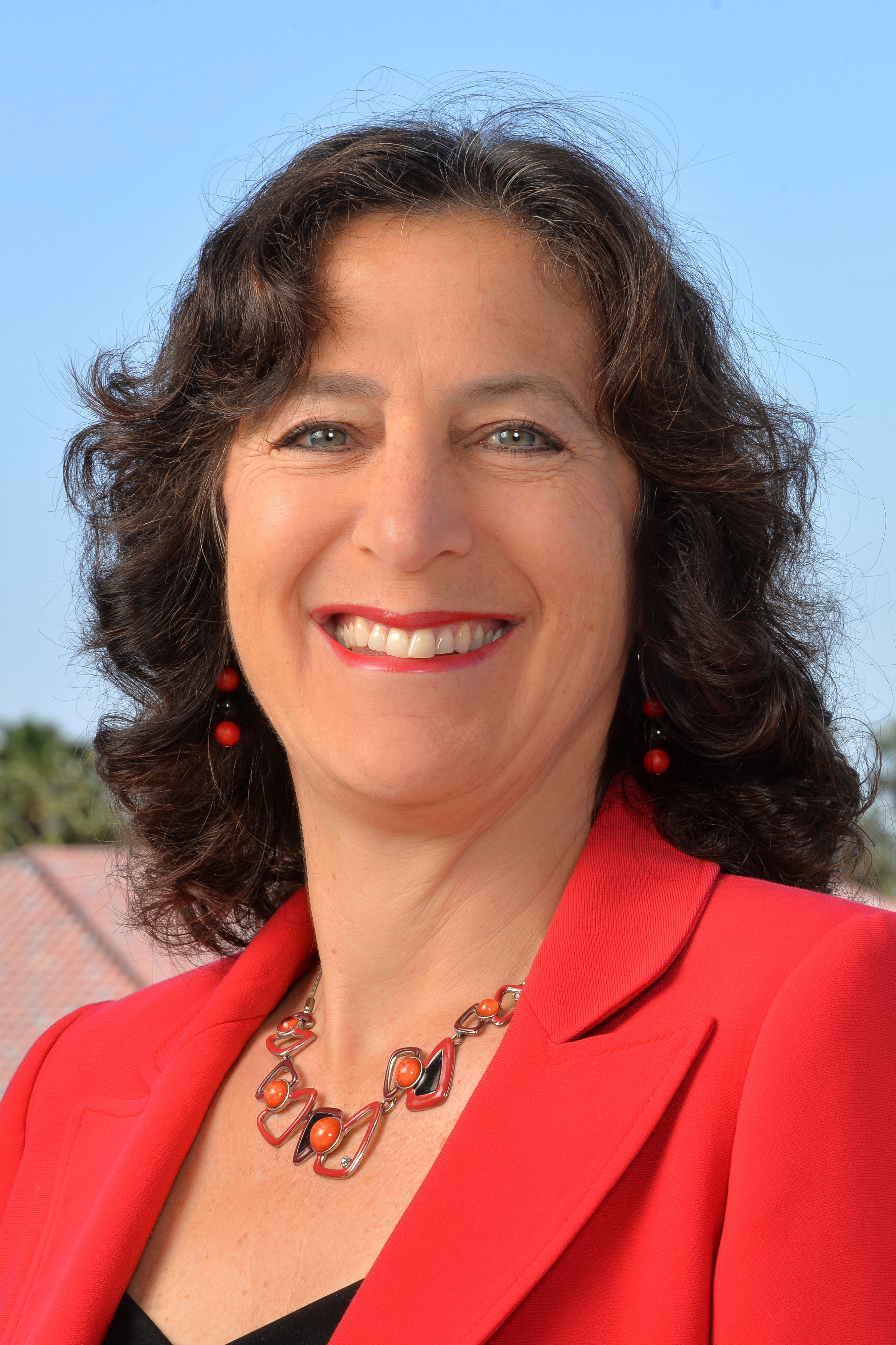}}]{Andrea J. Goldsmith (Fellow, IEEE)}
 received the B.S., M.S., and Ph.D. degrees in electrical engineering from University of California at Berkeley.  She is currently the Stephen Harris Professor with the School of Engineering and a Professor of electrical engineering with Stanford University. She has authored the book Wireless Communications and coauthored the books MIMO Wireless Communications and Principles of Cognitive Radio (Cambridge University Press), and an inventor on 29 patents. Her research interests are in information theory, communication theory,  signal processing, and their application to wireless communications, interconnected systems, and neuroscience. Dr. Goldsmith is a member of the National Academy of Engineering and the American Academy of Arts and Sciences, and a fellow of Stanford. She has received several awards for her work, including the IEEE Sumner Technical Field Award, the ACM Athena Lecturer Award, the ComSoc Armstrong Technical Achievement Award, the WICE Mentoring Award, and the Silicon Valley/San Jose Business Journal’s Women of Influence Award. She is currently the Founding Chair of the IEEE Board of Directors Committee on Diversity, Inclusion, and Ethics. She served as the President for the IEEE Information Theory Society in 2009, as the Founding Chair for its student committee, and as the Founding Editor-in-Chief for the IEEE Journal on Selected Areas of Information Theory. She has also served on the Board of Governors for both the IEEE Information Theory and Communications Societies. At Stanford, she has served as a Chair for Stanford’s Faculty Senate and for multiple terms as a Senator, and on its Academic Council Advisory Board, Budget Group, Committee on Research, Planning and Policy Board, Commissions on Graduate and on Undergraduate Education, Faculty Women’s Forum Steering Committee, and Task Force on Women and Leadership. 
\end{IEEEbiography}

% %If you have an EPS/PDF photo (graphicx package needed), extra braces are
 
\vspace{11pt}

% \end{IEEEbiography}
\end{document}